\documentclass[sigconf]{acmart}

\AtBeginDocument{%
  \providecommand\BibTeX{{%
    \normalfont B\kern-0.5em{\scshape i\kern-0.25em b}\kern-0.8em\TeX}}}

\copyrightyear{2020}
\acmYear{2020}
\setcopyright{acmcopyright}\acmConference[CCS '20]{Proceedings of the 2020 ACM SIGSAC Conference on Computer and Communications Security}{November 9--13, 2020}{Virtual Event, USA}
\acmBooktitle{Proceedings of the 2020 ACM SIGSAC Conference on Computer and Communications Security (CCS '20), November 9--13, 2020, Virtual Event, USA}
\acmPrice{15.00}
\acmDOI{10.1145/3372297.3417238}
\acmISBN{978-1-4503-7089-9/20/11}
\usepackage{url}
\usepackage{amsmath,amsfonts,amsthm}
\usepackage{algorithmic}
\usepackage{graphicx}
\usepackage{textcomp}
\usepackage{xcolor}
\usepackage{float}
\usepackage{epstopdf}
\usepackage{multicol}
\usepackage{booktabs}
\usepackage{subfigure}
\usepackage{hyperref}

\usepackage{bbm}
\usepackage[bottom,hang]{footmisc}
\usepackage{adjustbox}
\usepackage{xpatch}
\DeclareMathOperator*{\argmin}{argmin}
\xpretocmd{\proof}{\setlength{\parindent}{0pt}}{}{}

\newcommand{\myparatight}[1]{\smallskip\noindent{\bf {#1}:}~}
\newtheoremstyle{mytheoremstyle} %
    {\topsep}                    %
    {\topsep}                    %
    {\itshape}                   %
    {}                           %
    {\scshape}                   %
    {.}                          %
    {.5em}                       %
    {}  %
    \theoremstyle{mytheoremstyle}

\newtheorem{innercustomthm}{Theorem}
\newenvironment{customthm}[1]
  {\renewcommand\theinnercustomthm{#1}\innercustomthm}
  {\endinnercustomthm}
\begin{document}
\fancyhead{}
\fancyfoot[C]{\thepage}
\title{GAN-Leaks: A Taxonomy of Membership Inference Attacks against Generative Models}

\author{Dingfan Chen}
\affiliation{CISPA Helmholtz Center for Information Security}

\author{Ning Yu}
\affiliation{University of Maryland, College Park \\ Max Planck Institute for Informatics}

\author{Yang Zhang}
\affiliation{CISPA Helmholtz Center for Information Security}

\author{Mario Fritz}
\affiliation{CISPA Helmholtz Center for Information Security}

\renewcommand{\shortauthors}{Chen et al.}

\begin{abstract}
Deep learning has achieved overwhelming success, spanning from discriminative models to generative models. In particular, deep generative models have facilitated a new level of performance in a myriad of areas, ranging from media manipulation to sanitized dataset generation. Despite the great success, the potential risks of privacy breach caused by generative models have not been analyzed systematically. In this paper, we focus on membership inference attack against deep generative models that reveals information about the training data used for victim models. Specifically, we present the first taxonomy of membership inference attacks, encompassing not only existing attacks but also our novel ones. In addition, we propose the first generic attack model that can be instantiated in a large range of settings and is applicable to various kinds of deep generative models.  Moreover, we provide a theoretically grounded attack calibration technique, which consistently boosts the attack performance in all cases, across different attack settings, data modalities, and training configurations. We complement the systematic analysis of attack performance by a comprehensive experimental study, that investigates the effectiveness of various attacks w.r.t. model type and training configurations, over three diverse application scenarios (i.e., images, medical data, and location data).\footnote{Our code is available at \href{https://github.com/DingfanChen/GAN-Leaks}{https://github.com/DingfanChen/GAN-Leaks}.} 
\end{abstract}



\maketitle

\section{Introduction}

Over the last few years, two categories of deep learning techniques have made tremendous progress. The discriminative model has been successfully adopted in various prediction tasks, such as image classification~\cite{KSH12,SZ15,SLJSRAEVR15,HZRS16} 
and speech recognition~\cite{hinton2012deep,GMH13}. The generative model, on the other hand, has also gained increasing attention and has delivered appealing applications including photorealistic image synthesis~\cite{GPMXWOCB14,LW16,PKDDE16,YBSAL19}, text and sound generation~\cite{VTBE15,AGLMZ17,ODZSVGKSK16,MKGKJSCB17}, sanitized dataset generation~\cite{beaulieu2019privacy,AMCC17,ZJW18,XLWWZ18,JYS19}, etc. Most of such applications are supported by deep generative models, e.g., the generative adversarial networks (GANs)~\cite{GPMXWOCB14,RMC15,SGZCRC16,ACB17,GAADC17,KALL18,BDS19,KLA19,KLAHLA20,YLZMDF20} and variational autoencoder (VAE)~\cite{KW14,RMW14,YYSL16}.%

In line with the growing trend of deep learning in real business, many companies collect and process customer data which is then used to develop deep learning models for commercial use. %
However, data privacy violations frequently happened due to data misuse with an inappropriate legal basis, e.g., the misuse of National Health Service data in the DeepMind project.\footnote{\url{https://news.sky.com/story/google-received-1-6-million-nhs-patients-data-on-an-inappropriate-legal-basis-10879142}}
Data privacy can also be challenged by malicious users who intend to infer the original training data. The resulting privacy breach would raise serious issues as training data contains sensitive attributes such as diagnosis and income. One such attack is membership inference attack (MIA)~\cite{DSSUV15,BBHM16,SSSS17,HZHBTWB19,HMDC19,SZHBFB19} which aims to identify if a data record was used to train a machine learning model. Overfitting is the major cause for the feasibility of MIA, as the learned model tends to memorize training inputs and perform better on them.

While numerous literature is dedicated to MIA against discriminative models~\cite{SSSS17,LBWBWTGC18,YGFJ18,SZHBFB19,MSCS19,JSBZG19,LZ20}, the attack on generative models has not received equal attention, despite its practical importance. For instance, GANs have been applied to health record data and medical images~\cite{CBMDSS18,FKAGG18,yi2019generative} whose membership is sensitive as it may reveal a patient's disease history. Moreover, recent works in privacy preserving data sharing~\cite{beaulieu2019privacy,AMCC17,ZJW18,XLWWZ18,JYS19,COF20} propose to impose (membership) privacy constraints during GANs training for sanitized data generation. Understanding the membership privacy leakage under a practical threat model helps shed light on future research in this area.

Nevertheless, this is a highly challenging task from the adversary side. Unlike discriminative models, the victim generative models do not directly provide confidence values about the overfitting of data records, and thus leave little clues for conducting membership inference. In addition, current GAN models inevitably underrepresent certain data samples, i.e., encounter mode dropping and mode collapse, which pose additional difficulty to the attacker.

Unfortunately, none of the existing works~\cite{HMDC19,HHB19} provides a generic attack applicable to varying types of generative models. Nor do they report a complete and practical analysis of MIA against deep generative models. For example, Hayes~et~al.~\cite{HMDC19} %
do not consider the realistic situation where the GAN's discriminator is not accessible but only the generator is released. Hilprecht~et~al.~\cite{HHB19} investigate only on small-scale image datasets and do not involve white-box attack against GANs.  This motivates our contributions towards a simple and generic approach as well as a more systematic analysis. In general, we make the following contributions in the paper.

\begin{figure}[!t]
\centering
\includegraphics[width=0.9\columnwidth]{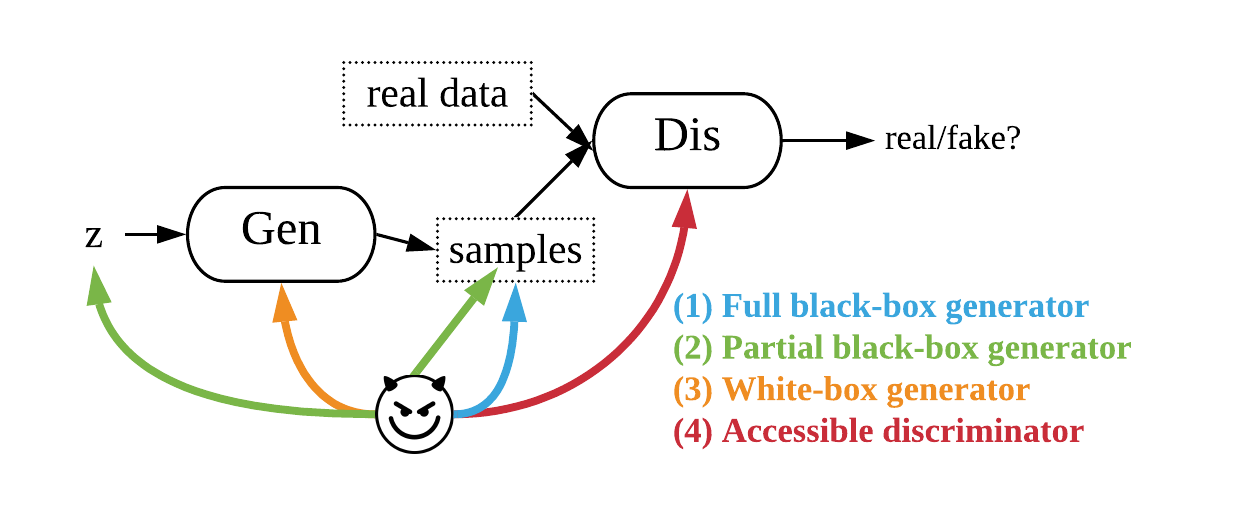}
\caption{Taxonomy of attack models against GANs. \textbf{Gen}: generator; \textbf{Dis}: discriminator; \textbf{z}: latent code input to Gen.}
\label{fig:taxonomy}
\end{figure}

\myparatight{Taxonomy of Membership Inference Attacks against Deep Generative Models} We conduct a pioneering study to categorize attack settings against deep generative models. Given the increasing order of the amount of knowledge about a victim model, the settings are benchmarked as (1) full black-box generator, (2) partial black-box generator, (3) white-box generator, and (4) accessible discriminator (full model). In particular, two of the settings, the partial black-box and white-box settings, are of practical value but have not been explored by previous works. We then establish the first taxonomy that comprises the existing and our proposed attacks. See \autoref{sec:taxonomy}, \autoref{table:taxonomy}, and \autoref{fig:taxonomy} for details.

\myparatight{Generic Attack Model and its Novel Instantiated Variants} We propose a simple and generic attack model (\autoref{sec:generic}) applicable to all the practical settings and various types of deep generative models. More specifically, our generic attack model can be instantiated to a preliminary low-skill attack for the full black-box setting (\autoref{sec:full bb}), a novel black-box optimization-based attack variant in the partial black-box (\autoref{sec:partial bb}), as well as a novel quasi-Newton optimization-based variant in the white-box settings (\autoref{sec:wb}). The consistent effectiveness of our attack model exhibited in all of the aforementioned settings bridges the assumption gap and performance gap between the full black-box attacks and discriminator-accessible attack in previous study~\cite{HMDC19,HHB19} through a complete performance spectrum (\autoref{sec:comparisons}). 

\myparatight{Novel Attack Calibration Technique}  To further improve the effectiveness of our attack model, we adjust our approach to each query sample and propose our novel attack calibration technique, which is naturally incorporated in our generic attack framework. Moreover, we prove its near-optimality under a Bayesian perspective. Through extensive experiments, we validate that our attack calibration technique boosts the attack performance noticeably in all cases, across different attack settings, data modalities, and training configurations. See \autoref{sec:calibration} for detailed explanation and \autoref{sec:exp_calibration} for experiment results.

\myparatight{Systematic Analysis in Each Setting} We progressively investigate attacks in each setting in the increasing order of amount of knowledge to adversary. See \autoref{sec:exp_fbb} to \autoref{sec:exp_wb} for detailed elaboration. In each setting, our research spans several orthogonal dimensions including three datasets with diverse modalities (\autoref{sec:setup}), five victim GAN models that were the state-of-the-art at their release time (\autoref{sec:setup}), two analysis study w.r.t. GAN training configuration (\autoref{sec:ablation}), attack performance gains introduced by attack calibration (\autoref{sec:calibration} and \autoref{sec:exp_calibration}) and differential private defense (\autoref{sec:dp}).

\section{Related Work}
\label{sec:related}

\myparatight{Generative Models}
Generative models are designed for approximating the probability distribution of the real data. In general, this is done by defining a parametric family of densities and finding the optimal parameters that either maximize the real data likelihood or minimize the divergence between generated and real data distribution. Recent generative models exploit the representation power of deep neural networks for constituting an exceptionally rich parametric family, resulting in tremendous success in modeling high-dimensional data distribution. In this work, we investigate the most widely used deep generative models, namely the generative adversarial networks (GANs)~\cite{GPMXWOCB14,RMC15,SGZCRC16,ACB17,GAADC17,KALL18,BDS19,KLA19,KLAHLA20,YLZMDF20} and variational autoencoders (VAEs)~\cite{DKB15,DSB17,KD18}. %
Briefly speaking, GANs are trained to minimize the divergence between the generated and real data distribution, while VAEs maximize a lower bound of the real data log-likelihood. %

\myparatight{Membership Inference Attacks (MIAs)}
Shokri~et~al.~\cite{SSSS17} specifies the first MIA against discriminative models in the black-box setting, where an attack has access to the victim model's full response (i.e., confidence scores for all classes) for a given input query. They propose to train shadow models that imitate the behavior of the victim model,
which generates data to train an attacker model.

Hayes~et~al.~\cite{HMDC19} consider MIA against GANs and also propose to retrain a shadow model of the victim model in the black-box case. They then check the discriminator's output scores to query inputs and set a threshold such that all the query inputs with scores larger than the threshold will be classified as in the training set.  

Another concurrent study by Hilprecht~et~al.~\cite{HHB19} investigates MIA against both GANs and VAEs. For VAEs, they assume the accessibility of the full model and propose to threshold the $L_2$ reconstruction error; For GANs, they only consider the full black-box setting. Their black-box attack is similar to ours in spirit, as they count the number of generated samples that are inside an $\epsilon$-ball of the query, while we exploit the reconstruction distance instead. 

\myparatight{Differential Privacy (DP)}
Differential privacy~\cite{DR14} is designed to protect the membership privacy of individual samples and is by constructing a defense mechanism against MIA. Recent works propose to train GAN models with differential privacy constraint~\cite{beaulieu2019privacy,AMCC17,ZJW18,XLWWZ18,JYS19,COF20} and publicize the DP-trained models instead of the raw data, which allows sharing sensitive data while preserving privacy. The differential privacy constraint is fulfilled by replacing the regular stochastic gradient descent with differential private stochastic gradient descent (DP-SGD)~\cite{ACGMMTZ16}, which injects calibrated noise in training gradients. As a result, it perturbs data-related objective functions and mitigates inference attacks.

\section{Background}
\subsection{Generative Model}
\label{sec:background generative}

\myparatight{Generative Adversarial Networks (GANs)} 
GANs consist of two neural network modules, a generator $G$ and a discriminator $D$, which are trained simultaneously in an adversarial manner. The generator takes random noise $z$ (latent code) as input and generates samples that approximate the training data distribution, while the discriminator receives samples from both the generator and training dataset and is trained to differentiate the two sources. During training, these two modules compete and evolve, such that the generator learns to generate more and more realistic samples aiming at fooling the discriminator, while the discriminator learns to tell the two sources apart more accurately. The training objective can be formulated as
\begin{equation*}
  \min_{\theta_G} \max_{\theta_D} \mathbb{E}_{x\sim P_\text{data}}[\log(D_{\theta_D}(x))] + \mathbb{E}_{z\sim P_z}[\log(1-D_{\theta_D}(G_{\theta_G}(z)))]
\end{equation*}
where $\theta_G, \theta_D$ denote the parameters of the generator and the discriminator. $P_\text{data}$ is the real data distribution, while the $P_z$ is the prior distribution of the latent code. The first term in the objective forces the discriminator to output high score given real data sample. The second term makes discriminator output low score on generated samples, while the generator is trained to maximize the discriminator output score. Once the training is done, the discriminator is no longer useful and will normally be discarded. The generator will receive new latent code samples $z$ drawn from the known prior distribution (normally Gaussian) and output the synthetic data samples, which will be collected and used for the downstream task.

\myparatight{Variational Autoencoder (VAE)} 
VAE is another widely used generative framework~\cite{KW14,RMW14,YYSL16} consists of an encoder and a decoder, which are cascaded to reconstruct data with pre-defined similarity metrics, e.g. $L_1$/$L_2$ loss. The encoder maps data into a latent space, while the decoder maps the encoded latent representation back to the data space. %
The VAE objective is composed of the reconstruction error and the prior regularization over the latent code distribution. Formally,
\begin{align*}
\label{eq:vae_loss}
\min_{\theta,\phi}-\mathbb{E}_{q_\phi(z|x)}[p_\theta(x|z)] + KL(q_\phi(z|x) \Vert P_z)
\end{align*}
where $z$ denotes the latent code, $x$ denotes the input data, $q_\phi(z|x)$ is the probabilistic encoder parameterized by $\phi$ which is introduced to approximate the intractable true posterior, $p_\theta(x|z)$ represents the probabilistic decoder parameterized by $\theta$, and $KL(\cdot \Vert \cdot)$ denotes the KL divergence. In practice, $q_\phi(z|x)$ is always constrained to be uni-modal Gaussian and $z$ is sampled via the reparameterization trick, which results in a closed-form derivation of the second term.

\myparatight{Hybrid Model} GANs often suffer from mode collapse and mode dropping issues, i.e., failing to generate appearances relevant to some training samples (low recall), due to the lack of explicit supervison (e.g. data reconstruction) for promoting data mode coverage. VAEs, on the contrary, attain better data coverage but often lack flexible generation capability (low precision). Therefore, a hybrid model, VAEGAN~\cite{LSLW16,BFS19}, is proposed to jointly train a VAE and a GAN, where the VAE decoder and the GAN generator are collapsed into one by sharing trainable parameters. The GAN discriminator is trained to complement the low-level $L_1$ or $L_2$ reconstruction loss, in order to improve the generation quality of fine-grained details.

\subsection{Membership Inference}
\label{sec:background MI}

We formulate the membership inference attack as a binary classification task where the attacker aims to classify whether a sample $x$ has been used to train a victim generative model. Formally, we define $$ \mathcal{A}: (x, \mathcal{M}(\theta)) \rightarrow \{0,1\} $$ where the attack model $\mathcal{A}$ output 1 if the attacker infers that the query sample $x$ is included in the training set, and 0 otherwise. $\theta$ denotes the victim model parameters while $\mathcal{M}$ represents the general model publishing mechanism, i.e., type of access available to the attacker. For example, the $\mathcal{M}$ is an identity function for the white-box access case and can be the inference function for the black-box case. For simplicity, we may omit the dependence on $\mathcal{M}$ if the type of access is irrelevant for illustration. With a Bayesian perspective~\cite{SDSOJ19}, the optimal attacker aims to compute the probability $P(x\in D_{\text{train}}|  x, \theta)$ and predict the query sample to be in the training set if the log-likelihood ratio is non-negative, i.e. the query sample is more likely to be contained in the training set than not. Mathematically,
\begin{equation}
\mathcal{A}(x,\mathcal{M}(\theta)) = \mathbbm{1}  \left[ \log \frac{P(x\in D_{\text{train}}|  x, \theta)}{P(x\notin D_{\text{train}}|  x, \theta)} \geq 0 \right]
\end{equation}
where $\mathbbm{1}(\cdot)$ is the indicator function, and the training set is denoted by $D_\text{train}$. We denote the query sample set as $S=\{(x_i,m_i)\}_{i=1}^N$ that contains both training set samples ($x_i \in D_{\text{train}}, m_i=1$) as well as hold-out set samples ($x_i \notin {D_{\text{train}}}, m_i=0$), where $m$ is the membership indicator variable.  The true positive and true negative rate of the attacker can be measure by $\mathbb{E}_{x_i}  [P(\mathcal{A}(x_i,\mathcal{M}(\theta))=1 | m_i=1)]$ and  $\mathbb{E}_{x_i}  [P(\mathcal{A}(x_i,\mathcal{M}(\theta))=0 | m_i=0)]$, respectively.

\section{Taxonomy}
\label{sec:taxonomy} 

\begin{table}[!t]
\centering
\begin{adjustbox}{max width=\columnwidth}
\begin{tabular}{lcccc}
\toprule
&  Latent & Gen- & Dis- \\
& code & erator & criminator \\
\midrule
\cite{HMDC19} full black-box & $\times$ & $\blacksquare$ & $\times$ \\
\cite{HHB19} full black-box & $\times$ & $\blacksquare$ & $\times$ \\
Our full black-box (\autoref{sec:full bb}) & $\times$ & $\blacksquare$ & $\times$ \\
Our partial black-box (\autoref{sec:partial bb}) & $\checkmark$ & $\blacksquare$ & $\times$ \\
Our white-box (\autoref{sec:wb}) & $\checkmark$ & $\square$ & $\times$ \\
\cite{HMDC19} accessible discriminator (full model) & $\checkmark$ & $\square$  & $\checkmark$ \\
\bottomrule
\end{tabular}
\end{adjustbox}
\caption {Taxonomy of attack settings against GANs over the previous work and ours. ($\times$: without access; $\checkmark$: with access; $\blacksquare$: black-box; $\square$: white-box). 
}
\label{table:taxonomy}
\end{table}
 
The attack scenarios can be categorized into either white-box or black-box one. In the white-box setting, the adversary has access to the victim model internals, whereas in the black-box setting, the internal workings are unknown to the attackers. For attacks against GANs, we further distinguish the settings based on the accessibility of GANs' components, i.e., the latent code, generator model, and the discriminator model, according to the following criteria: (1) whether the discriminator is accessible, (2) whether the generator is accessible, and (3) whether the latent code is accessible. 
We elaborate on each category in the following in a \textbf{decreasing} order of the amount of knowledge to attackers. Note that we define the taxonomy in a fully attack-agnostic way, i.e. the attacker can freely decide which part of the available information to use.

\subsection{Accessible Discriminator (Full Model)} 
\label{sec:taxonomy dis}

By construction, the discriminator is only used for the adversarial training and normally will be discarded after the training stage is completed. The only scenario in which the discriminator is accessible to the attacker is that the developers publish the whole GAN model along with the source code and allow fine-tuning. In this case, both the discriminator and the generator are accessible to the adversary in a white-box manner. This is the most knowledgeable setting for attackers. And the existing attack methods against discriminative models~\cite{SSSS17} can be applied to this setting. This setting is also considered in~\cite{HMDC19}, corresponding to the last row in \autoref{table:taxonomy}. In practice, however, the discriminator of a well-trained GAN is discarded without being deployed to APIs, and thus not accessible to attackers. %
We, therefore, devote less effort to investigating the discriminator and mainly focus on the following practical and generic settings where the attackers only have access to the generator. 

\subsection{White-box Generator}
\label{sec:taxonomy wb}

Following the common practice, researchers from the generative modeling community always publish their well-trained generators and code, which allows users to generate new samples and validate the results. 
This corresponds to the settings that the generator is accessible to the adversary in a white-box manner, i.e. the attackers have access to the internals of the generator. This scenario is also commonly studied in the community of differential privacy~\cite{DMNS06} and privacy preserving data generation~\cite{beaulieu2019privacy,AMCC17,ZJW18,XLWWZ18,JYS19,COF20}, where people enforce privacy guarantee by training and sharing their generative models instead of sharing the raw private data. Our attack model under this setting can serve as a practical tool for empirically estimating the privacy risk incurred by sharing the differentially private generative models, which offers clear interpretability towards bridging between theory and practice. However, this setting has not been explored by any previous work and is a novel case for constructing a membership inference attack against GANs. It corresponds to the second last row in \autoref{table:taxonomy} and \autoref{sec:wb}.

\subsection{Partial Black-box Generator (Known Input-output Pair)}
\label{sec:taxonomy pbb}

This is a less knowledgeable setting to attackers where they have no access to the internals of the generator but have access to the latent code of each generated sample. This is a practical setting where the developers retain ownership of their well-trained models while allowing users to control the properties of the generated samples by manipulating the latent code distribution~\cite{JCI19}, which is a desired feature for application scenarios such as GAN-based image processing~\cite{GSZ19} and facial attribute editing~\cite{KLA19,HZKSC18}. This is another novel setting and not considered in previous works \cite{HMDC19,HHB19}. It corresponds to the third last row in \autoref{table:taxonomy} and \autoref{sec:partial bb}.

\subsection{Full Black-box Generator (Known Output Only)}
\label{sec:taxonomy fbb}

This is the least knowledgeable setting to attackers where they are passive, i.e., unable to provide input, but are only permitted to access the generated samples set from the well-trained black-box generator. Hayes~et~al.~\cite{HMDC19} investigate attacks in this setting by retraining a local copy of the victim model. Hilprecht~et~al.~\cite{HHB19} count the number of generated samples that are inside an $\epsilon$-ball of the query, based on an elaborate design of distance metric. Our idea is similar in spirit to Hilprecht~et~al.~\cite{HHB19} but we score each query by the reconstruction error directly, which does not introduce additional hyperparameter while achieving superior performance. In short, we design a low-skill attack method with a simpler implementation (\autoref{sec:full bb}) that achieves comparable or better performance (\autoref{sec:exp_fbb}). Our attack and theirs correspond to the third, second, and first rows in \autoref{table:taxonomy}, respectively.

\section{Attack Model}
\label{sec:attack model}
\begin{table}[!t]
\centering
\begin{adjustbox}{max width=\columnwidth}
\begin{tabular}{cl}
\toprule
Notation & Description \\
\midrule
$\mathcal{A}$ & Attacker\\
$\mathcal{M}$ & model publishing mechanism \\
$D_{\text{train}}$ & Training set of the victim generator\\
$S$ & Query set \\
$\mathcal{R}$ & Attacker's reconstructor\\
$x$ & Query sample\\
$m$ & Membership indicator variable\\
$z$ & Latent code (input to the generator)\\ 
$\mathcal{G}_v $ & Victim generator\\
$\mathcal{G}_r$ & Attacker's reference generator, described in \autoref{sec:calibration}\\
$\theta_v$ & Victim model's parameter\\
$\theta_r$ & Attacker's reference model's parameter\\
\bottomrule
\end{tabular}
\end{adjustbox}
\caption {Notations.}
\label{table:notations}
\end{table}

\begin{figure*}[!t]
\centering
\subfigure[Generic attack model  (\autoref{sec:generic})]{
\includegraphics[width=0.46\columnwidth]{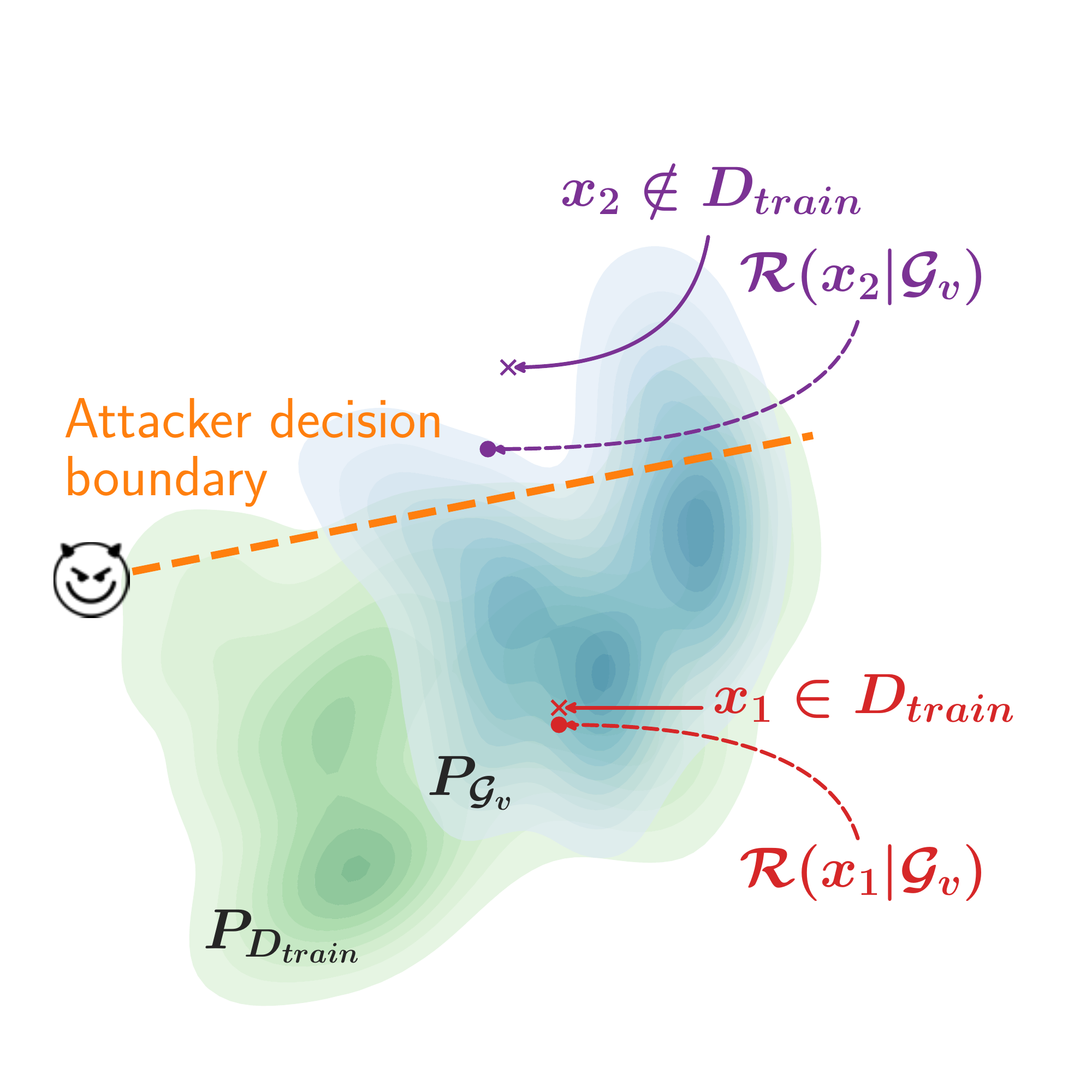}
\label{fig:diagram_generic}
}
\subfigure[Full black-box attack (\autoref{sec:full bb})]{
\includegraphics[width=0.46\columnwidth]{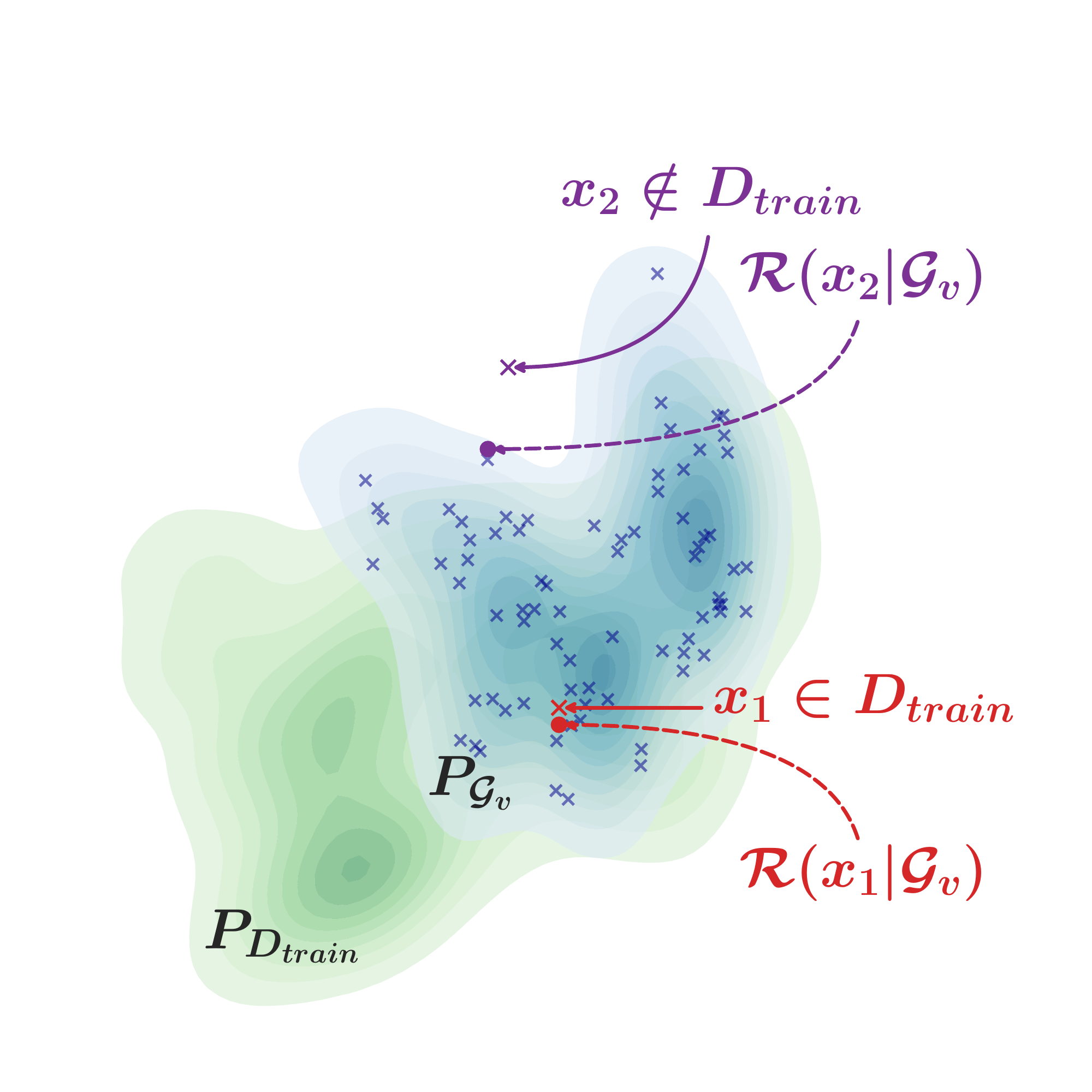}
\label{fig:diagram_fbb}
}
\subfigure[Partial black-box and white-box attack (\autoref{sec:partial bb} and \autoref{sec:wb})]{
\includegraphics[width=0.46\columnwidth]{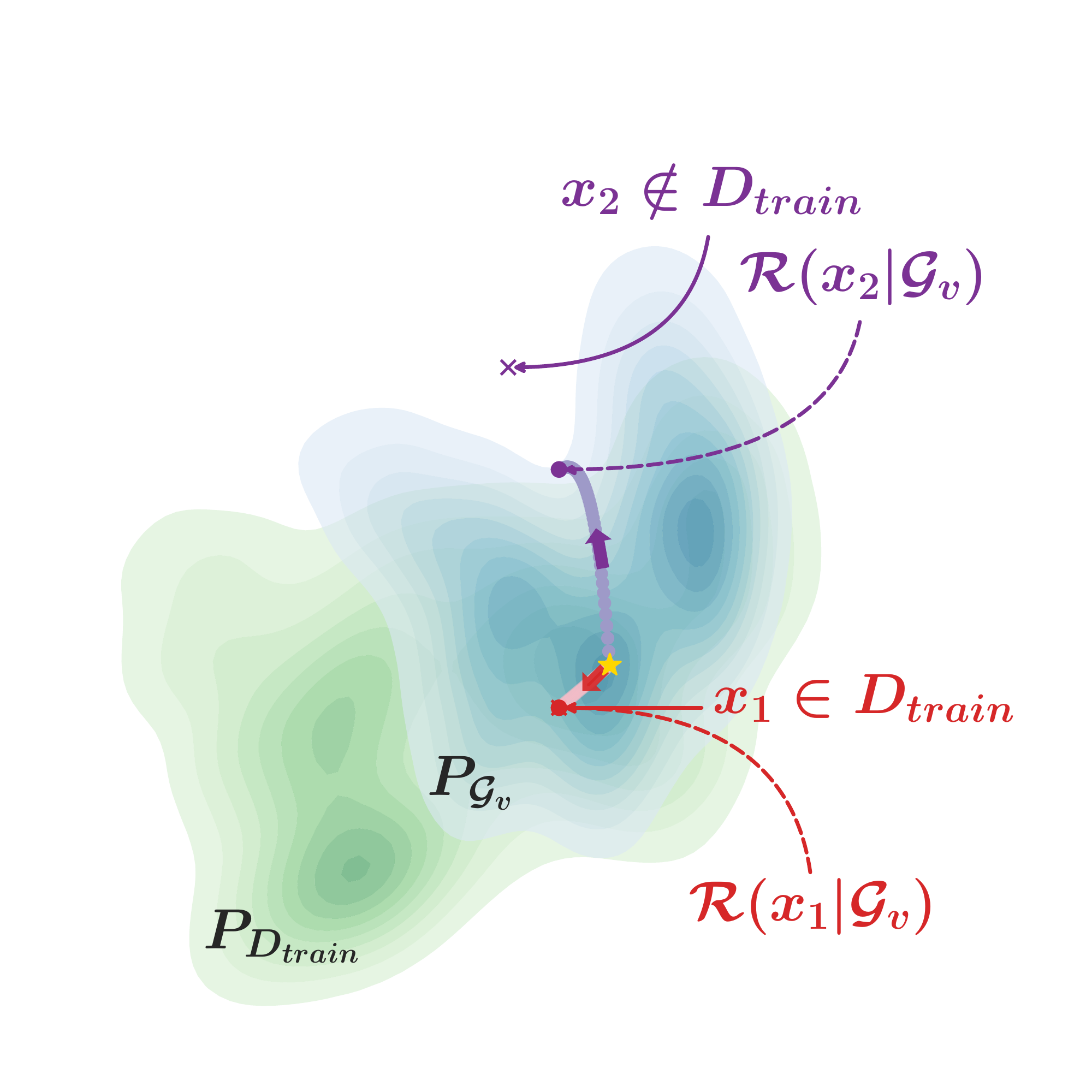}
\label{fig:diagram_pbb_wb}
}
\subfigure[Attack calibration (\autoref{sec:calibration})]{
\includegraphics[width=0.46\columnwidth]{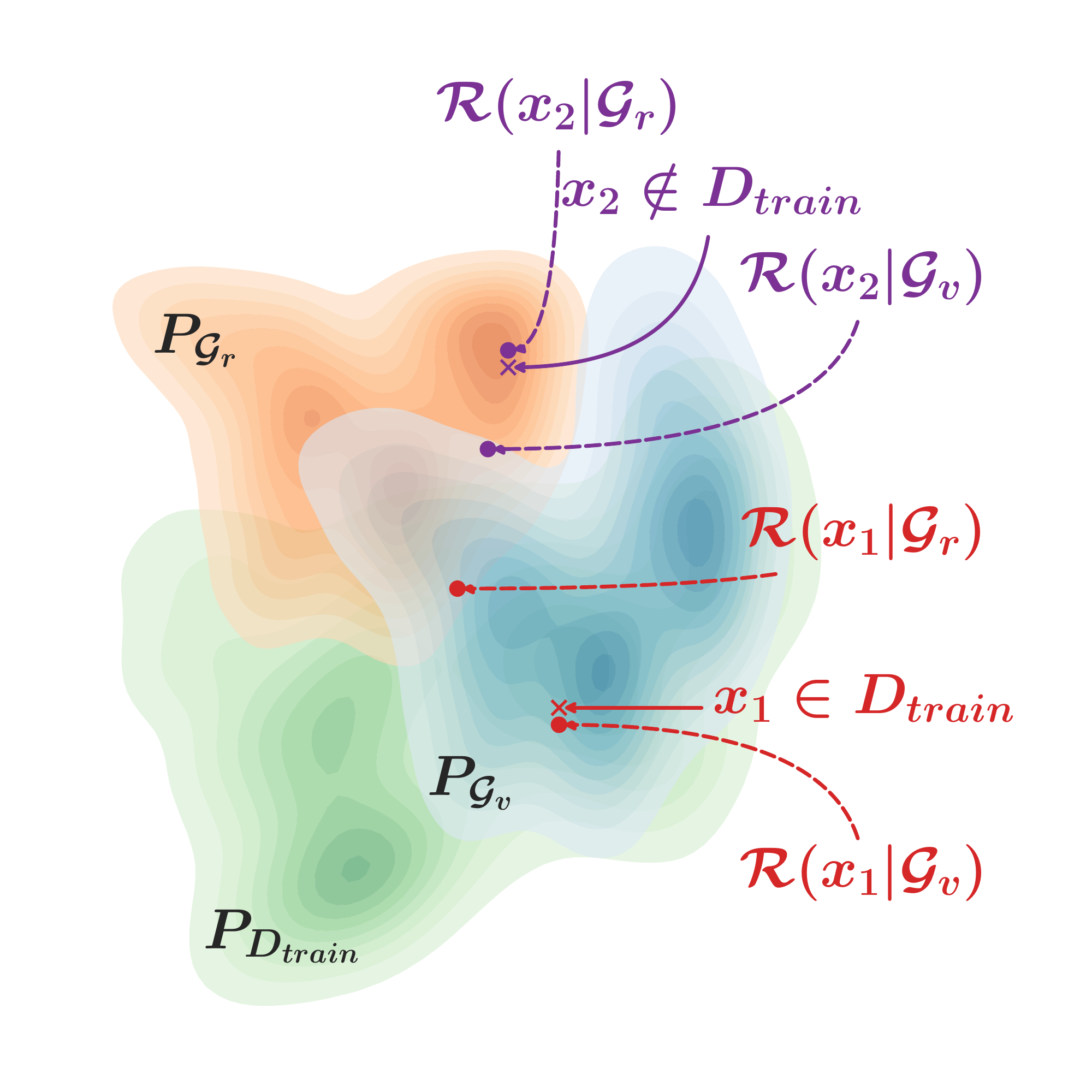}
\label{fig:diagram_calibration}
}
\caption{Diagram of our attacks. Mathematical notations refer to \autoref{table:notations}. $P$ represents data distribution.
$x_1$ belongs to $D_{\text{train}}$ so that it should be better represented by $\mathcal{G}_v$ with a smaller distance to its reconstructed copy $\mathcal{R}(x_1\vert\mathcal{G}_v)$. $x_2$ does not belong to $D_{\text{train}}$ so that it should have a larger distance to its best approximation $\mathcal{R}(x_2\vert\mathcal{G}_v)$ in $P_{\mathcal{G}_v}$. (a) Our generic attacker set a decision boundary based on the reconstruction distance to infer membership. (b) The best reconstruction is determined over random samples from $P_{\mathcal{G}_v}$ while in (c) it is found by optimization on the manifold of $P_{\mathcal{G}_v}$. (d) $P_{\mathcal{G}_r}$ is a third-party reference GAN distribution where the reconstruction distance is calibrated by the distance between $x$ and $\mathcal{R}(x\vert\mathcal{G}_r)$.
}
\end{figure*}

\subsection{Generic Attack Model}
\label{sec:generic}

As mentioned in \autoref{sec:background MI}, the optimal attacker computes the probability $P(m_i=1|x_i,\theta_v)$. Specifically for the generative model, we make the assumption that this probability should be proportional to the probability that the query sample can be generated by the generator. This assumption holds in general as the generative model is trained to approximate the training data distribution, i.e., $P_{\mathcal{G}_v} \approx P_{D_\text{train}}$ where $\mathcal{G}_v$ denotes the victim generator. And if the probability that the query sample is generated by the victim generator is large, it is more likely that the query sample is used to train the generative model. Formally,
\begin{equation}
\label{eq:bayes error}
P(m_i=1|x_i,\theta_v)\propto P_{\mathcal{G}_v}(x|\theta_v)
\end{equation}
However, computing the exact probability is intractable as the distribution of the generated data cannot be represented with an explicit density function. Therefore, we adopt the Parzen window density estimation~\cite{DHS12} and approximate the probability as below,
\begin{align}
P_{\mathcal{G}_v}(x|\theta_v
) &= \frac{1}{k}\sum_{i=1}^k \phi(x, \mathcal{G}_v(z_i)); \quad z_i \sim P_z \label{eq:kernel density 1}\\
&\approx \frac{1}{k}\sum_{i=1}^k \exp(-L(x,\mathcal{G}_v(z _i))); \quad z_i \sim P_z
\label{eq:kernel density 2}
\end{align}
where $\phi(\cdot,\cdot)$ denotes the kernel function, $L(\cdot,\cdot)$ is the general distance metric defined in \autoref{sec:distance metric}, and $k$ is the number of samples.  Note that this can be further simplified and well approximated using only few samples~\cite{BSI08}, as all of the terms in the summation of \autoref{eq:kernel density 1}, except for a few, will be negligible since $\phi(x,y)$ exponentially decreases with distance between $x,y$.

\subsection{Full Black-box Attack}
\label{sec:full bb}

We start with the least knowledgeable setting where an attacker only has access to a black-box generator $\mathcal{G}_v$. The attacker is allowed no other operation but blindly collecting $k$ samples from $\mathcal{G}_v$, denoted as $\{\mathcal{G}_v(\cdot)_i\}_{i=1}^k$. $\mathcal{G}_v(\cdot)$ indicates that the attacker has neither access nor control over latent code input. We then approximate the probability in \autoref{eq:kernel density 2} using the largest term which is given by the nearest neighbor to $x$ among $\{\mathcal{G}_v(\cdot)_i\}_{i=1}^k$.
Formally,
\begin{equation}
\mathcal{R}(x\vert\mathcal{G}_v) = \argmin_{\hat{x}\in\{\mathcal{G}_v(\cdot)_i\}_{i=1}^k}L(x, \hat{x})
\label{eq:fbb_rec}
\end{equation}
See \autoref{fig:diagram_fbb} for a diagram. This approximation bound the complete Parzen window from below, but in practice we observe almost no difference when incorporating more terms in the summation for a fixed $k$. However, we find the estimation more sensitive to $k$, and in general a larger $k$ leads to better reconstructions (\autoref{fig:fbb_numqueries}) but at the price of a higher query and computation cost. Throughout the experiments, we consider a practical and limited budget and choose $k$ to be of the same magnitude as the training dataset size.

\subsection{Partial Black-box Attack}
\label{sec:partial bb}

In some practical scenario discussed in \autoref{sec:taxonomy pbb}, the access to the latent code $z$ is permitted. We then propose to exploit $z$ in order to find a better reconstruction of the query sample and thus improve the $P_{\mathcal{G}_v}(x|\theta_v)$ estimation. Concretely, the attacker performs an black-box optimization with respect to $z$. Formally,
\begin{equation}
\mathcal{R}(x\vert\mathcal{G}_v) = \mathcal{G}_v(z^*)
\end{equation}
where
\begin{equation}
z^* = \argmin_z L\big(x, \mathcal{G}_v(z)\big)
\end{equation}

Without knowing the internals of $\mathcal{G}_v$, the optimization is not differentiable and no gradient information is available. As only the evaluation of function (forward-pass through the generator) is allowed by the access of $\{z, \mathcal{G}_v(z)\}$ pair, we propose to approximate the optimum via the Powell's Conjugate Direction Method~\cite{powell1964efficient}.

\subsection{White-box Attack}
\label{sec:wb}

In the white-box setting, we have the same reconstruction formulation as in \autoref{sec:partial bb}. See \autoref{fig:diagram_pbb_wb} for a diagram. More advantageously to attackers, the reconstruction quality can be further boosted thanks to access to the internals of $\mathcal{G}_v$. With access to the gradient information, the optimization problem can be more accurately solved by advanced first-order optimization algorithms~\cite{KB15,tieleman2012lecture,liu1989limited}. In our experiment, we apply the L-BFGS algorithm for its robustness against suboptimal initialization and its superior convergence rate in comparison to the other methods.

\begin{figure*}[!t]
\centering
\includegraphics[width=0.85\textwidth,trim={2cm 0 2cm 0},clip]{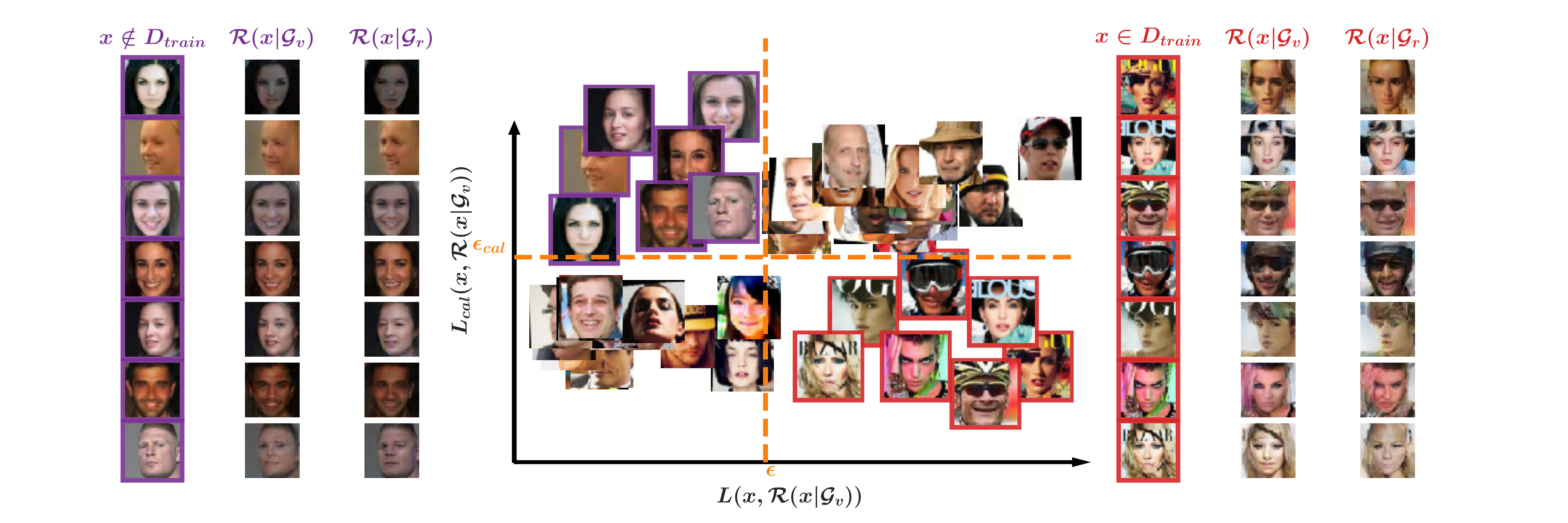}
\caption{The effectiveness of calibration when attacking PGGAN on CelebA. The x- and y-axes respectively represent the distance before ($L$) and after calibration ($L_{\text{cal}}$) between a query sample $x$ and its reconstruction $\mathcal{R}(x\vert\mathcal{G}_v)$. $\epsilon$ and $\epsilon_{\text{cal}}$ are the corresponding thresholds for classification. The false-positive (in purple frame) as well as the false-negative samples (in red frame) before ($L$) calibration can be corrected by calibration ($L_\text{cal}$).
}
\label{fig:calibration}
\end{figure*}

\subsection{Distance Metric}
\label{sec:distance metric}

Our distance metric $L(\cdot, \cdot)$ consists of three terms: the element-wise (pixel-wise) difference term $L_2$ targets low-frequency components, the deep image feature term $L_{\text{lpips}}$ (i.e., the Learned Perceptual Image Patch Similarity (LPIPS) metric~\cite{ZIESW18}) targets realism details, and the regularization term penalizes latent code far from the prior distribution. Mathematically,
\begin{align}
\label{eq:distance_metric}
L\big(x, \mathcal{G}_v(z)\big) = & \lambda_1 L_2\big(x, \mathcal{G}_v(z)\big) + \lambda_2 L_{\text{lpips}}\big(x, \mathcal{G}_v(z)\big) \nonumber \\
&+ \lambda_3 L_{\text{reg}}(z)
\end{align}
where
\begin{equation}
L_2\big(x, \mathcal{G}_v(z)\big) = \Vert x - \mathcal{G}_v(z)\Vert_2^2
\end{equation}
\begin{equation}
L_{\text{reg}}(z) = \big(\Vert z \Vert_2^2 - \text{dim}(z)\big)^2
\end{equation}
$\lambda_1$, $\lambda_2$ and $\lambda_3$ are used to enable/disable and balance the order of magnitude of each loss term. For non-image data, $\lambda_2 = 0$ because LPIPS is no longer applicable. For full black-box attack, $\lambda_3 = 0$ as the constraint $z\sim P_z$ is satisfied by the sampling process.

\subsection{Attack Calibration}
\label{sec:calibration}

We noticed that the reconstruction error is query-dependent, i.e., some query samples are more (less) difficult to reconstruct due to their intrinsically more (less) complicated representations, regardless of which generator is used. In this case, the reconstruction error is dominated by the representations rather than by the membership clues. We, therefore, propose to mitigate the query dependency by first independently training a reference GAN $\mathcal{G}_r$ with a relevant but disjoint dataset, and then calibrating our base reconstruction error according to the reference reconstruction error. Formally,
\begin{equation}
L_{\text{cal}}\big(x, \mathcal{R}(x\vert\mathcal{G}_v)\big) = L\big(x, \mathcal{R}(x\vert\mathcal{G}_v)\big) - L\big(x, \mathcal{R}(x\vert\mathcal{G}_r)\big)
\end{equation}
with $\mathcal{R}$ the reconstruction.
As demonstrated in \autoref{fig:calibration}, we show in the up-left quadrant the query samples in purple frame that are classified as \textbf{in} $D_{\text{train}}$ by $L$ and as \textbf{not in} $D_{\text{train}}$ by $L_{\text{cal}}$. They are false-positive to $L$ but are corrected to true-negative by $L_{\text{cal}}$. On the other hand, we show in the bottom-right quadrant the query samples in red frame that are classified as \textbf{not in} $D_{\text{train}}$ by $L$ and as \textbf{in} $D_{\text{train}}$ by $L_{\text{cal}}$. They are false-negative to $L$ but are corrected to true-positive by $L_{\text{cal}}$. We compare all these samples, their reconstructions from the victim generator $\mathcal{G}_v$, and their reconstructions from the reference generator $\mathcal{G}_r$ on the two sides of the plot. The false-positive samples by $L$ on the left-hand side are those with less complicated appearances such that their reconstruction errors are not high given arbitrary generators. In contrast, the false-negative samples by $L$ on the right-hand side are those with more complicated appearances such that their reconstruction errors are high given arbitrary generators. Our calibration can effectively mitigate these two types of misclassification that depend on sample representations.

As discussed in  \autoref{sec:background MI}, the optimal attacker aims to compute the membership probability 
\begin{equation}
    P(m_i = 1|\theta_v, x_i) = \mathbb{E}_S [P(m_i=1|\theta_v, x_i, S)] 
\end{equation}

\noindent Specifically, inferring the membership of the query sample $x_i$ amounts to approximating the value of $P(m_i=1|\theta_v, x_i, S)$~\cite{SDSOJ19}.  We show that our calibrated loss well approximate this probability by the following theorem, whose proof is provided in Appendix.

\begin{theorem}
\label{theorem:calibration} 
Given the victim model with parameter $\theta_v$, a query dataset $S$, the membership probability of a query sample $x_i$ is well approximated by the sigmoid of minus calibrated reconstruction error. 
\begin{equation}
P(m_i = 1|\theta_v, x_i, S) \approx \sigma ( -L_{\text{cal}} (x_i,\mathcal{R}(x_i|\mathcal{G}_v))  
\end{equation}
And the optimal attack is equivalent to 
\begin{align}
\label{eq:attack threshold}
\mathcal{A}(x_i,\mathcal{M}(\theta_v)) = \mathbbm{1}[L_{\text{cal}}(x_i,\mathcal{R}(x_i|\mathcal{G}_v)) < \epsilon ]
\end{align}
i.e., the attacker checks whether the calibrated reconstruction error of the query sample $x_i$ is smaller than a threshold $\epsilon$.
\end{theorem}

\begin{figure*}[!t]
\centering
\subfigure[PGGAN]{
\includegraphics[width=0.33\columnwidth]{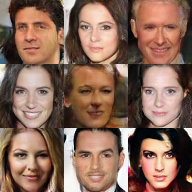}
\label{fig:pggan}
}
\subfigure[WGANGP]{
\includegraphics[width=0.33\columnwidth]{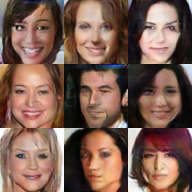}
\label{fig:wgangp}
}
\subfigure[DCGAN]{
\includegraphics[width=0.33\columnwidth]{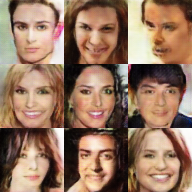}
\label{fig:dcgan}
}
\subfigure[VAEGAN]{
\includegraphics[width=0.33\columnwidth]{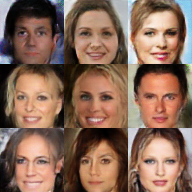}
\label{fig:vaegan}
}
 \subfigure[PGGAN w/ DP]{
\includegraphics[width=0.33\columnwidth]{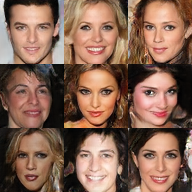}
\label{fig:pggan_dp}
}
\caption{Generated images from different victim GAN models trained on CelebA.}
\label{fig:gan_eval}
\end{figure*}

In the white-box case, the reference model has the same architecture as the victim model as this information is accessible to the attacker. In the full black-box and partial black-box settings, $\mathcal{G}_r$ has irrelevant network architectures to $\mathcal{G}_v$, which is fixed across attack scenarios. The optimization on the well-trained $\mathcal{G}_r$ is the same as on the white-box $\mathcal{G}_v$. See \autoref{fig:diagram_calibration} for a diagram, and  
\autoref{sec:exp_calibration} for implementation details.

\section{Experiments}
\label{sec:experiment}

Based on the proposed taxonomy, we present the most comprehensive evaluation to date on the membership inference attacks against deep generative models. While prior studies have singled out few data sets from constraint domains on selected models, our evaluation includes three diverse datasets,  five different generative models, and systematic analysis of attack vectors -- including more viable threat models. Via this approach, we present key discoveries, that connect for the first time the effectiveness of the attacks to the model types, data sets, and training configuration.

\subsection{Setup}
\label{sec:setup}

\myparatight{Datasets}
We conduct experiments on three diverse modalities of datasets covering images, medical records, and location check-ins, which are considered with a high risk of privacy breach.

\emph{CelebA~\cite{LLWT15}} is a large-scale face attributes dataset with 200k RGB images. Images are aligned to each other based on facial landmarks, which benefits GAN performance. We select at most 20k images, center-crop them, and resize them to $64\times64$ before GAN training.

\emph{MIMIC-\uppercase\expandafter{\romannumeral3}~\cite{johnson2016mimic}} is a public Electronic Health Records (EHR) database containing medical records of $46,520$ intensive care unit (ICU) patients. We follow the same procedure as in \cite{CBMDSS18} to pre-process the data, where each patient is represented by a 1071-dimensional binary feature vector. We filter out patients with repeated vector presentations and yield $41,307$ unique samples. 

\emph{Instagram New-York~\cite{BHPZ17}} contains Instagram users' check-ins at various locations in New York at different time stamps from 2013 to 2017. We filter out users with less than 100 check-ins and yield $34,336$ remaining samples. For sample representation, we first select $2,024$ evenly-distributed time stamps. We then concatenate the longitude and latitude values of the check-in location at each time stamp, and yield a 4048-dimensional vector for each sample. The longitude and latitude values are either retrieved from the dataset or linearly interpolated from the available neighboring time stamps. We then perform zero-mean normalization before GAN training.

\myparatight{Victim GAN Models} We select PGGAN~\cite{KALL18}, WGANGP~\cite{GAADC17}, DCGAN~\cite{RMC15}, MEDGAN~\cite{CBMDSS18}, and VAEGAN~\cite{BFS19} into the victim model set, considering their pleasing performance on generating images and/or other data representations.

It is important to guarantee the high quality of well-trained GANs because attackers are more likely to target high-quality GANs with practical effectiveness. We noticed previous works~\cite{HMDC19,HHB19} only show qualitative results of their victim GANs. In particular, Hayes~et~al.~\cite{HMDC19} did not show visually pleasing generated results on the Labeled Faces in the Wild (LFW) dataset~\cite{huang2008labeled}. Rather, we present better qualitative results of different GANs on CelebA (\autoref{fig:gan_eval}), and further present the corresponding quantitative evaluation in terms of Fr\'{e}chet Inception Distance (FID) metric~\cite{HRUNH17} (\autoref{table:gan_eval}). A smaller FID indicates the generated image set is more realistic and closer to real-world data distribution. We show that our GAN models are in a reasonable range to the state of the art.

\begin{table}[!t]
\centering
\begin{adjustbox}{width=\columnwidth}
\begin{tabular}{lccccc|c}
\toprule
& PG- & WGAN- & DC- & VAE- &SOTA & PGGAN\\
& GAN & GP & GAN & GAN  &ref & w/ DP\\
\midrule
FID & 14.86 &  24.26 & 35.40 & 53.08  & 7.40 & 15.63\\
\bottomrule
\end{tabular}
\end{adjustbox}
\caption {FID for different GAN models trained on CelebA. ``SOTA ref'' represents the state-of-the-art result reported in \cite{BDS19} over $128\times 128$ ImageNet ILSVRC 2012 dataset~\cite{RDSKSMHKKBBF15}. ``w/ DP'' represents the GAN model with DP privacy protection~\cite{ACGMMTZ16} (see \autoref{sec:dp}).}
\label{table:gan_eval}
\end{table}

\myparatight{Attack Evaluation} The proposed membership inference attack is formulated as a binary classification given a threshold $\epsilon$ in \autoref{eq:attack threshold}. Through varying $\epsilon$, we measure the area under the receiver operating characteristic curve (AUCROC) to evaluate the attack performance. 

\subsection{Analysis Study}
\label{sec:ablation}

We first list two dimensions of analysis study across attack settings. There are also some other dimensions specifically for the white-box attack, which are elaborated in \autoref{sec:exp_wb}.

\subsubsection{GAN Training Set Size} Training set size is highly related to the degree of overfitting of GAN training. A GAN model trained with a smaller size tends to more easily memorize individual training images and is thus more vulnerable to membership inference attack. Moreover, training set size is the main factor that affects the privacy cost computation for differential privacy. Therefore, we evaluate the attack performance w.r.t. training set size. We exclude DCGAN and VAEGAN from evaluation since they yield unstable training for small training sets.
\begin{figure*}[!t]
\centering
\subfigure[]{
\includegraphics[width=0.55\columnwidth]{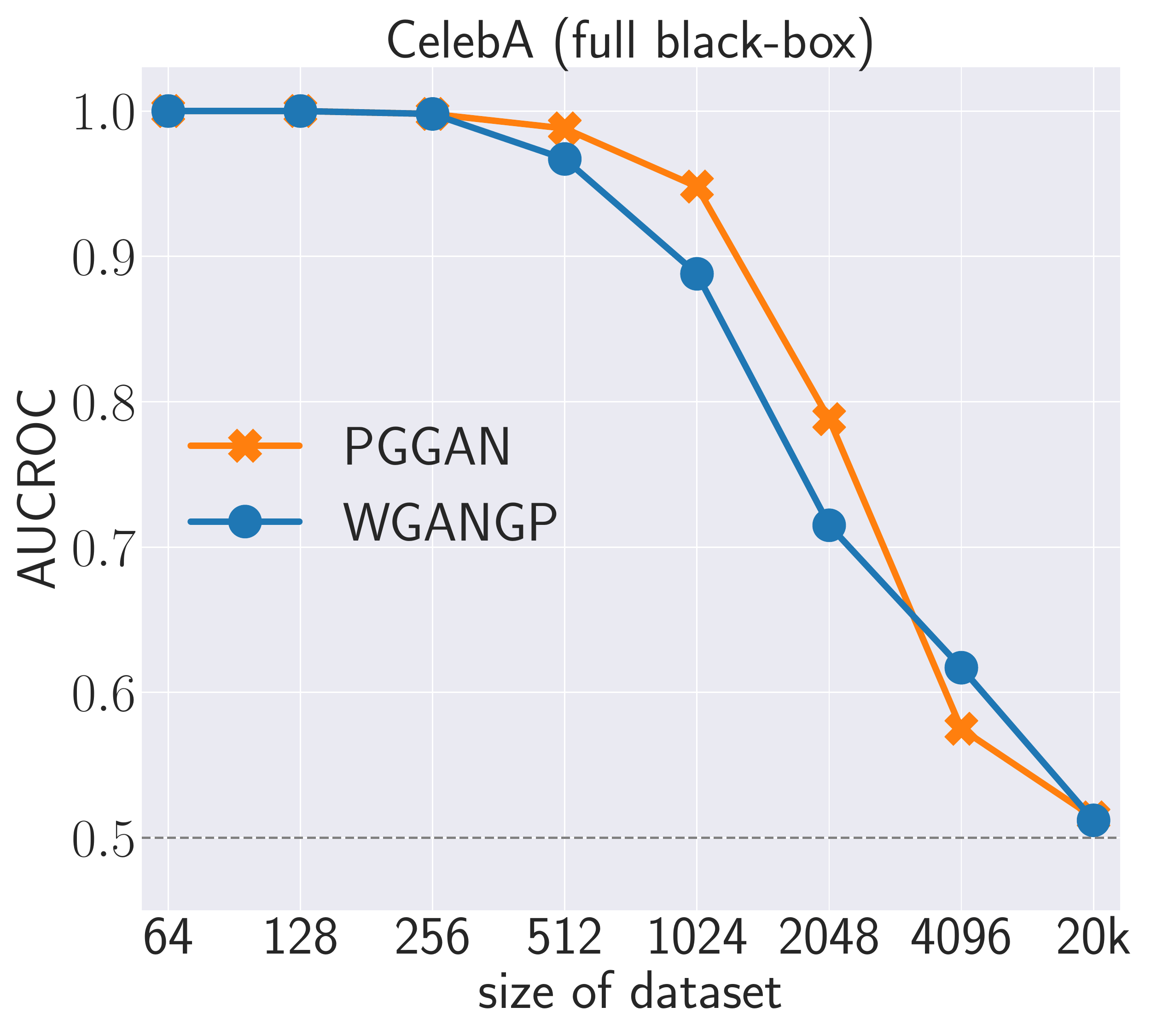}
\label{fig:bb_size_1}
}
\subfigure[]{
\includegraphics[width=0.55\columnwidth]{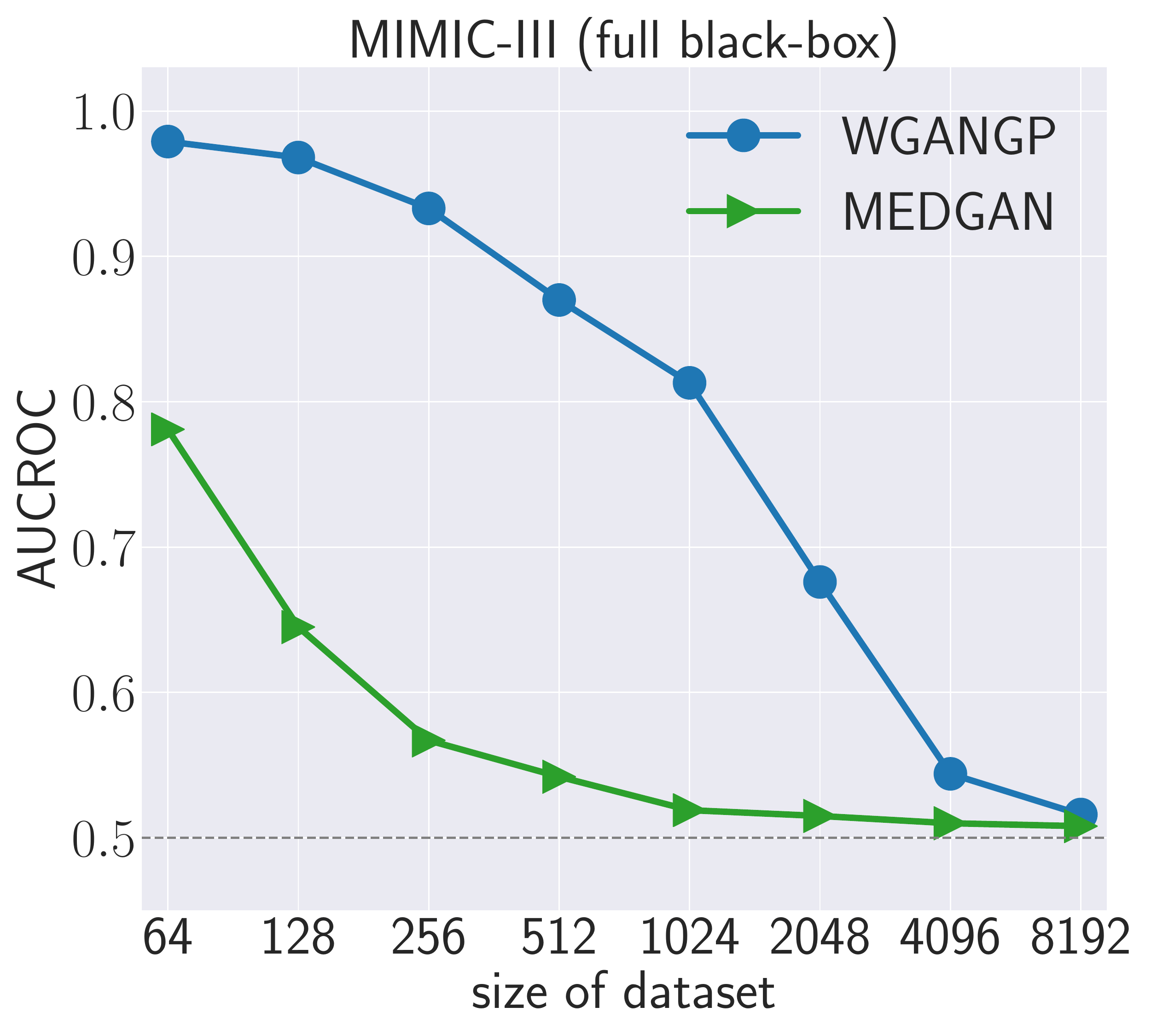}
\label{fig:bb_size_2}
}
\subfigure[]{
\includegraphics[width=0.55\columnwidth]{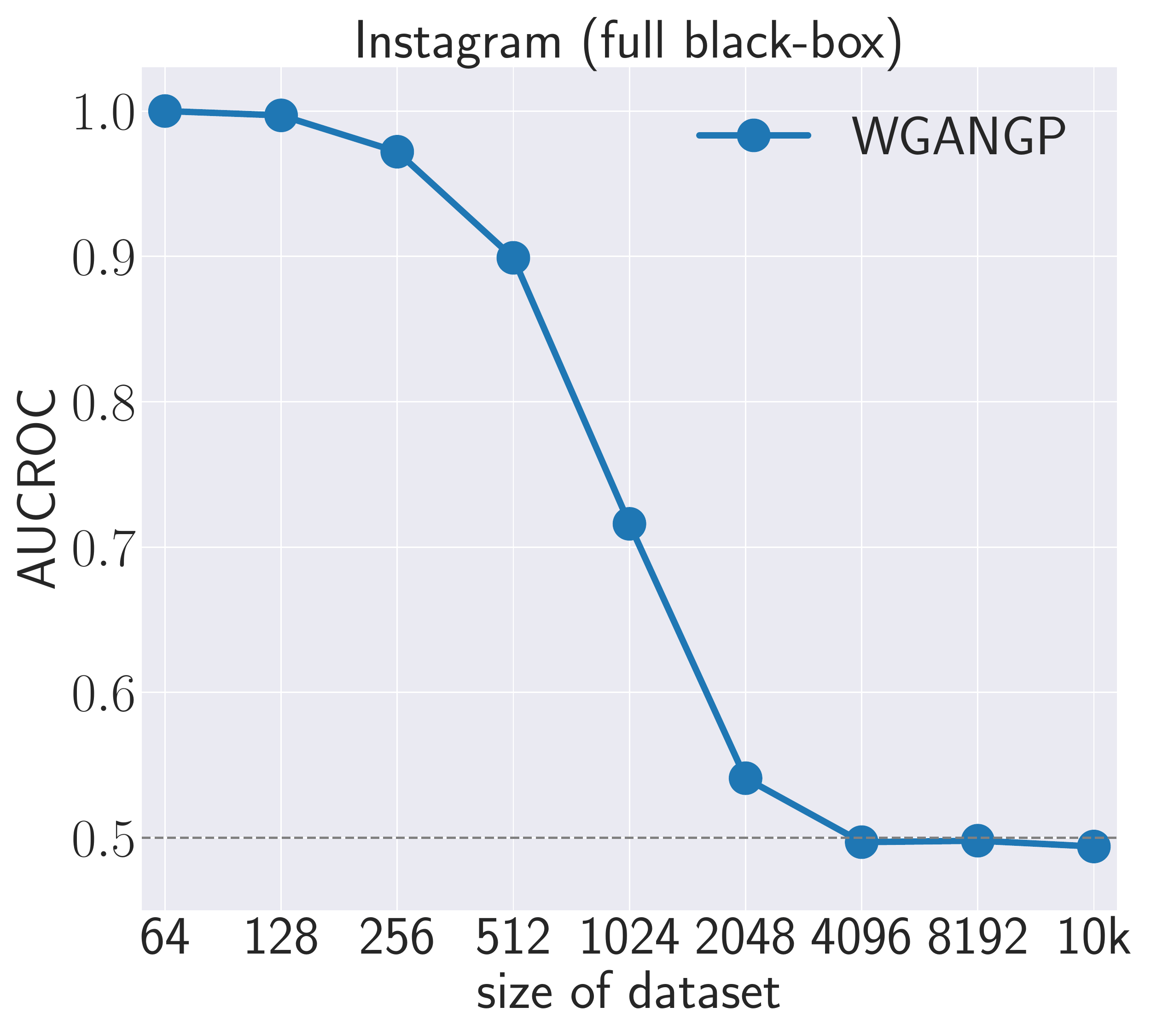}
\label{fig:bb_size_3}
}
\subfigure[]{
\includegraphics[width=0.55\columnwidth]{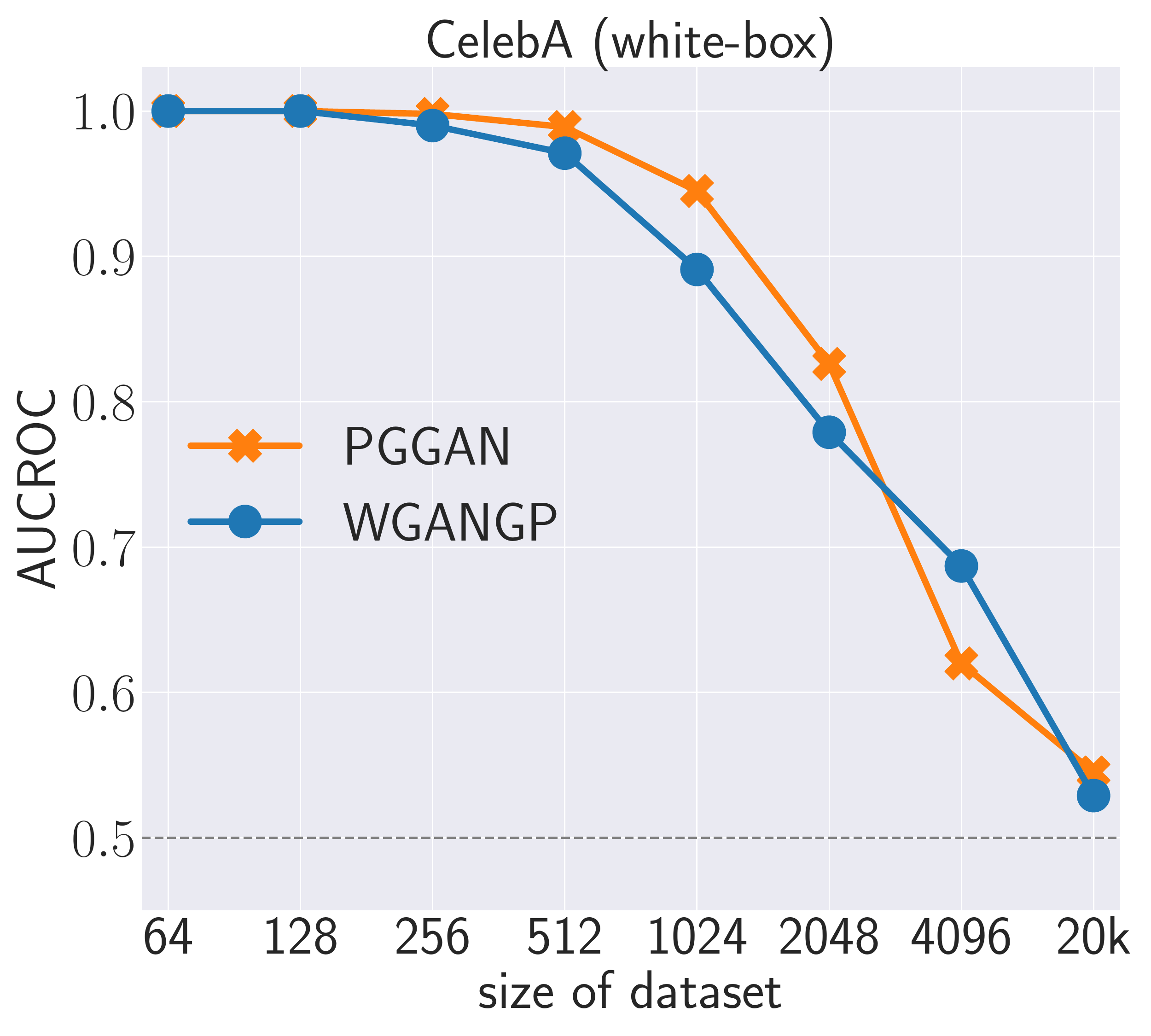}
\label{fig:wb_size_1}
}
\subfigure[]{
\includegraphics[width=0.55\columnwidth]{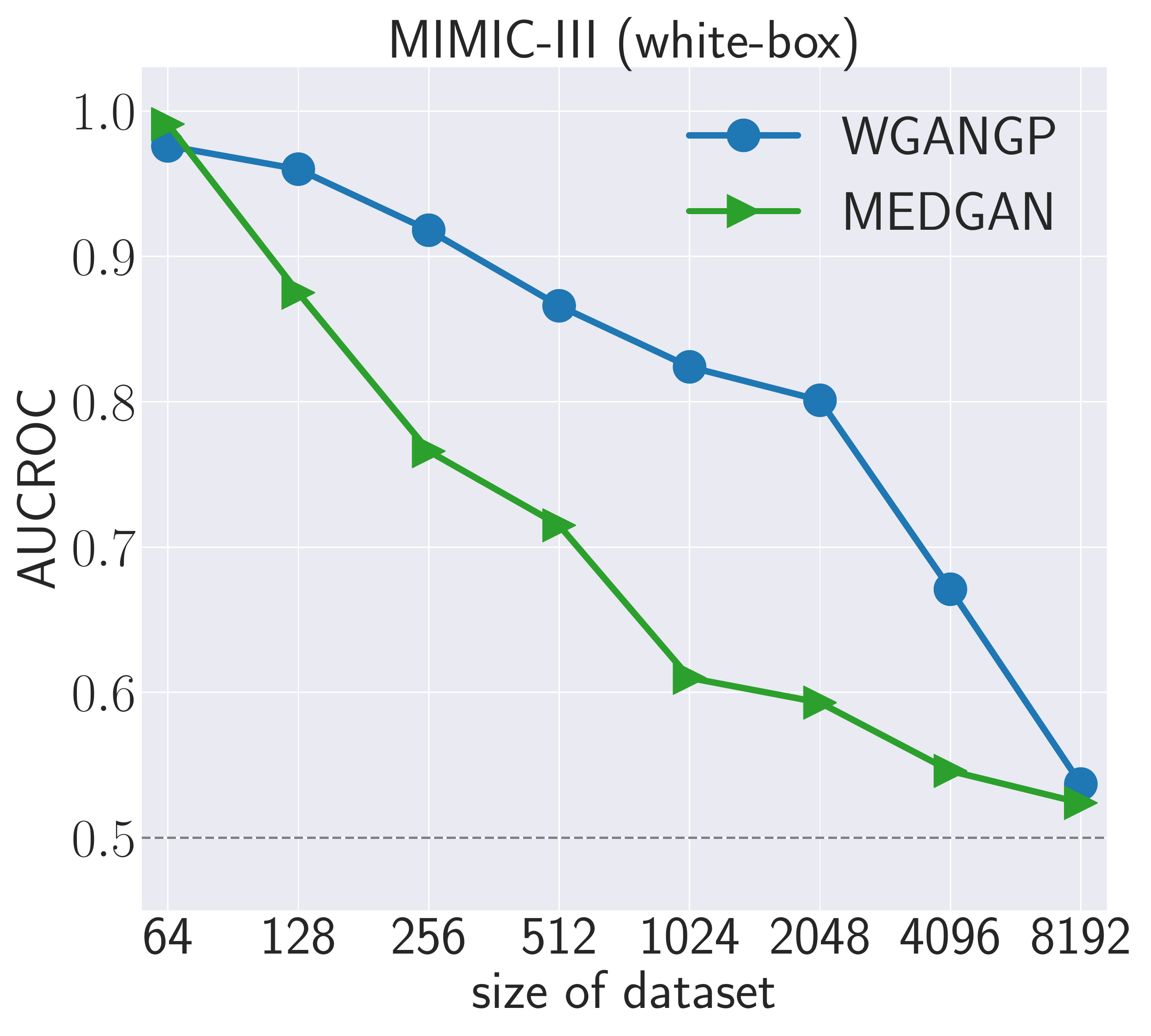}
\label{fig:wb_size_2}
}
\subfigure[]{
\includegraphics[width=0.55\columnwidth]{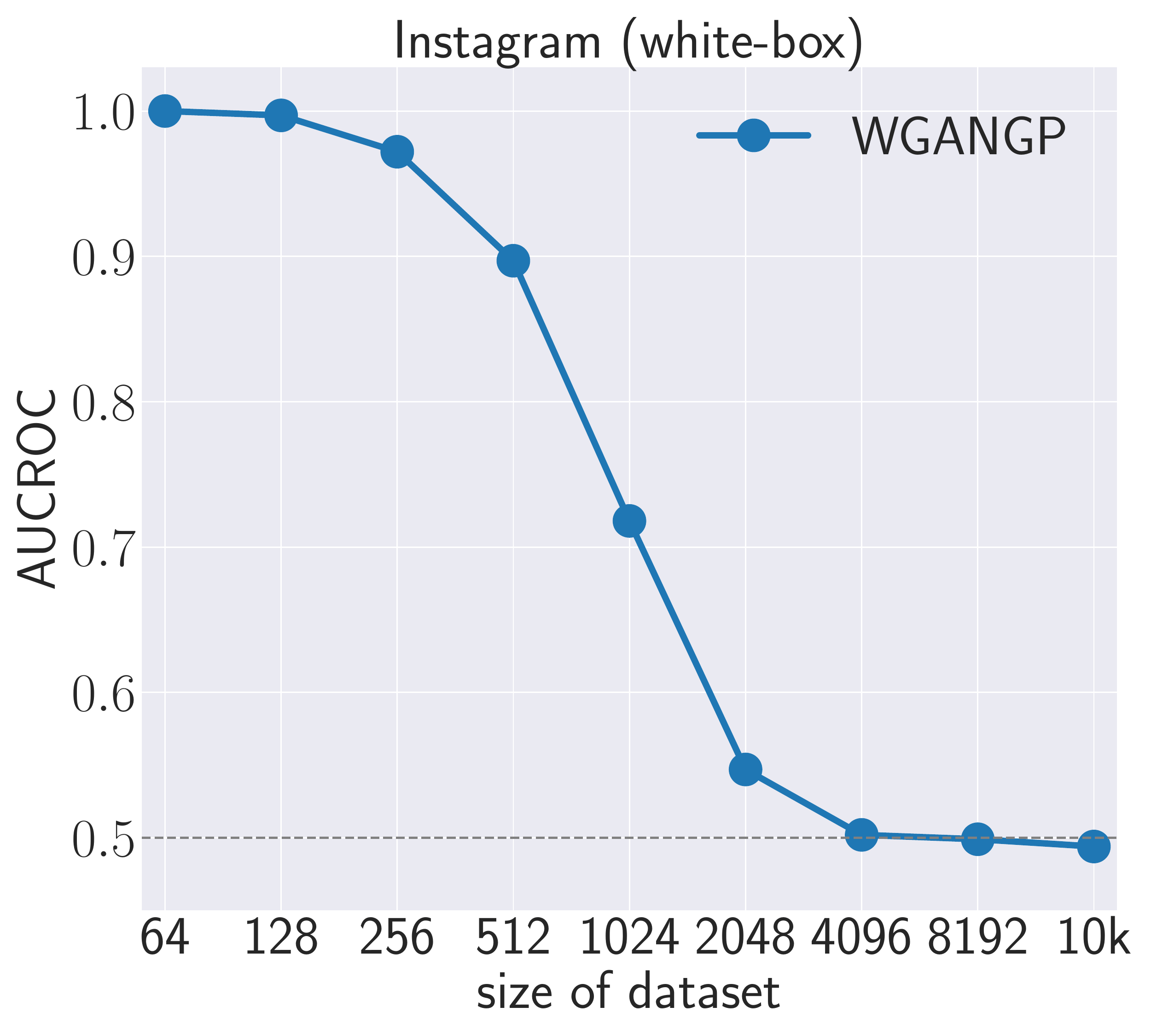}
\label{fig:wb_size_3}
}
\caption{Full black-box attack (the first row) and white-box attack (the second row) performance w.r.t. GAN training set size.
}
\end{figure*}

\subsubsection{Random v.s.\ Identity-based Selection for GAN Training Set} There are different levels of difficulty for membership inference attack. For example, CelebA contains person identity information and we can design attack difficulty by composing GAN training set based on identity or not. In one case, we include all images of the selected individuals for training(\emph{identity}). In the other case, we ignore identity information and randomly select images for training(\emph{random}), i.e., it is possible that some images for an individual are in the training dataset while some are not. The former case is relatively easier to attackers with a larger margin between membership image set and non-membership image set. In line with previous work~\cite{HMDC19}, we evaluate these two kinds of training set selection schemes on CelebA for a complete and fair comparison.

\begin{figure*}[!t]
\centering
\subfigure[]{
\includegraphics[width=0.55\columnwidth]{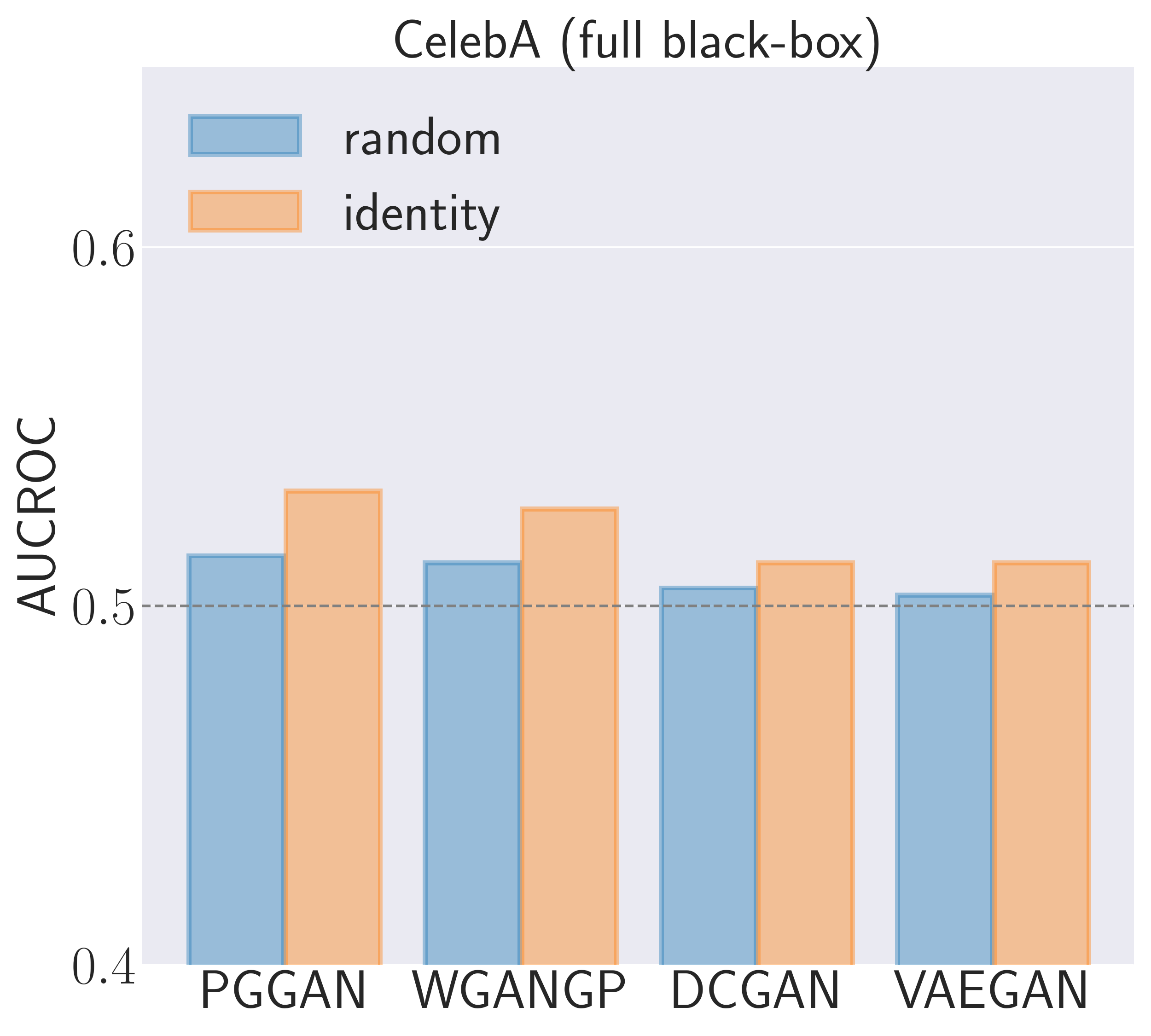}
\label{fig:celeba_iden_rand_bb}
}
\subfigure[]{
\includegraphics[width=0.55\columnwidth]{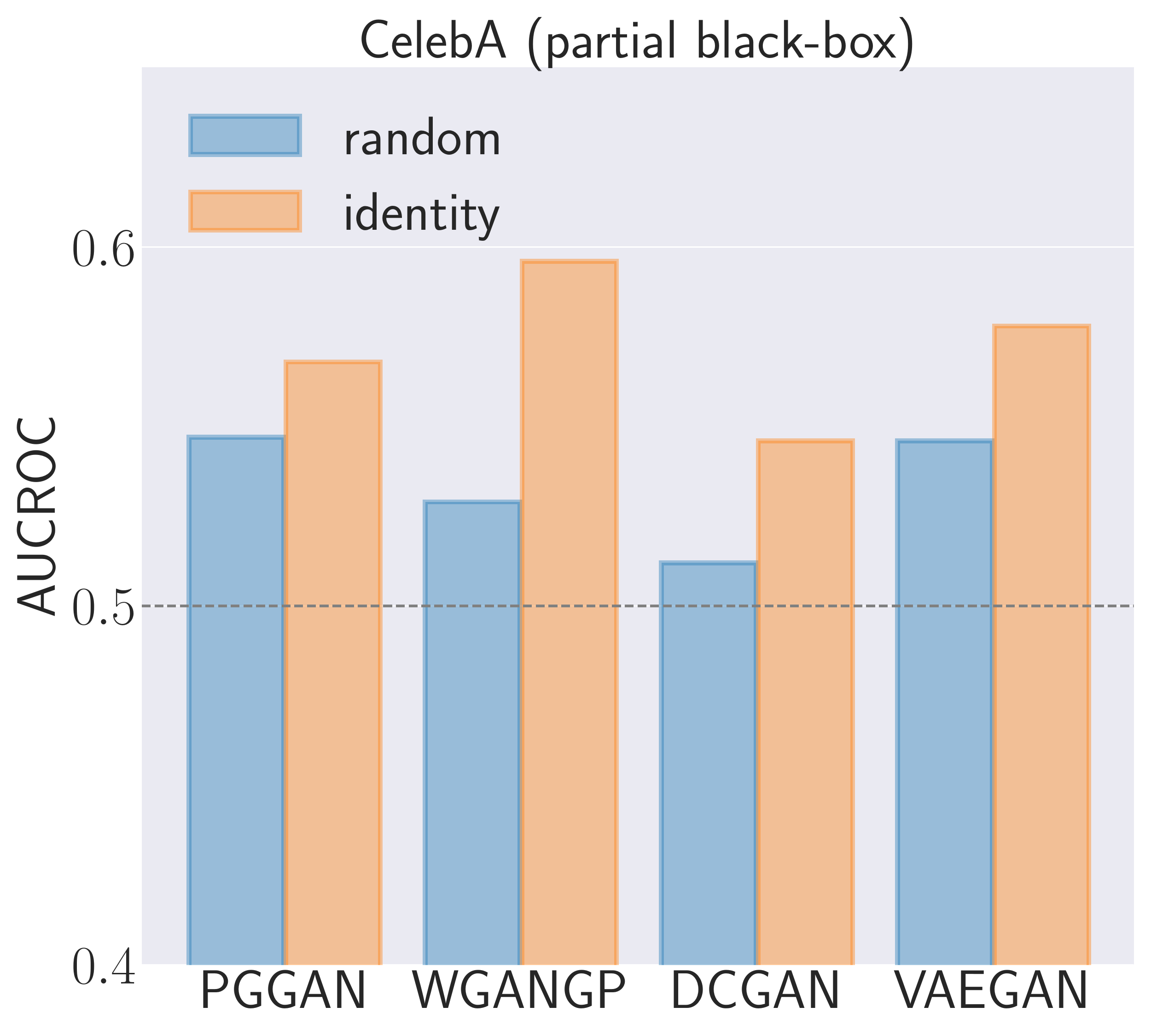}
\label{fig:celeba_iden_rand_pbb}
}
\subfigure[]{
\includegraphics[width=0.55\columnwidth]{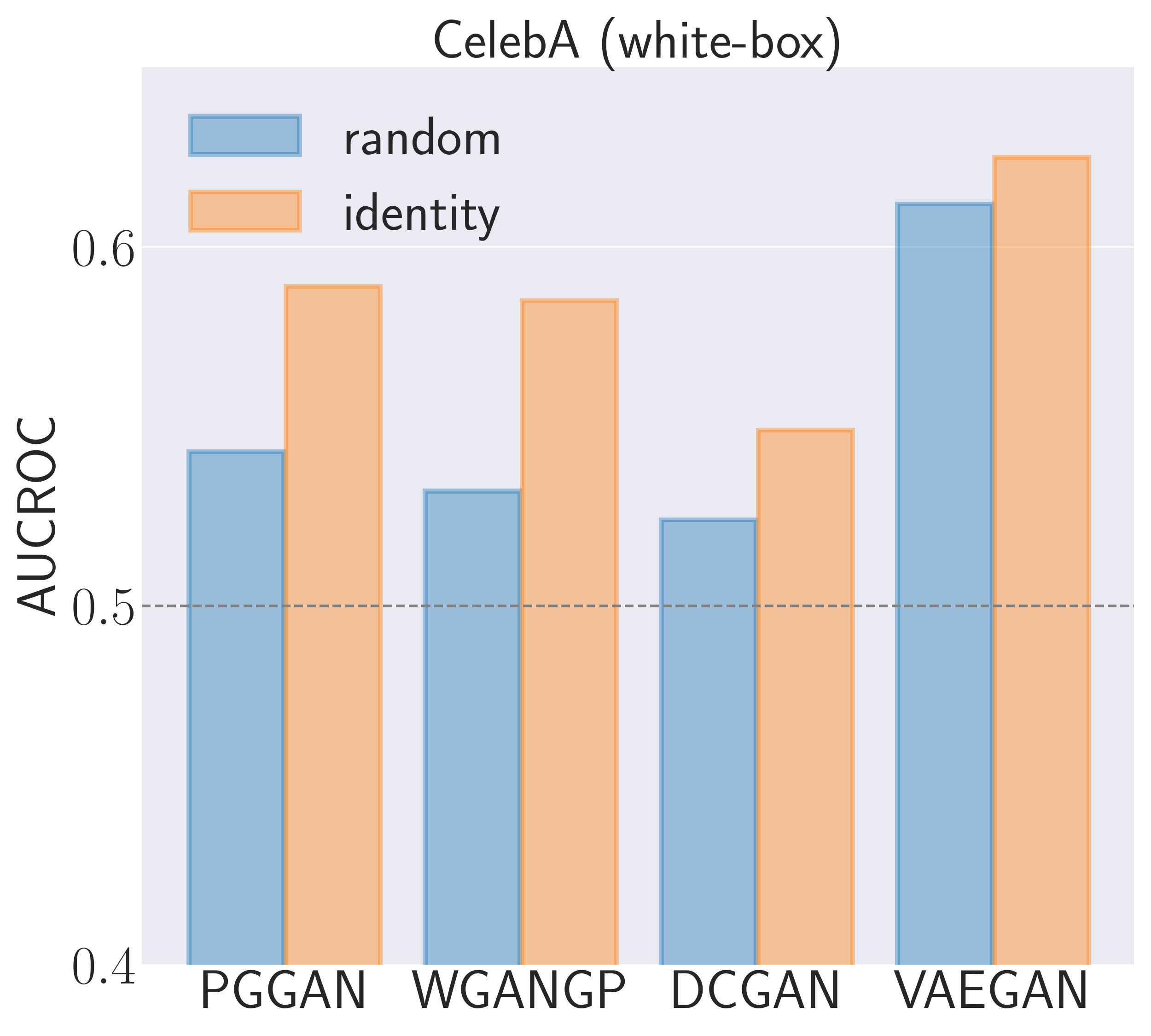}
\label{fig:celeba_iden_rand_wb}
}
\caption{Attack performance on the random v.s. identity-based training set selection (CelebA with size=20k).
}
\label{fig:celeba_iden_rand}
\end{figure*}

\subsection{Evaluation on Full Black-box Attack}
\label{sec:exp_fbb}

We start with evaluating our preliminary low-skill black-box attack model in order to gain a sense of the difficulty of the whole problem. 

\subsubsection{Performance w.r.t. GAN Training Set Size} \autoref{fig:bb_size_1} to \autoref{fig:bb_size_3} plot the attack performance against different GAN models on the three datasets. As shown in the plots, the attack performs sufficiently well when the training set is small for all three datasets. For instance, on CelebA, when the training set contains up to 512 images, attacker's AUCROC on both PGGAN and WGANGP are above 0.95. This indicates an almost perfect attack and a serious privacy breach. For larger training sets, however, the attacks become less effective as the degree of overfitting decreases and GAN's capability shifts from memorization to generalization. It is also consistent with the objective of GAN, i.e., to model the underlying distribution of the whole population instead of fitting a particular data sample. Hence, the collection of more data for GAN training can reduce privacy breach of individual samples. Moreover, PGGAN becomes more vulnerable than WGANGP on CelebA when the training size becomes larger. WGANGP is consistently more vulnerable than MEDGAN on MIMIC-\uppercase\expandafter{\romannumeral3} regardless of training size.

\subsubsection{Performance w.r.t. GAN Training Set Selection} \autoref{fig:celeba_iden_rand_bb} shows the attack performance w.r.t. training set selection schemes on four victim GAN models when fixing the training set size. We observe that, consistently, all the GAN models are more vulnerable when the training set is selected based on identity. Hence, more attention needs to be paid to an identity-based privacy breach, which is more likely to happen than an instance-based privacy breach. Moreover, when compared among different victim GAN models, DCGAN and VAEGAN are more resistant against the full black-box attack with AUCROC only marginally above 0.5 (random guess baseline). This may be attributed to the poor generation quality of DCGAN and VAEGAN (\autoref{table:gan_eval}), as it indicates that a certain amount of data samples can not be well represented by the victim model and thus the reconstruction error will be a less accurate approximation of the true membership probability in \autoref{eq:bayes error}.

\subsection{Evaluation on Partial Black-box Attack}
\label{sec:exp_pbb}

\subsubsection{Performance w.r.t. GAN Training Set Selection} \autoref{fig:celeba_iden_rand_pbb} shows the comparison on four victim GAN models. Similar to the case of the full black-box attack (\autoref{sec:exp_fbb}), we find that all models become more vulnerable to identity-based selection. Still, DCGAN is the most resistant victim against membership inference in both training set selection schemes, probably due to its inferior generation quality.

\subsubsection{Comparison to Full Black-box Attack} Comparing between \autoref{fig:celeba_iden_rand_bb} and \autoref{fig:celeba_iden_rand_pbb}, the attack performance against each GAN model consistently and significantly improves from black-box setting to partial black-box setting. We attribute this improvement to a better reconstruction of query samples found by the attacker via optimization. Hence, we conclude that providing the input interface to a generator suffers from an increased privacy risk.

\subsection{Evaluation on White-box Attack}
\label{sec:exp_wb}

We further investigate the case where the victim generator is published in a white-box manner. This scenario is commonly studied in the field of privacy preserving data generation~\cite{beaulieu2019privacy,AMCC17,ZJW18,XLWWZ18,JYS19,COF20}, where our approach can serve as a simple and interpretable framework for empirically quantifying the privacy leakage.
As the optimization in the white-box attack involves more technical details, we conduct additional analysis study and sanity check in this setting. See Appendix~\ref{sec:wb sanity check} for more details.

\begin{figure*}[!t]
\centering
\subfigure[]{
\label{fig:celeba_calibrate_id(1)}
\includegraphics[width=0.55\columnwidth]{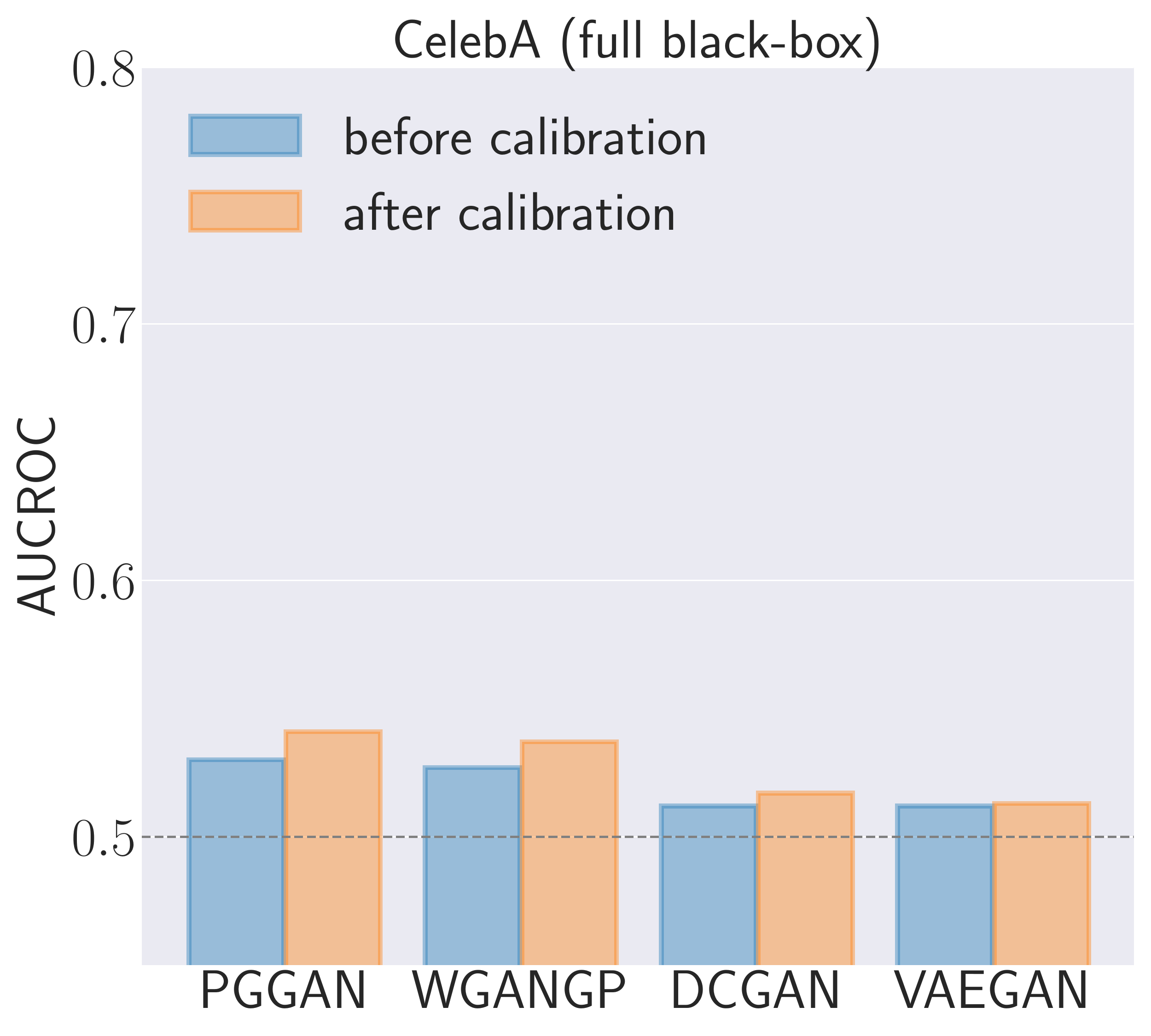}
}
\subfigure[]{
\label{fig:celeba_calibrate_id(2)}
\includegraphics[width=0.55\columnwidth]{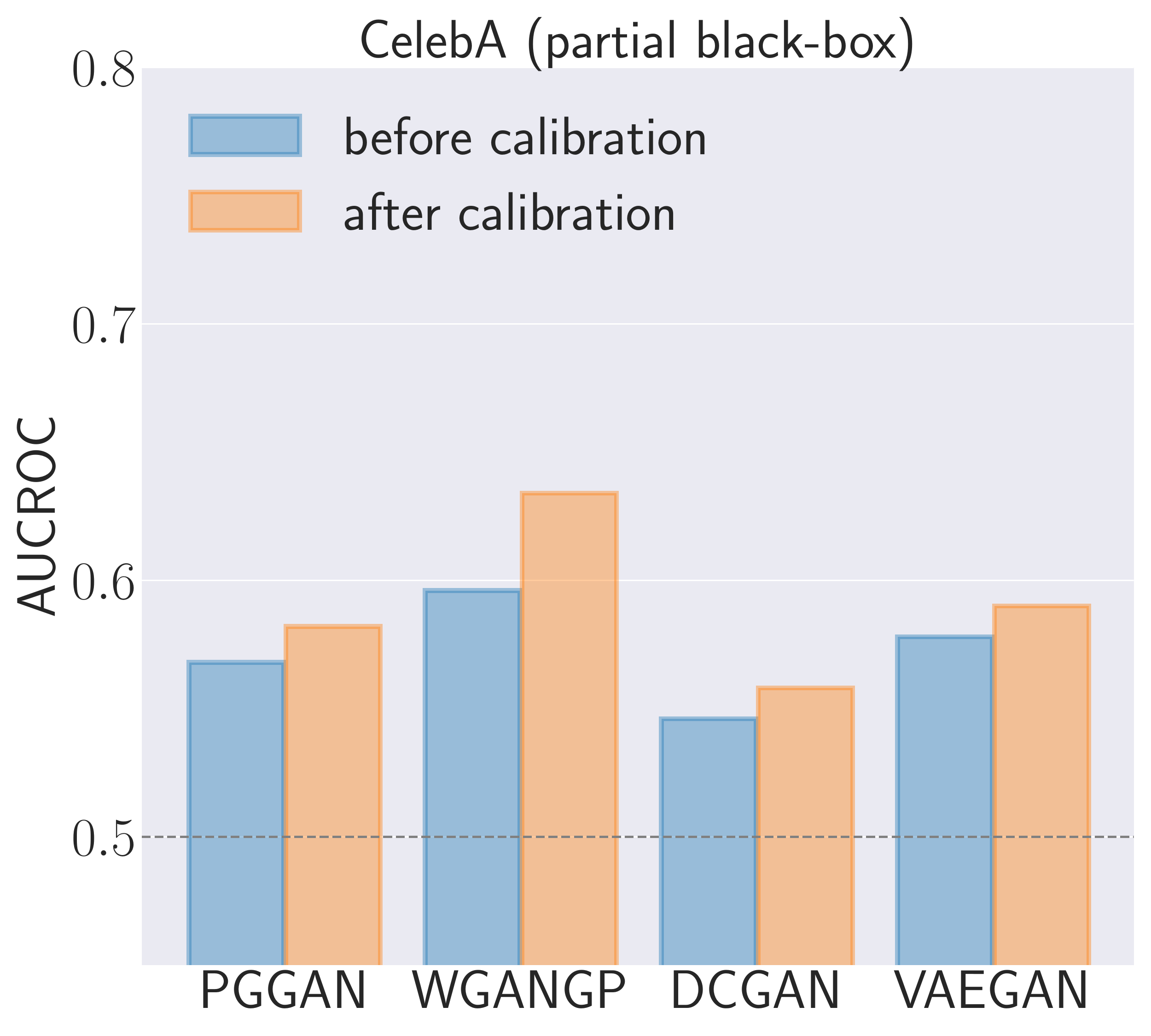}
}
\subfigure[]{
\label{fig:celeba_calibrate_id(3)}
\includegraphics[width=0.55\columnwidth]{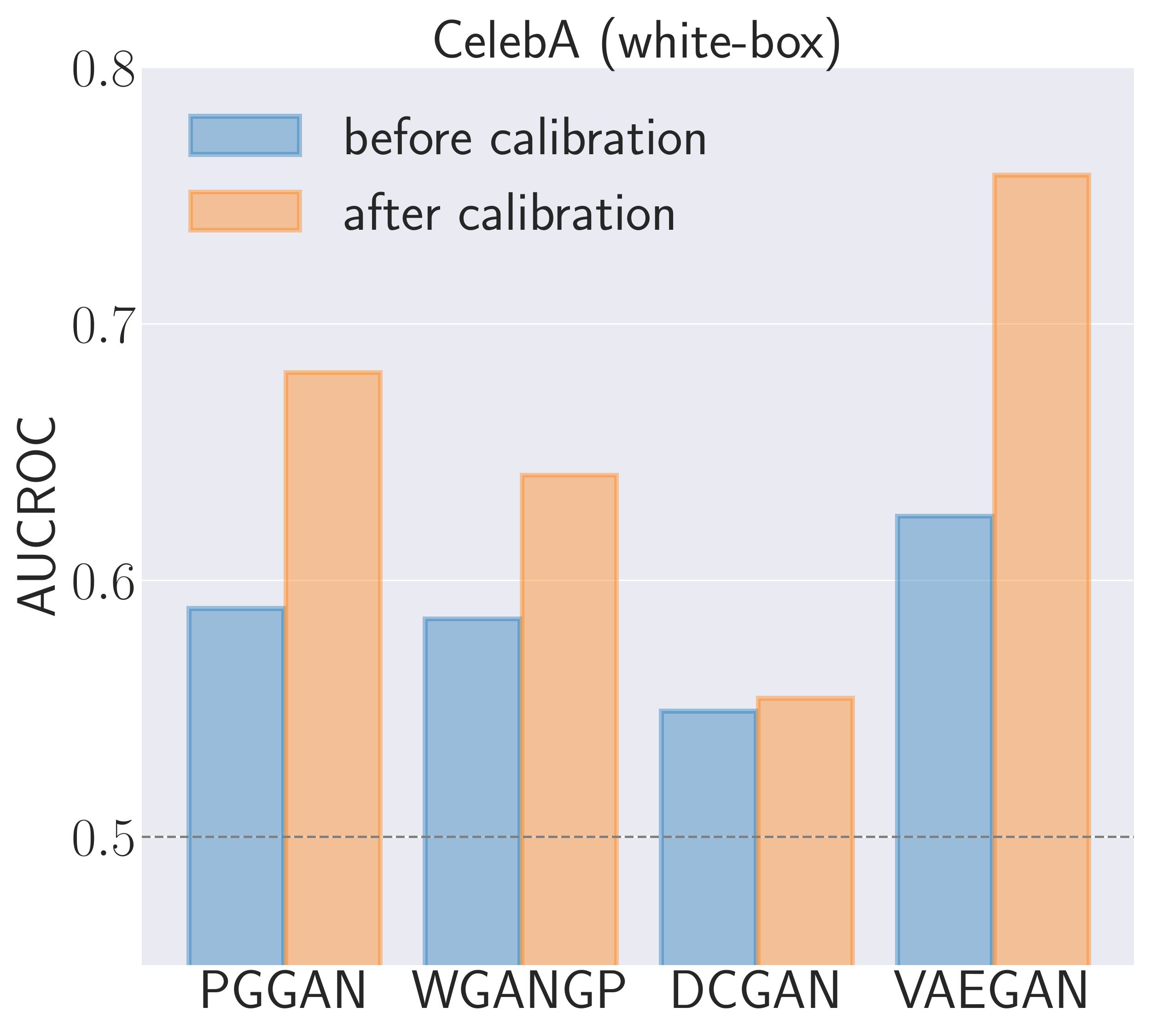}
}
\caption{Attack performance before and after calibration on CelebA (size=20k). }
\label{fig:celeba_calibrate_iden}
\end{figure*}

\begin{figure*}[!t]
\centering
\subfigure[]{
\includegraphics[width=0.55\columnwidth]{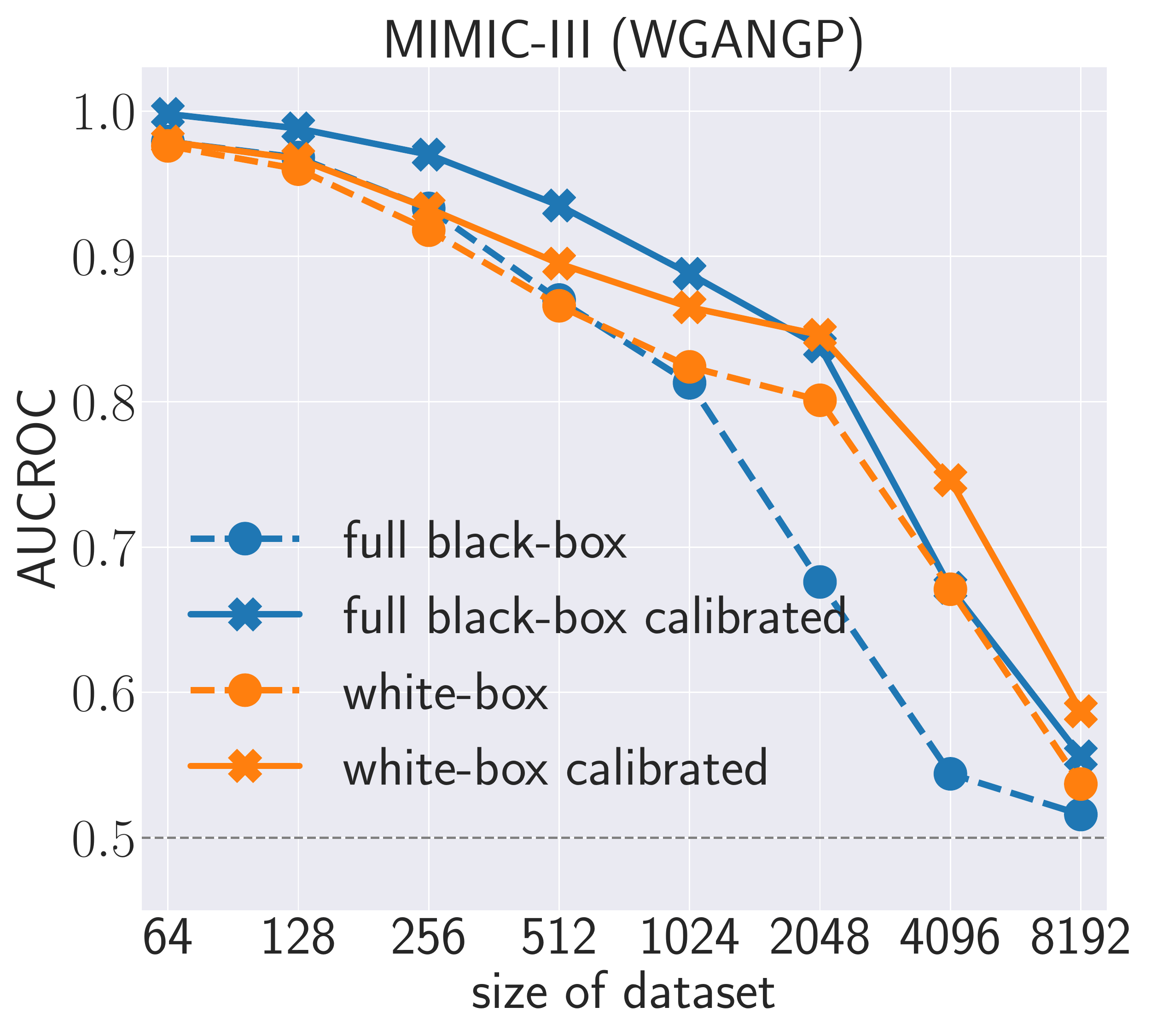}
}
\subfigure[]{
\includegraphics[width=0.55\columnwidth]{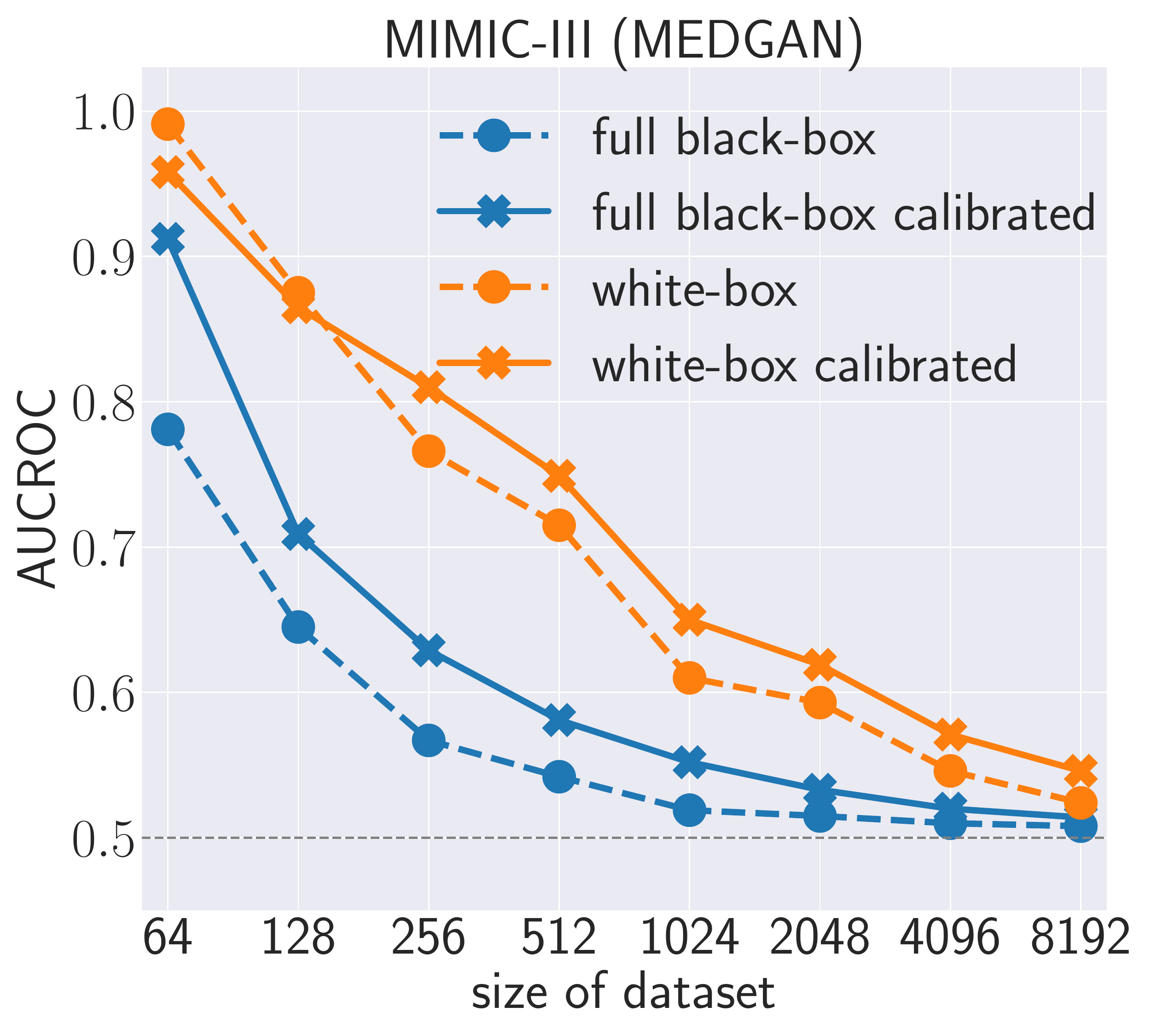}
}
\subfigure[]{
\includegraphics[width=0.55\columnwidth]{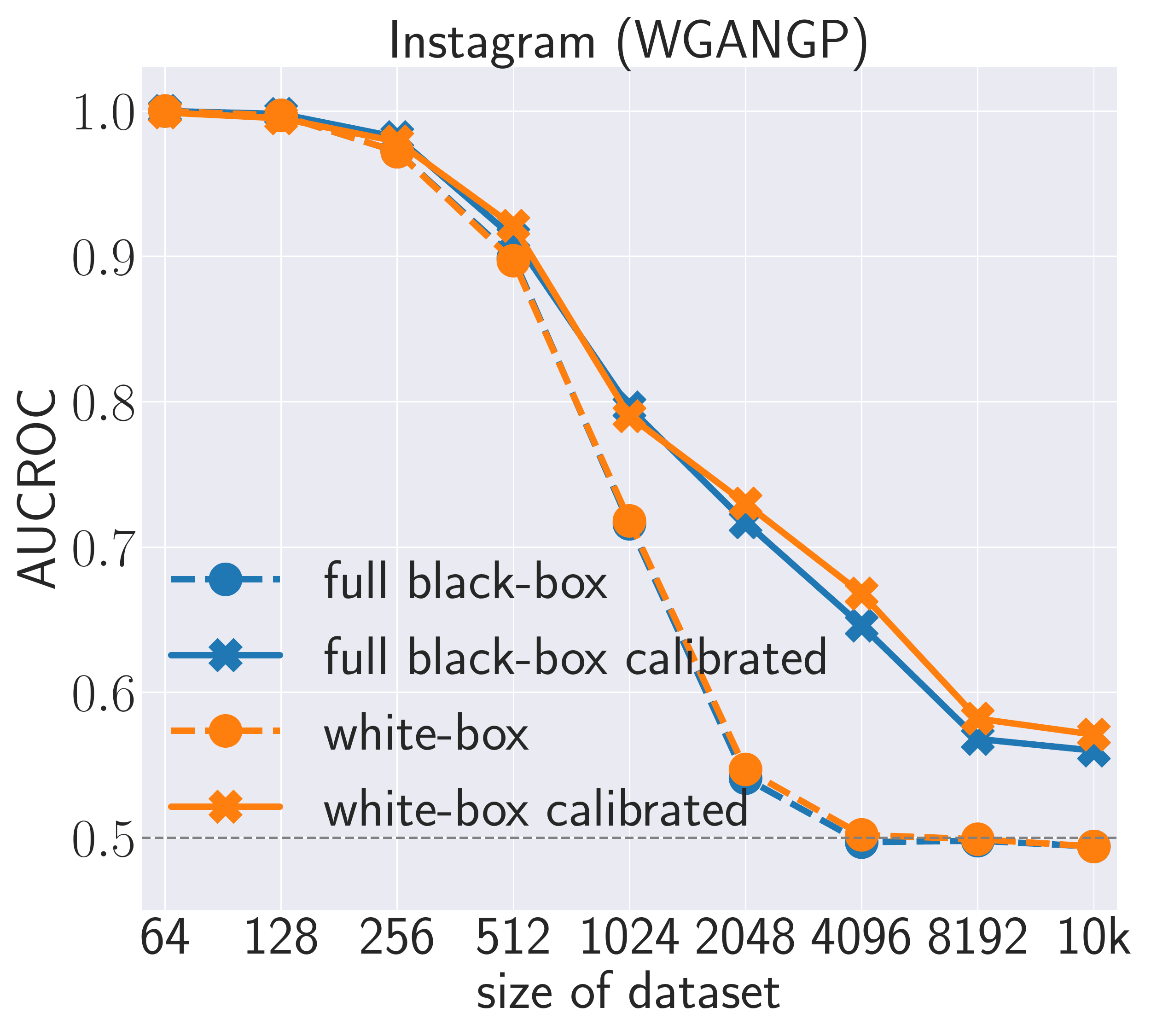}
}
\caption{Attack performance before and after calibration for non-image datasets w.r.t. GAN training set sizes.}
\label{fig:wbbb_calibrate}
\end{figure*}

\subsubsection{Performance w.r.t. GAN Training Set Size} \autoref{fig:wb_size_1} to \autoref{fig:wb_size_3} plot the attack performance against different GAN models on the three datasets when varying training set size. We find that the attack becomes less effective as the training set becomes larger, similar to that in the black-box setting. For CelebA, the attack remains effective for 20k training samples, while for MIMIC-\uppercase\expandafter{\romannumeral3} and Instagram, this number decreases to 8192 and 2048, respectively. The strong similarity between the member and non-member in these two non-image datasets increases the difficulty of attack, which explains the deteriorated effectiveness of the attack model.

\subsubsection{Performance w.r.t. GAN Training Set Selection} \autoref{fig:celeba_iden_rand_wb} shows the comparisons against four victim GAN models. Our attack is much more effective when composing GAN training set according to identity, which is similar to those in the full and partial black-box settings. 

\subsubsection{Comparison to Full and Partial Black-box Attacks} For membership inference attack, it is an important question whether or to what extent the white-box attack is more effective than the black-box ones. For discriminative (classification) models, recent literature reports that the state-of-the-art black-box attack performs almost as well as the white-box attack~\cite{SDSOJ19,NSH19}. In contrast, we find that against generative models the white-box attack is much more effective. Comparisons across subfigures in \autoref{fig:celeba_iden_rand} show that the AUCROC values increase by at least 0.03 when changing from full black-box to white-box setting. Compared to the partial black-box attack, the white-box attack achieves noticeably better performance against PGGAN and VAEGAN. Moreover, conducting the white-box attack requires much less computation cost than conducting the partial black-box attack. Therefore, we conclude that publicizing model parameters (white-box setting) does incur high privacy breach risk.

\subsection{Performance Gain from Attack Calibration}
\label{sec:exp_calibration}

We perform calibration on all the settings. Note that for full and partial black-box settings, attackers do not have prior knowledge of victim model architectures. We thus train a PGGAN on LFW face dataset~\cite{huang2008labeled} and use it as the generic reference model for calibrating all victim models trained on CelebA in the black-box settings. Similarly, for MIMIC-\uppercase\expandafter{\romannumeral3}, we use WGANGP as the reference model for MedGAN and vice versa. In other words, we have to guarantee that our calibrated attacks strictly follow the black-box assumption.

\autoref{fig:celeba_calibrate_iden} compares attack performance on CelebA before and after applying calibration. The AUCROC values are improved consistently across all the GAN architectures in all the settings. In general, the white-box attack calibration yields the greatest performance gain. Moreover, the improvement is especially significant when attacking against VAEGAN, as the AUCROC value increases by 0.2 after applying calibration.       

\autoref{fig:wbbb_calibrate} compares attack performance on the other two non-image datasets. The performance is also consistently boosted for all training set sizes after calibration.

\begin{figure}[!t]
\centering
\includegraphics[width=0.77\columnwidth]{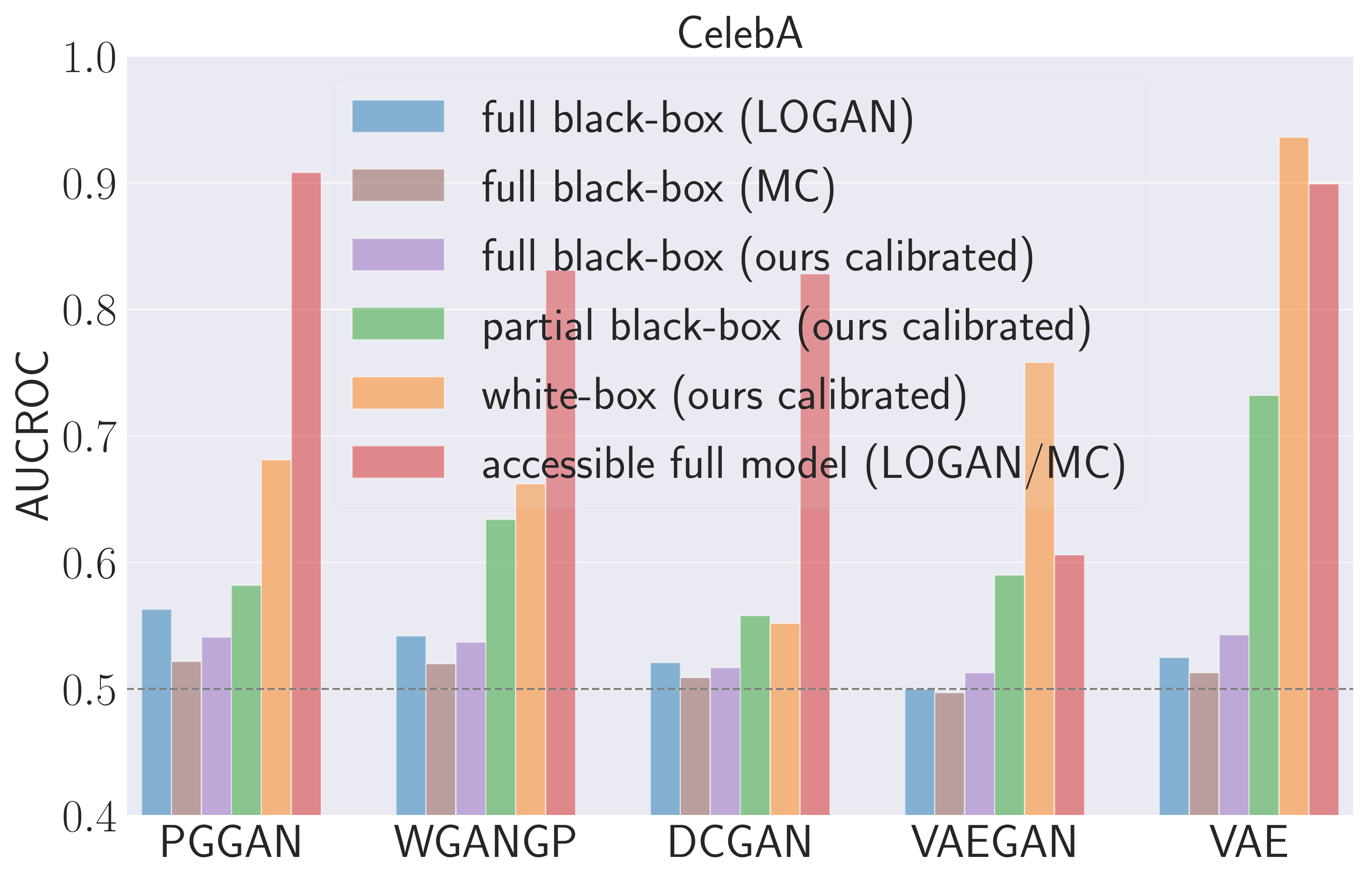}
\caption{Comparison of different attacks on CelebA. See \autoref{table:celeba_allattack} in Appendix for quantitative results.}
\label{fig:celeba_iden_allattack}
\end{figure}

\begin{figure}[!t]
\centering
\includegraphics[width=0.8\columnwidth]{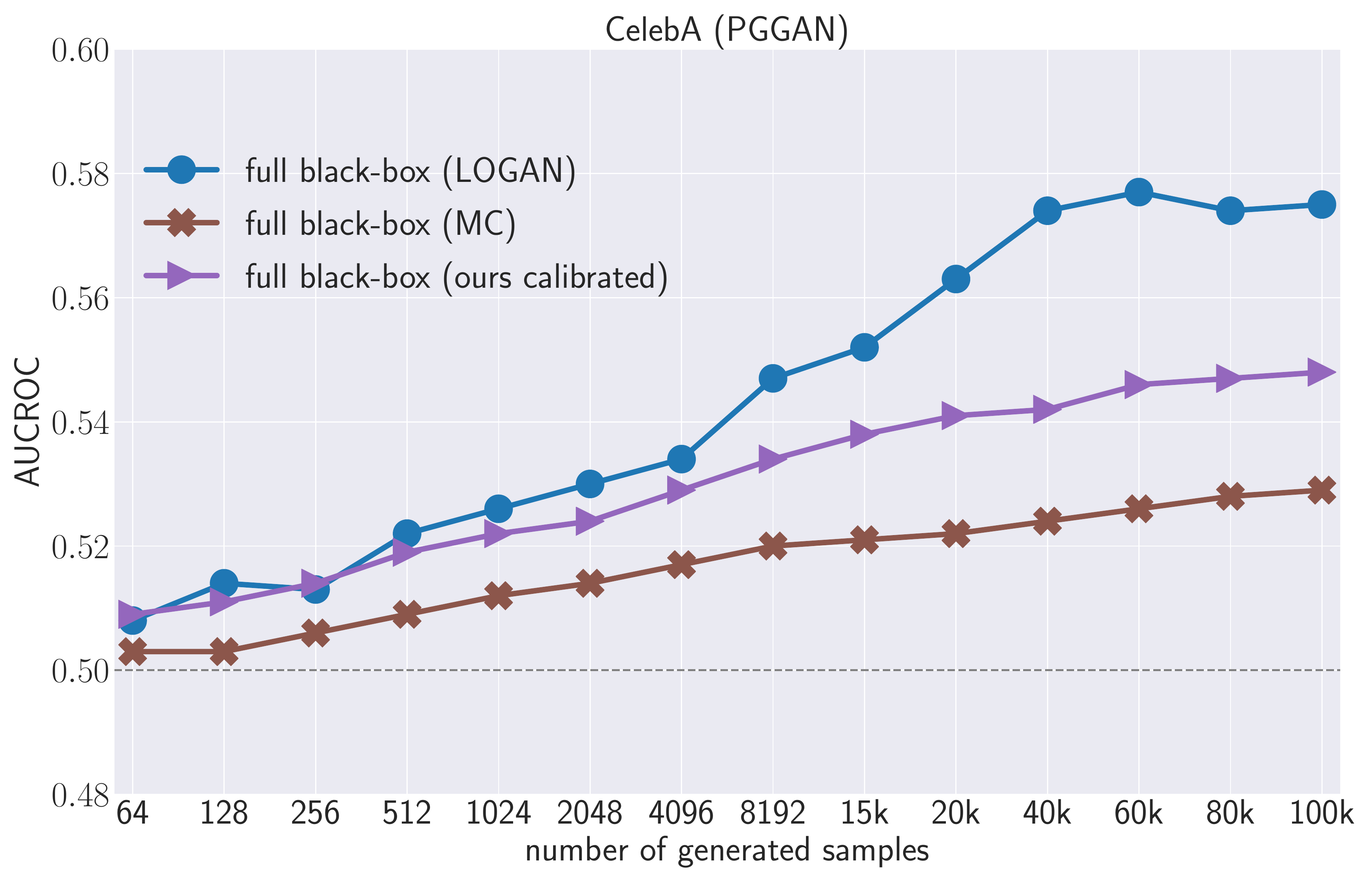}
\caption{Full black-box attack performance against PGGAN on CelebA w.r.t. $k$ in \autoref{eq:fbb_rec}, the number of generated samples. See \autoref{table:fbb_numqueries} in Appendix for quantitative results.}
\label{fig:fbb_numqueries}
\end{figure}

\subsection{Comparison to Baseline Attacks}
\label{sec:comparisons}

We compare our calibrated attack to two recent membership inference attack baselines: Hayes~et~al.~\cite{HMDC19} (denoted as \textbf{LOGAN}) and Hilprecht~et~al.~\cite{HHB19} (denoted as \textbf{MC}, standing for their proposed Monte Carlo sampling method). As described in our taxonomy (\autoref{sec:taxonomy}), LOGAN includes a full black-box attack model and a discriminator-accessible attack model against GANs. The latter is regarded as the most knowledgeable but unrealistic setting because the discriminator in GAN is usually not accessible in practice. But we still compare to both settings for the completeness of our taxonomy and experiments. MC includes a full black-box attack against GANs and a full-model-accessible attack against VAEs. We evaluate our generic attack model on both GANs and VAEs for a complete comparison, though we mainly focus on GANs in this work. Note that, to the best of our knowledge, there does not exist any attack against GANs in the partial black-box or white-box settings.

\begin{figure*}[!t]
\centering
\subfigure[]{
\label{fig:mimic_instagram_allattack_a}
\includegraphics[width=0.57\columnwidth]{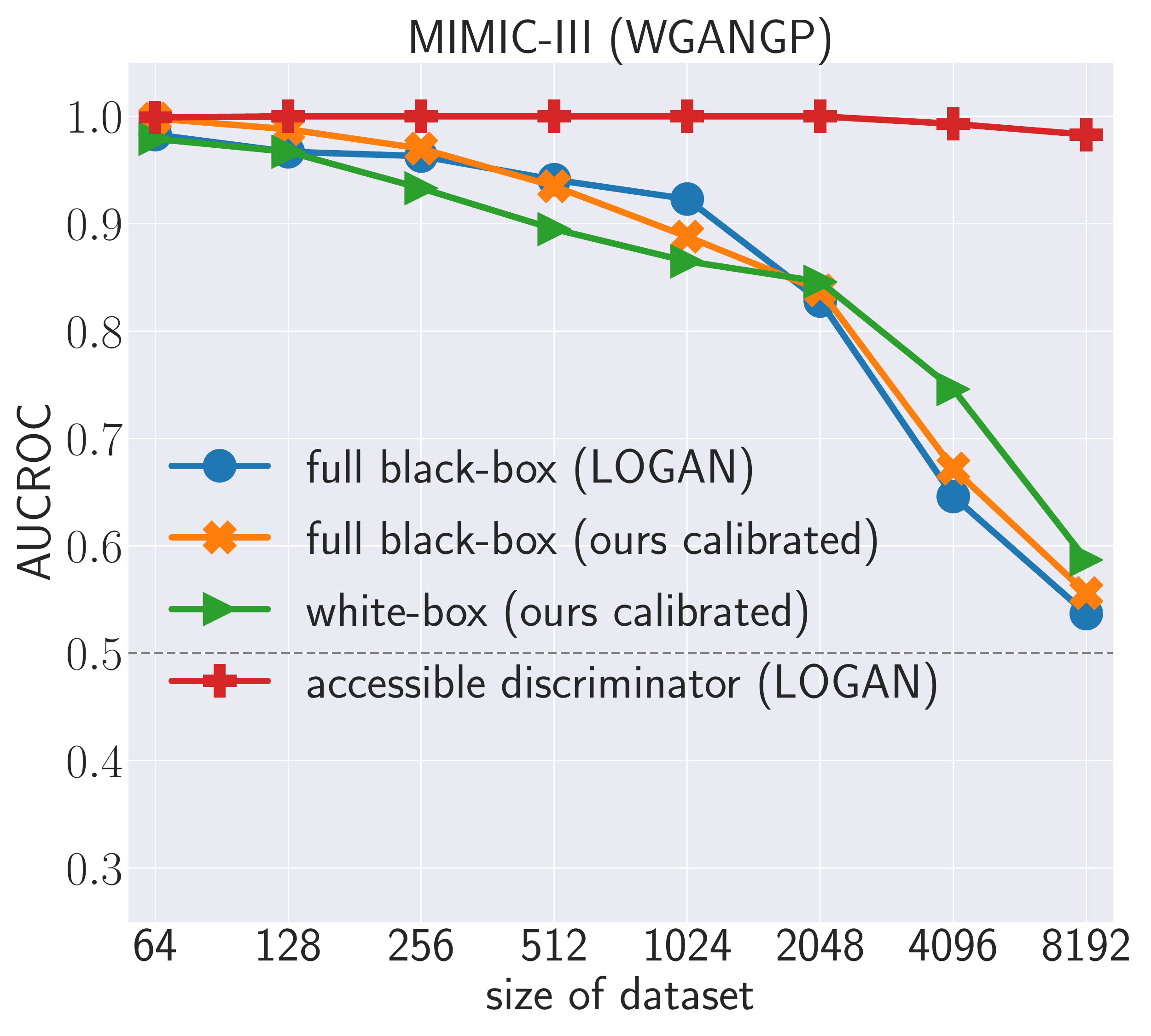}
}
\subfigure[]{
\label{fig:mimic_instagram_allattack_b}
\includegraphics[width=0.57\columnwidth]{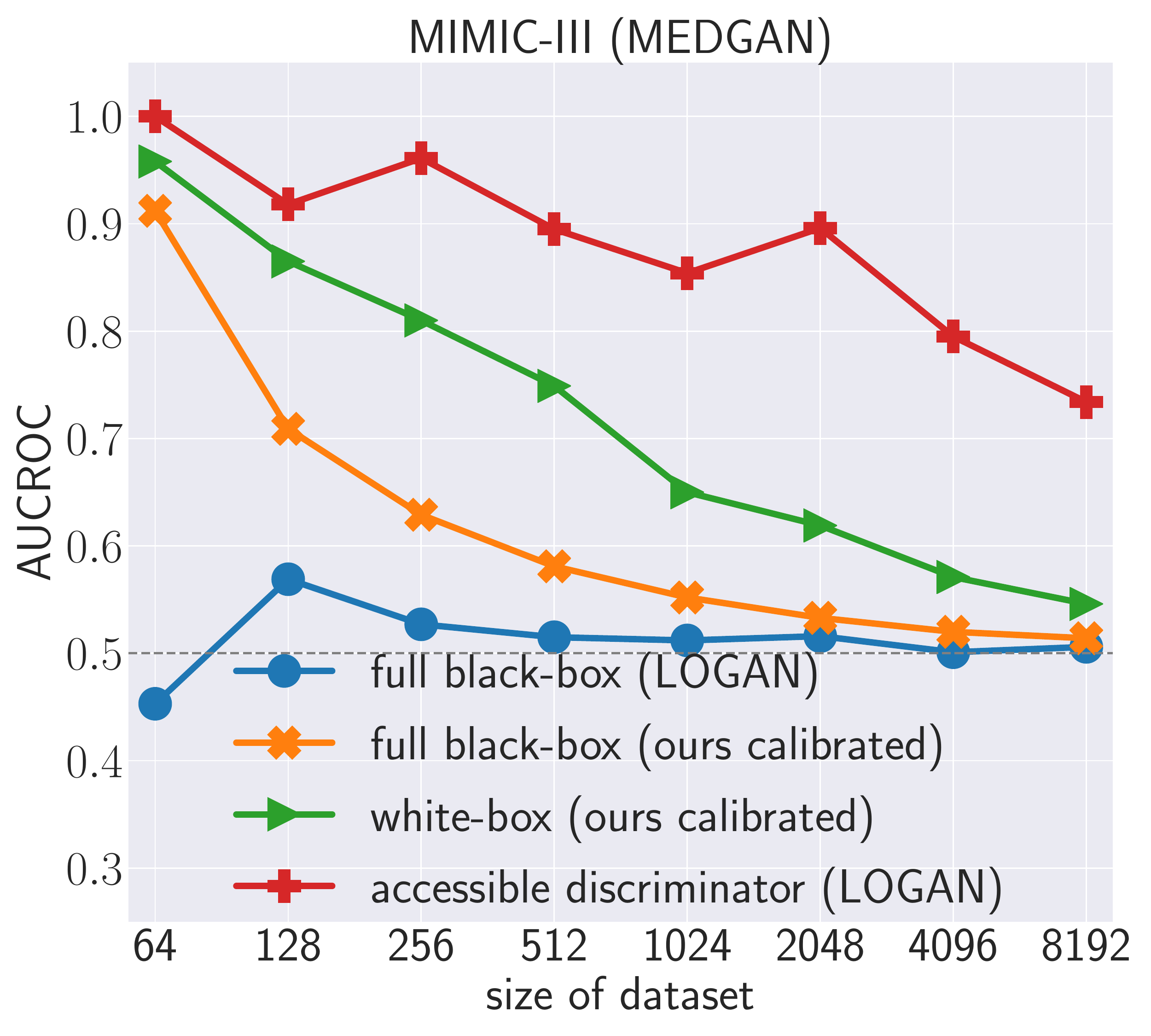}
}
\subfigure[]{
\label{fig:mimic_instagram_allattack_c}
\includegraphics[width=0.57\columnwidth]{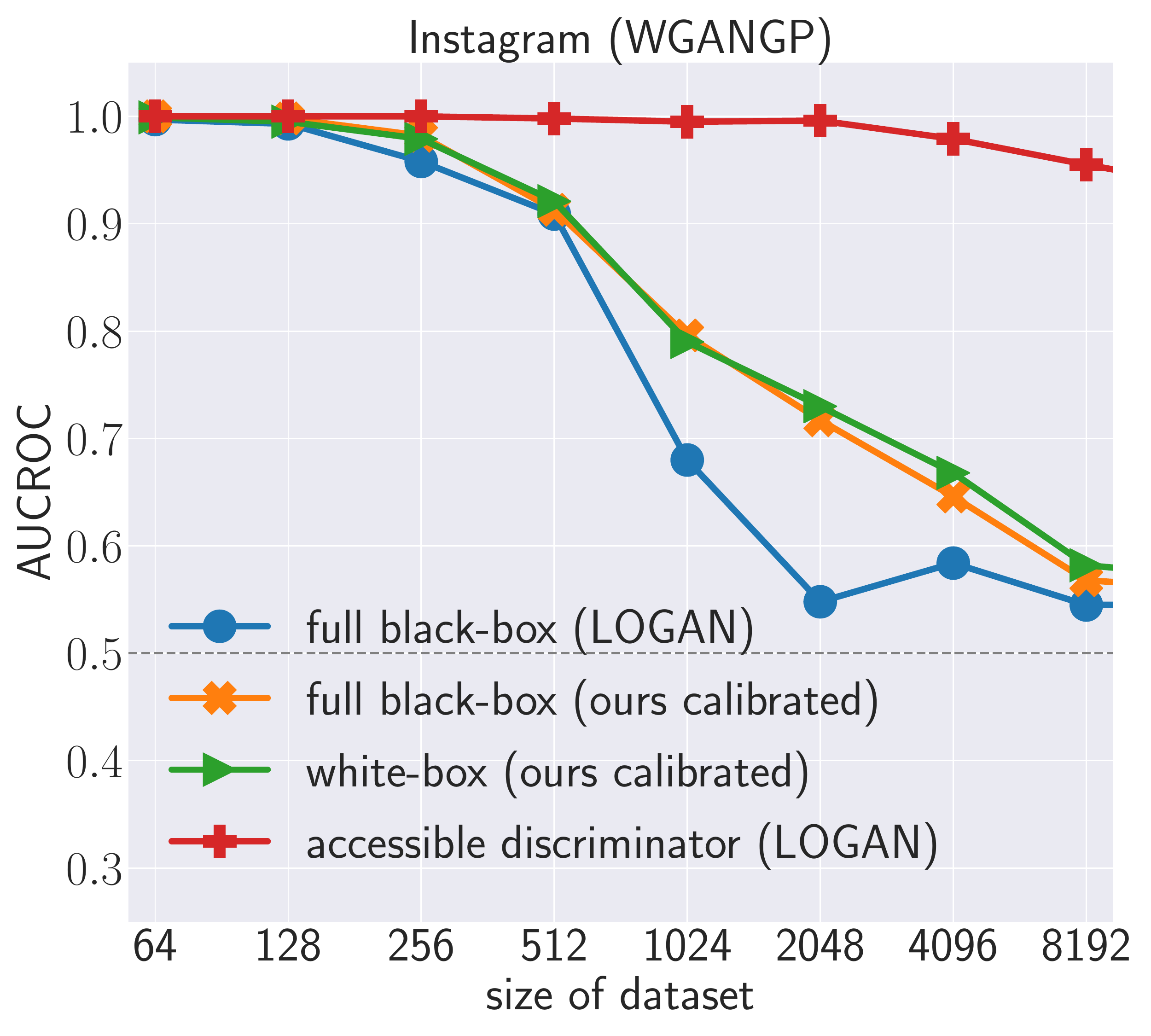}
}
\caption{Comparison of different attacks on the other two non-image datasets w.r.t. GAN training set size. See \autoref{table:mimic_instagram_all_attack} in Appendix for quantitative results.}
\label{fig:mimic_instagram_allattack}
\end{figure*}

\autoref{fig:celeba_iden_allattack}, \autoref{fig:fbb_numqueries}, and \autoref{fig:mimic_instagram_allattack} show the comparisons, considering several datasets, victim models, training set sizes, numbers of query images (full black-box), and different attack settings. We skip MC on the non-image datasets as it is not directly applicable in terms of their distance calculation. Our findings are as follows.

In the black-box setting, our low-skill attack consistently outperforms MC and outperforms LOGAN on the non-image datasets. It also achieves comparable performance to LOGAN on CelebA but with a much simpler and learning-free implementation. 

Our white-box and even partial black-box attacks consistently outperform the other full black-box attacks. Hence, publicizing the generator or even just the input to the generator can lead to a considerably higher risk of privacy breach. With a complete spectrum of performance across settings, they bridge the performance gap between the highly constrained full black-box attack and the unrealistic discriminator-accessible attack. Moreover, our proposed white-box attack model is of practical value for the differential privacy community. 

Assuming the accessibility of discriminator (full model) normally results in the most effective attack. This can be explained by the fact that the discriminator is explicitly trained to maximize the margin between training set (membership samples) and generated set (a subset of non-membership samples), which eventually yields very accurate confidence scores for membership inference. Surprisingly, our calibrated white-box attack even outperforms baseline methods in more knowledgeable settings, i.e., LOGAN (accessible discriminator) for VAEGAN and MC (accessible full model) for VAE. This shows that when data coverage is explicitly enforced, which probably leads to overfitting and data memorization if not properly regularized, our attack models are highly effectively and achieve superior performance with a more realistic assumption.

\begin{figure}[!t]
\centering
\subfigure[]{
\includegraphics[width=0.57\columnwidth]{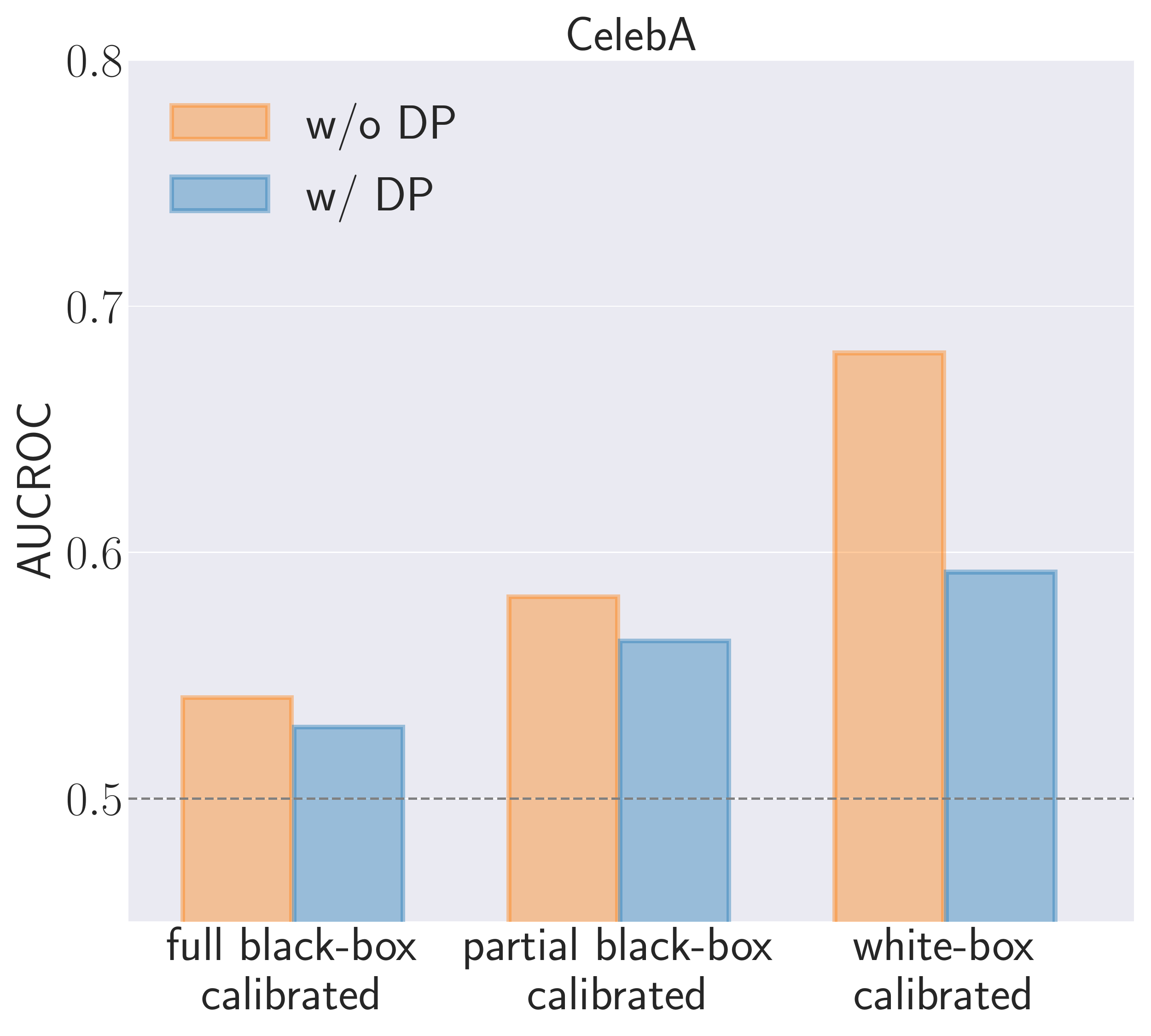}
\label{fig:celeba_defense_allattack}
}
\subfigure[]{
\includegraphics[width=0.57\columnwidth]{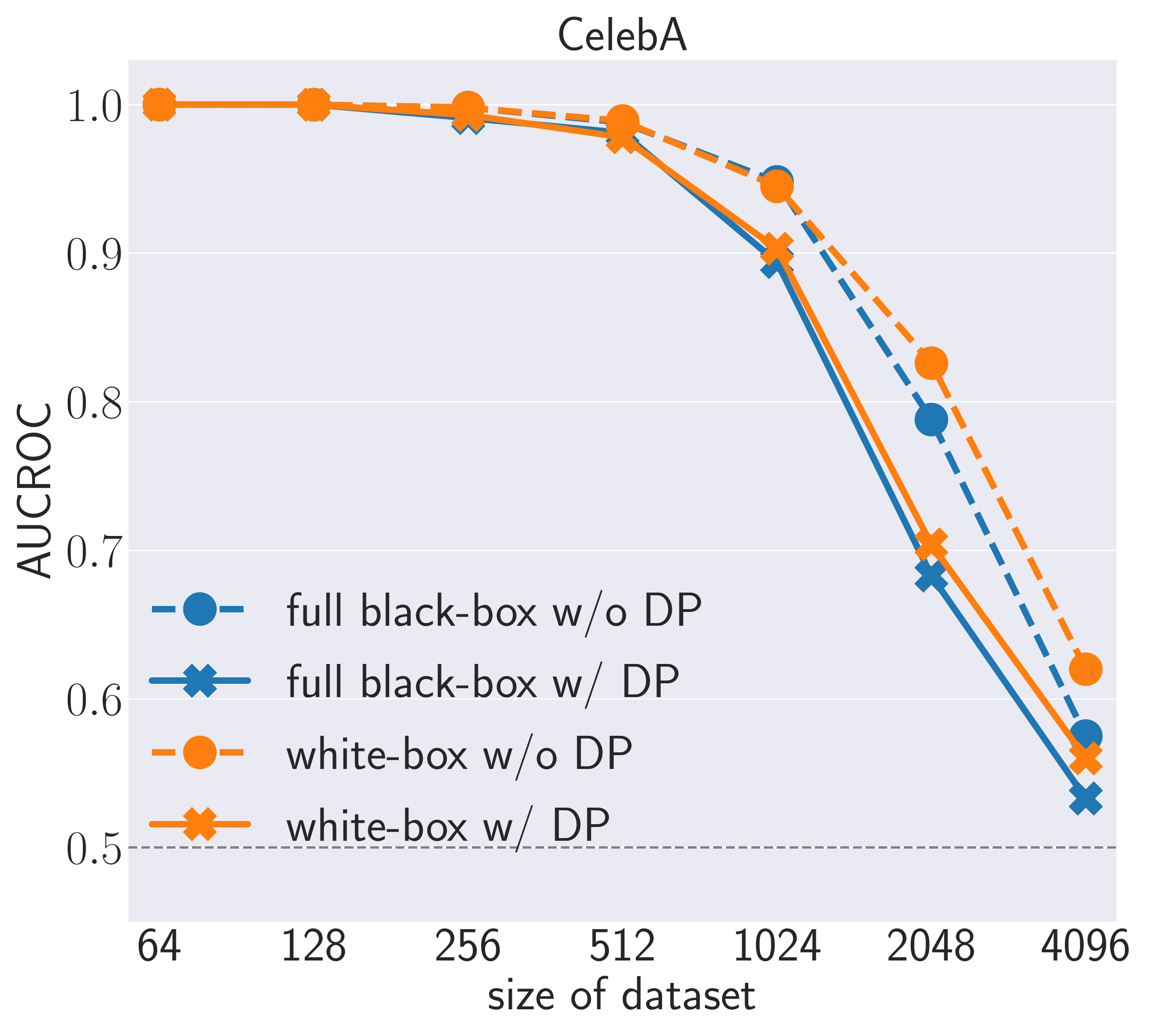}
\label{fig:defense_size}
}
\caption{
(a) Attack performance against PGGAN on CelebA with or without DP defense.
(b)Attack performance against PGGAN on CelebA with or without DP defense, w.r.t. GAN training set size. We fix all the other control factors (training iterations, batch size, noise scale, norm threshold) apart from training set size, which results in less privacy guarantee for a smaller dataset.
}
\end{figure}

\subsection{Defense}
\label{sec:dp}

We investigate the most effective defense mechanism against MIA to date that is applicable to GANs~\cite{HMDC19,beaulieu2019privacy,ZJW18,XLWWZ18}, i.e., the differential private (DP) stochastic gradient descent \cite{ACGMMTZ16}. The algorithm can be summarized into two steps. First, the per-sample gradient computed at each training iteration is clipped by its $L_2$ norm with a pre-defined threshold. Subsequently, calibrated random noise is added to the gradient in order to inject stochasticity for protecting privacy. %
In this scheme, however, privacy protection is at the cost of computational complexity and utility deterioration, i.e., slower training and lower generation quality.

We conduct attacks against PGGAN on CelebA, which has been defended by DP. We skip the other cases because DP always deteriorates generation quality to an unacceptable level. The hyper-parameters are selected through the grid search. We fix the norm threshold to 1.0 (average gradient norm magnitude during pre-training) and the noise scale to $10^{-4}$ (the largest value with which we obtain samples of good visual quality). However, this results in high $\epsilon$ values ($>10^{10}$ for a default value of $\delta=10^{-5}$), while it still reduces the effectiveness of the membership inference attack.

\autoref{fig:celeba_defense_allattack} and \autoref{fig:defense_size} depict the attack performance in different settings. We observe a consistent decrease in AUCROC in all the settings. Therefore, DP is effective in general against our attack. However, applying DP into training leads to a much higher computation cost ($10\times$ slower) in practice due to the per-sample gradient modification. Moreover, DP results in a deterioration of GAN utility, which is witnessed by an increasing FID (comparing the last and second columns in \autoref{table:gan_eval}). Moreover, for obtaining a pleasing level of utility, the noise scale has to be limited to a small value, which, in turn, cannot defend the membership inference attack completely. For example, for all the settings, our attack still achieves better performance than the random guess baseline (AUCROC $= 0.5$).

\subsection{Summary}
\label{sec:summary}

Before ending this section, we show a few insights over the experiment results and list practical considerations relevant to the deployment of GANs and potential privacy breaches.

\begin{itemize}
\item The vulnerability of models under MIA heavily relies on the attackers' knowledge about victim models. Releasing the discriminator (full model) results in an exceptionally high risk of privacy breach, which can be explained by the fact that the discriminator had full access to the training data and thus easily memorizes the private information about the training data. Similarly, the release of the generator and/or the control over the input noise $z$ also incurs a relatively high privacy risk.

\item The vulnerability of different generative models under MIA varies. Although the effectiveness of MIA mainly depends on the generation quality of victim models, the objective function and training paradigm also play important roles. Specifically, when data reconstruction is explicitly formulated in the training objective to improve data mode coverage, e.g. in VAEGAN and VAE, the resulting models become highly vulnerable to MIA.

\item A smaller training dataset leads to a higher risk of revealing information of individual samples. In particular, if the magnitude of training set size is less than $10k$ where most existing GAN models have sufficient modeling capacity for overfitting to individual sample, the membership privacy is highly likely to be compromised once the GAN model and/or its generated sample set is released.  This causes special concern when dealing with real-world privacy sensitive datasets (e.g. medical records),  which typically contain very limited data samples.

\item Differential private defense on GAN training is effective against practical MIA, but at the cost of high computation burden and deteriorated generation quality. 
\end{itemize}

\section{Conclusion}
\label{sec:conclusion}

We have established the first taxonomy of membership inference attacks against GANs, with which we hope to benchmark research in this direction in the future. %
We have also proposed the first generic attack model based on reconstruction, which is applicable to all the settings according to the amount of the attacker's knowledge about the victim model. In particular, the instantiated attack variants in the partial black-box and white-box settings are another novelty that bridges the assumption gap and performance gap in the previous work~\cite{HMDC19,HHB19}. In addition, we proposed a novel theoretically grounded attack calibration technique, which consistently improve the attack performance in all cases. Comprehensive experiments show consistent effectiveness and a broad spectrum of performance in a variety of setups spanning diverse dataset modalities, various victim models, two directions of analysis study, attack calibration, as well as differential privacy defense, which conclusively provide a better understanding of privacy risks  associated with deep generative models. %

\begin{acks} 
This work is partially funded by the Helmholtz Association within the projects "Trustworthy Federated Data Analytics" (TFDA) (funding number ZT-I-OO1 4) and "Protecting Genetic Data with Synthetic Cohorts from Deep Generative Models" (PRO-GENE-GEN) (funding number ZT-I-PF-5-23).
\end{acks}

\bibliographystyle{ACM-Reference-Format}
\bibliography{comlimentary,normal_generated_py3}


\begin{thebibliography}{73}


\ifx \showCODEN    \undefined \def \showCODEN     #1{\unskip}     \fi
\ifx \showDOI      \undefined \def \showDOI       #1{#1}\fi
\ifx \showISBNx    \undefined \def \showISBNx     #1{\unskip}     \fi
\ifx \showISBNxiii \undefined \def \showISBNxiii  #1{\unskip}     \fi
\ifx \showISSN     \undefined \def \showISSN      #1{\unskip}     \fi
\ifx \showLCCN     \undefined \def \showLCCN      #1{\unskip}     \fi
\ifx \shownote     \undefined \def \shownote      #1{#1}          \fi
\ifx \showarticletitle \undefined \def \showarticletitle #1{#1}   \fi
\ifx \showURL      \undefined \def \showURL       {\relax}        \fi
\providecommand\bibfield[2]{#2}
\providecommand\bibinfo[2]{#2}
\providecommand\natexlab[1]{#1}
\providecommand\showeprint[2][]{arXiv:#2}

\bibitem[\protect\citeauthoryear{Abadi, Chu, Goodfellow, McMahan, Mironov,
  Talwar, and Zhang}{Abadi et~al\mbox{.}}{2016}]%
        {ACGMMTZ16}
\bibfield{author}{\bibinfo{person}{Martin Abadi}, \bibinfo{person}{Andy Chu},
  \bibinfo{person}{Ian Goodfellow}, \bibinfo{person}{Brendan McMahan},
  \bibinfo{person}{Ilya Mironov}, \bibinfo{person}{Kunal Talwar}, {and}
  \bibinfo{person}{Li Zhang}.} \bibinfo{year}{2016}\natexlab{}.
\newblock \showarticletitle{{Deep Learning with Differential Privacy}}. In
  \bibinfo{booktitle}{\emph{{ACM SIGSAC Conference on Computer and
  Communications Security (CCS)}}}. \bibinfo{publisher}{ACM},
  \bibinfo{pages}{308--318}.
\newblock


\bibitem[\protect\citeauthoryear{Acs, Melis, Castelluccia, and Cristofaro}{Acs
  et~al\mbox{.}}{2017}]%
        {AMCC17}
\bibfield{author}{\bibinfo{person}{Gergely Acs}, \bibinfo{person}{Luca Melis},
  \bibinfo{person}{Claude Castelluccia}, {and} \bibinfo{person}{Emiliano~De
  Cristofaro}.} \bibinfo{year}{2017}\natexlab{}.
\newblock \showarticletitle{{Differentially Private Mixture of Generative
  Neural Networks}}. In \bibinfo{booktitle}{\emph{{International Conference on
  Data Mining (ICDM)}}}. \bibinfo{publisher}{IEEE}, \bibinfo{pages}{715--720}.
\newblock


\bibitem[\protect\citeauthoryear{Arjovsky, Chintala, and Bottou}{Arjovsky
  et~al\mbox{.}}{2017}]%
        {ACB17}
\bibfield{author}{\bibinfo{person}{Martin Arjovsky}, \bibinfo{person}{Soumith
  Chintala}, {and} \bibinfo{person}{L{\'e}on Bottou}.}
  \bibinfo{year}{2017}\natexlab{}.
\newblock \showarticletitle{{Wasserstein Generative Adversarial Networks}}. In
  \bibinfo{booktitle}{\emph{{International Conference on Machine Learning
  (ICML)}}}. \bibinfo{publisher}{JMLR}, \bibinfo{pages}{214--223}.
\newblock


\bibitem[\protect\citeauthoryear{Arora, Ge, Liang, Ma, and Zhang}{Arora
  et~al\mbox{.}}{2017}]%
        {AGLMZ17}
\bibfield{author}{\bibinfo{person}{Sanjeev Arora}, \bibinfo{person}{Rong Ge},
  \bibinfo{person}{Yingyu Liang}, \bibinfo{person}{Tengyu Ma}, {and}
  \bibinfo{person}{Yi Zhang}.} \bibinfo{year}{2017}\natexlab{}.
\newblock \showarticletitle{{Generalization and Equilibrium in Generative
  Adversarial Nets (GANs)}}. In \bibinfo{booktitle}{\emph{{International
  Conference on Machine Learning (ICML)}}}. \bibinfo{publisher}{JMLR},
  \bibinfo{pages}{224--232}.
\newblock


\bibitem[\protect\citeauthoryear{Backes, Berrang, Humbert, and
  Manoharan}{Backes et~al\mbox{.}}{2016}]%
        {BBHM16}
\bibfield{author}{\bibinfo{person}{Michael Backes}, \bibinfo{person}{Pascal
  Berrang}, \bibinfo{person}{Mathias Humbert}, {and} \bibinfo{person}{Praveen
  Manoharan}.} \bibinfo{year}{2016}\natexlab{}.
\newblock \showarticletitle{{Membership Privacy in MicroRNA-based Studies}}. In
  \bibinfo{booktitle}{\emph{{ACM SIGSAC Conference on Computer and
  Communications Security (CCS)}}}. \bibinfo{publisher}{ACM},
  \bibinfo{pages}{319--330}.
\newblock


\bibitem[\protect\citeauthoryear{Backes, Humbert, Pang, and Zhang}{Backes
  et~al\mbox{.}}{2017}]%
        {BHPZ17}
\bibfield{author}{\bibinfo{person}{Michael Backes}, \bibinfo{person}{Mathias
  Humbert}, \bibinfo{person}{Jun Pang}, {and} \bibinfo{person}{Yang Zhang}.}
  \bibinfo{year}{2017}\natexlab{}.
\newblock \showarticletitle{{walk2friends: Inferring Social Links from Mobility
  Profiles}}. In \bibinfo{booktitle}{\emph{{ACM SIGSAC Conference on Computer
  and Communications Security (CCS)}}}. \bibinfo{publisher}{ACM},
  \bibinfo{pages}{1943--1957}.
\newblock


\bibitem[\protect\citeauthoryear{Beaulieu-Jones, Wu, Williams, Lee, Bhavnani,
  Byrd, and Greene}{Beaulieu-Jones et~al\mbox{.}}{2019}]%
        {beaulieu2019privacy}
\bibfield{author}{\bibinfo{person}{Brett~K Beaulieu-Jones},
  \bibinfo{person}{Zhiwei~Steven Wu}, \bibinfo{person}{Chris Williams},
  \bibinfo{person}{Ran Lee}, \bibinfo{person}{Sanjeev~P Bhavnani},
  \bibinfo{person}{James~Brian Byrd}, {and} \bibinfo{person}{Casey~S Greene}.}
  \bibinfo{year}{2019}\natexlab{}.
\newblock \showarticletitle{Privacy-Preserving Generative Deep Neural Networks
  Support Clinical Data Sharing}.
\newblock \bibinfo{journal}{\emph{Circulation: Cardiovascular Quality and
  Outcomes}} \bibinfo{volume}{12}, \bibinfo{number}{7} (\bibinfo{year}{2019}),
  \bibinfo{pages}{e005122}.
\newblock


\bibitem[\protect\citeauthoryear{Bhattacharyya, Fritz, and
  Schiele}{Bhattacharyya et~al\mbox{.}}{2019}]%
        {BFS19}
\bibfield{author}{\bibinfo{person}{Apratim Bhattacharyya},
  \bibinfo{person}{Mario Fritz}, {and} \bibinfo{person}{Bernt Schiele}.}
  \bibinfo{year}{2019}\natexlab{}.
\newblock \showarticletitle{{``Best-of-Many-Samples'' Distribution Matching}}.
\newblock \bibinfo{journal}{\emph{{CoRR abs/1909.12598}}}
  (\bibinfo{year}{2019}).
\newblock


\bibitem[\protect\citeauthoryear{Boiman, Shechtman, and Irani}{Boiman
  et~al\mbox{.}}{2008}]%
        {BSI08}
\bibfield{author}{\bibinfo{person}{Oren Boiman}, \bibinfo{person}{Eli
  Shechtman}, {and} \bibinfo{person}{Michal Irani}.}
  \bibinfo{year}{2008}\natexlab{}.
\newblock \showarticletitle{{In Defense of Nearest-Neighbor based Image
  Classification}}. In \bibinfo{booktitle}{\emph{{IEEE Conference on Computer
  Vision and Pattern Recognition (CVPR)}}}. \bibinfo{publisher}{IEEE}.
\newblock


\bibitem[\protect\citeauthoryear{Brock, Donahue, and Simonyan}{Brock
  et~al\mbox{.}}{2019}]%
        {BDS19}
\bibfield{author}{\bibinfo{person}{Andrew Brock}, \bibinfo{person}{Jeff
  Donahue}, {and} \bibinfo{person}{Karen Simonyan}.}
  \bibinfo{year}{2019}\natexlab{}.
\newblock \showarticletitle{{Large Scale GAN Training for High Fidelity Natural
  Image Synthesis}}. In \bibinfo{booktitle}{\emph{{International Conference on
  Learning Representations (ICLR)}}}.
\newblock


\bibitem[\protect\citeauthoryear{Chen, Orekondy, and Fritz}{Chen
  et~al\mbox{.}}{2020}]%
        {COF20}
\bibfield{author}{\bibinfo{person}{Dingfan Chen}, \bibinfo{person}{Tribhuvanesh
  Orekondy}, {and} \bibinfo{person}{Mario Fritz}.}
  \bibinfo{year}{2020}\natexlab{}.
\newblock \showarticletitle{{GS-WGAN: A Gradient-Sanitized Approach for
  Learning Differentially Private Generators}}.
\newblock \bibinfo{journal}{\emph{{CoRR abs/2006.08265}}}
  (\bibinfo{year}{2020}).
\newblock


\bibitem[\protect\citeauthoryear{Choi, Biswal, Malin, Duke, Stewart, and
  Sun}{Choi et~al\mbox{.}}{2018}]%
        {CBMDSS18}
\bibfield{author}{\bibinfo{person}{Edward Choi}, \bibinfo{person}{Siddharth
  Biswal}, \bibinfo{person}{Bradley Malin}, \bibinfo{person}{Jon Duke},
  \bibinfo{person}{Walter~F. Stewart}, {and} \bibinfo{person}{Jimeng Sun}.}
  \bibinfo{year}{2018}\natexlab{}.
\newblock \showarticletitle{{Generating Multi-label Discrete Patient Records
  using Generative Adversarial Networks}}.
\newblock \bibinfo{journal}{\emph{{CoRR abs/1703.06490}}}
  (\bibinfo{year}{2018}).
\newblock


\bibitem[\protect\citeauthoryear{Dinh, Krueger, and Bengio}{Dinh
  et~al\mbox{.}}{2015}]%
        {DKB15}
\bibfield{author}{\bibinfo{person}{Laurent Dinh}, \bibinfo{person}{David
  Krueger}, {and} \bibinfo{person}{Yoshua Bengio}.}
  \bibinfo{year}{2015}\natexlab{}.
\newblock \showarticletitle{{NICE: Non-linear Independent Components
  Estimation}}.
\newblock \bibinfo{journal}{\emph{{CoRR abs/1410.8516}}}
  (\bibinfo{year}{2015}).
\newblock


\bibitem[\protect\citeauthoryear{Dinh, Sohl-Dickstein, and Bengio}{Dinh
  et~al\mbox{.}}{2017}]%
        {DSB17}
\bibfield{author}{\bibinfo{person}{Laurent Dinh}, \bibinfo{person}{Jascha
  Sohl-Dickstein}, {and} \bibinfo{person}{Samy Bengio}.}
  \bibinfo{year}{2017}\natexlab{}.
\newblock \showarticletitle{{Density Estimation using Real NVP}}. In
  \bibinfo{booktitle}{\emph{{International Conference on Learning
  Representations (ICLR)}}}.
\newblock


\bibitem[\protect\citeauthoryear{Duda, Hart, and Stork}{Duda
  et~al\mbox{.}}{2012}]%
        {DHS12}
\bibfield{author}{\bibinfo{person}{Richard~O Duda}, \bibinfo{person}{Peter~E
  Hart}, {and} \bibinfo{person}{David~G Stork}.}
  \bibinfo{year}{2012}\natexlab{}.
\newblock \bibinfo{booktitle}{\emph{{Pattern Classification}}}.
\newblock \bibinfo{publisher}{John Wiley \& Sons}.
\newblock


\bibitem[\protect\citeauthoryear{Dwork, McSherry, Nissim, and Smith}{Dwork
  et~al\mbox{.}}{2006}]%
        {DMNS06}
\bibfield{author}{\bibinfo{person}{Cynthia Dwork}, \bibinfo{person}{Frank
  McSherry}, \bibinfo{person}{Kobbi Nissim}, {and} \bibinfo{person}{Adam
  Smith}.} \bibinfo{year}{2006}\natexlab{}.
\newblock \showarticletitle{{Calibrating Noise to Sensitivity in Private Data
  Analysis}}. In \bibinfo{booktitle}{\emph{{Theory of Cryptography Conference
  (TCC)}}}. \bibinfo{publisher}{Springer}, \bibinfo{pages}{265--284}.
\newblock


\bibitem[\protect\citeauthoryear{Dwork and Roth}{Dwork and Roth}{2014}]%
        {DR14}
\bibfield{author}{\bibinfo{person}{Cynthia Dwork} {and} \bibinfo{person}{Aaron
  Roth}.} \bibinfo{year}{2014}\natexlab{}.
\newblock \bibinfo{booktitle}{\emph{{The Algorithmic Foundations of
  Differential Privacy}}}.
\newblock \bibinfo{publisher}{Now Publishers Inc.}
\newblock


\bibitem[\protect\citeauthoryear{Dwork, Smith, Steinke, Ullman, and
  Vadhan}{Dwork et~al\mbox{.}}{2015}]%
        {DSSUV15}
\bibfield{author}{\bibinfo{person}{Cynthia Dwork}, \bibinfo{person}{Adam~D.
  Smith}, \bibinfo{person}{Thomas Steinke}, \bibinfo{person}{Jonathan Ullman},
  {and} \bibinfo{person}{Salil~P. Vadhan}.} \bibinfo{year}{2015}\natexlab{}.
\newblock \showarticletitle{{Robust Traceability from Trace Amounts}}. In
  \bibinfo{booktitle}{\emph{{Annual Symposium on Foundations of Computer
  Science (FOCS)}}}. \bibinfo{publisher}{IEEE}, \bibinfo{pages}{650--669}.
\newblock


\bibitem[\protect\citeauthoryear{Frid-Adar, Klang, Amitai, Goldberger, and
  Greenspan}{Frid-Adar et~al\mbox{.}}{2018}]%
        {FKAGG18}
\bibfield{author}{\bibinfo{person}{Maayan Frid-Adar}, \bibinfo{person}{Eyal
  Klang}, \bibinfo{person}{Michal Amitai}, \bibinfo{person}{Jacob Goldberger},
  {and} \bibinfo{person}{Hayit Greenspan}.} \bibinfo{year}{2018}\natexlab{}.
\newblock \showarticletitle{{Synthetic Data Augmentation using GAN for Improved
  Liver Lesion Classification}}. In \bibinfo{booktitle}{\emph{{IEEE
  International Symposium on Biomedical Imaging (ISBI)}}}.
  \bibinfo{publisher}{IEEE}, \bibinfo{pages}{289--293}.
\newblock


\bibitem[\protect\citeauthoryear{Goodfellow, Pouget-Abadie, Mirza, Xu,
  Warde-Farley, Ozair, Courville, and Bengio}{Goodfellow et~al\mbox{.}}{2014}]%
        {GPMXWOCB14}
\bibfield{author}{\bibinfo{person}{Ian Goodfellow}, \bibinfo{person}{Jean
  Pouget-Abadie}, \bibinfo{person}{Mehdi Mirza}, \bibinfo{person}{Bing Xu},
  \bibinfo{person}{David Warde-Farley}, \bibinfo{person}{Sherjil Ozair},
  \bibinfo{person}{Aaron Courville}, {and} \bibinfo{person}{Yoshua Bengio}.}
  \bibinfo{year}{2014}\natexlab{}.
\newblock \showarticletitle{{Generative Adversarial Nets}}. In
  \bibinfo{booktitle}{\emph{{Annual Conference on Neural Information Processing
  Systems (NIPS)}}}. \bibinfo{publisher}{NIPS}, \bibinfo{pages}{2672--2680}.
\newblock


\bibitem[\protect\citeauthoryear{Graves, rahman Mohamed, and Hinton}{Graves
  et~al\mbox{.}}{2013}]%
        {GMH13}
\bibfield{author}{\bibinfo{person}{Alex Graves}, \bibinfo{person}{Abdel rahman
  Mohamed}, {and} \bibinfo{person}{Geoffrey~E. Hinton}.}
  \bibinfo{year}{2013}\natexlab{}.
\newblock \showarticletitle{{Speech Recognition with Deep Recurrent Neural
  Networks}}. In \bibinfo{booktitle}{\emph{{IEEE International Conference on
  Acoustics, Speech and Signal Processing (ICASSP)}}}.
  \bibinfo{publisher}{IEEE}, \bibinfo{pages}{6645--6649}.
\newblock


\bibitem[\protect\citeauthoryear{Gu, Shen, and Zhou}{Gu et~al\mbox{.}}{2019}]%
        {GSZ19}
\bibfield{author}{\bibinfo{person}{Jinjin Gu}, \bibinfo{person}{Yujun Shen},
  {and} \bibinfo{person}{Bolei Zhou}.} \bibinfo{year}{2019}\natexlab{}.
\newblock \showarticletitle{{Image Processing Using Multi-Code GAN Prior}}.
\newblock \bibinfo{journal}{\emph{{CoRR abs/1912.07116}}}
  (\bibinfo{year}{2019}).
\newblock


\bibitem[\protect\citeauthoryear{Gulrajani, Ahmed, Arjovsky, Dumoulin, and
  Courville}{Gulrajani et~al\mbox{.}}{2017}]%
        {GAADC17}
\bibfield{author}{\bibinfo{person}{Ishaan Gulrajani}, \bibinfo{person}{Faruk
  Ahmed}, \bibinfo{person}{Martin Arjovsky}, \bibinfo{person}{Vincent
  Dumoulin}, {and} \bibinfo{person}{Aaron~C. Courville}.}
  \bibinfo{year}{2017}\natexlab{}.
\newblock \showarticletitle{{Improved Training of Wasserstein GANs}}. In
  \bibinfo{booktitle}{\emph{{Annual Conference on Neural Information Processing
  Systems (NIPS)}}}. \bibinfo{publisher}{NIPS}, \bibinfo{pages}{5767--5777}.
\newblock


\bibitem[\protect\citeauthoryear{Hagestedt, Zhang, Humbert, Berrang, Tang,
  Wang, and Backes}{Hagestedt et~al\mbox{.}}{2019}]%
        {HZHBTWB19}
\bibfield{author}{\bibinfo{person}{Inken Hagestedt}, \bibinfo{person}{Yang
  Zhang}, \bibinfo{person}{Mathias Humbert}, \bibinfo{person}{Pascal Berrang},
  \bibinfo{person}{Haixu Tang}, \bibinfo{person}{XiaoFeng Wang}, {and}
  \bibinfo{person}{Michael Backes}.} \bibinfo{year}{2019}\natexlab{}.
\newblock \showarticletitle{{MBeacon: Privacy-Preserving Beacons for DNA
  Methylation Data}}. In \bibinfo{booktitle}{\emph{{Network and Distributed
  System Security Symposium (NDSS)}}}. \bibinfo{publisher}{Internet Society}.
\newblock


\bibitem[\protect\citeauthoryear{Hayes, Melis, Danezis, and Cristofaro}{Hayes
  et~al\mbox{.}}{2019}]%
        {HMDC19}
\bibfield{author}{\bibinfo{person}{Jamie Hayes}, \bibinfo{person}{Luca Melis},
  \bibinfo{person}{George Danezis}, {and} \bibinfo{person}{Emiliano~De
  Cristofaro}.} \bibinfo{year}{2019}\natexlab{}.
\newblock \showarticletitle{{LOGAN: Evaluating Privacy Leakage of Generative
  Models Using Generative Adversarial Networks}}.
\newblock \bibinfo{journal}{\emph{{Symposium on Privacy Enhancing Technologies
  Symposium}}} (\bibinfo{year}{2019}).
\newblock


\bibitem[\protect\citeauthoryear{He, Zhang, Ren, and Sun}{He
  et~al\mbox{.}}{2016}]%
        {HZRS16}
\bibfield{author}{\bibinfo{person}{Kaiming He}, \bibinfo{person}{Xiangyu
  Zhang}, \bibinfo{person}{Shaoqing Ren}, {and} \bibinfo{person}{Jian Sun}.}
  \bibinfo{year}{2016}\natexlab{}.
\newblock \showarticletitle{{Deep Residual Learning for Image Recognition}}. In
  \bibinfo{booktitle}{\emph{{IEEE Conference on Computer Vision and Pattern
  Recognition (CVPR)}}}. \bibinfo{publisher}{IEEE}, \bibinfo{pages}{770--778}.
\newblock


\bibitem[\protect\citeauthoryear{He, Zuo, Kan, Shan, and Chen}{He
  et~al\mbox{.}}{2018}]%
        {HZKSC18}
\bibfield{author}{\bibinfo{person}{Zhenliang He}, \bibinfo{person}{Wangmeng
  Zuo}, \bibinfo{person}{Meina Kan}, \bibinfo{person}{Shiguang Shan}, {and}
  \bibinfo{person}{Xilin Chen}.} \bibinfo{year}{2018}\natexlab{}.
\newblock \showarticletitle{{AttGAN: Facial Attribute Editing by Only Changing
  What You Want}}.
\newblock \bibinfo{journal}{\emph{{CoRR abs/1711.10678}}}
  (\bibinfo{year}{2018}).
\newblock


\bibitem[\protect\citeauthoryear{Heusel, Ramsauer, Unterthiner, Nessler, and
  Hochreiter}{Heusel et~al\mbox{.}}{2017}]%
        {HRUNH17}
\bibfield{author}{\bibinfo{person}{Martin Heusel}, \bibinfo{person}{Hubert
  Ramsauer}, \bibinfo{person}{Thomas Unterthiner}, \bibinfo{person}{Bernhard
  Nessler}, {and} \bibinfo{person}{Sepp Hochreiter}.}
  \bibinfo{year}{2017}\natexlab{}.
\newblock \showarticletitle{{GANs Trained by a Two Time-Scale Update Rule
  Converge to a Local Nash Equilibrium}}. In \bibinfo{booktitle}{\emph{{Annual
  Conference on Neural Information Processing Systems (NIPS)}}}.
  \bibinfo{publisher}{NIPS}, \bibinfo{pages}{6626--6637}.
\newblock


\bibitem[\protect\citeauthoryear{Hilprecht, H{\"{a}}rterich, and
  Bernau}{Hilprecht et~al\mbox{.}}{2019}]%
        {HHB19}
\bibfield{author}{\bibinfo{person}{Benjamin Hilprecht}, \bibinfo{person}{Martin
  H{\"{a}}rterich}, {and} \bibinfo{person}{Daniel Bernau}.}
  \bibinfo{year}{2019}\natexlab{}.
\newblock \showarticletitle{{Monte Carlo and Reconstruction Membership
  Inference Attacks against Generative Models}}.
\newblock \bibinfo{journal}{\emph{{Symposium on Privacy Enhancing Technologies
  Symposium}}} (\bibinfo{year}{2019}).
\newblock


\bibitem[\protect\citeauthoryear{Hinton, Deng, Yu, Dahl, rahman Mohamed,
  Jaitly, Senior, Vanhoucke, Nguyen, Kingsbury, et~al\mbox{.}}{Hinton
  et~al\mbox{.}}{2012}]%
        {hinton2012deep}
\bibfield{author}{\bibinfo{person}{Geoffrey Hinton}, \bibinfo{person}{Li Deng},
  \bibinfo{person}{Dong Yu}, \bibinfo{person}{George Dahl},
  \bibinfo{person}{Abdel rahman Mohamed}, \bibinfo{person}{Navdeep Jaitly},
  \bibinfo{person}{Andrew Senior}, \bibinfo{person}{Vincent Vanhoucke},
  \bibinfo{person}{Patrick Nguyen}, \bibinfo{person}{Brian Kingsbury},
  {et~al\mbox{.}}} \bibinfo{year}{2012}\natexlab{}.
\newblock \showarticletitle{Deep Neural Networks for Acoustic Modeling in
  Speech Recognition}.
\newblock \bibinfo{journal}{\emph{IEEE Signal Processing Magazine}}
  \bibinfo{volume}{29} (\bibinfo{year}{2012}).
\newblock


\bibitem[\protect\citeauthoryear{Huang, Mattar, Berg, and Learned-Miller}{Huang
  et~al\mbox{.}}{2008}]%
        {huang2008labeled}
\bibfield{author}{\bibinfo{person}{Gary~B Huang}, \bibinfo{person}{Marwan
  Mattar}, \bibinfo{person}{Tamara Berg}, {and} \bibinfo{person}{Eric
  Learned-Miller}.} \bibinfo{year}{2008}\natexlab{}.
\newblock \showarticletitle{Labeled Faces in the Wild: A Database for Studying
  Face Recognition in Unconstrained Environments}.
\newblock


\bibitem[\protect\citeauthoryear{Jahanian, Chai, and Isola}{Jahanian
  et~al\mbox{.}}{2019}]%
        {JCI19}
\bibfield{author}{\bibinfo{person}{Ali Jahanian}, \bibinfo{person}{Lucy Chai},
  {and} \bibinfo{person}{Phillip Isola}.} \bibinfo{year}{2019}\natexlab{}.
\newblock \showarticletitle{{On the ``Steerability'' of Generative Adversarial
  Networks}}.
\newblock \bibinfo{journal}{\emph{{CoRR abs/1907.07171}}}
  (\bibinfo{year}{2019}).
\newblock


\bibitem[\protect\citeauthoryear{Jia, Salem, Backes, Zhang, and Gong}{Jia
  et~al\mbox{.}}{2019}]%
        {JSBZG19}
\bibfield{author}{\bibinfo{person}{Jinyuan Jia}, \bibinfo{person}{Ahmed Salem},
  \bibinfo{person}{Michael Backes}, \bibinfo{person}{Yang Zhang}, {and}
  \bibinfo{person}{Neil~Zhenqiang Gong}.} \bibinfo{year}{2019}\natexlab{}.
\newblock \showarticletitle{{MemGuard: Defending against Black-Box Membership
  Inference Attacks via Adversarial Examples}}. In
  \bibinfo{booktitle}{\emph{{ACM SIGSAC Conference on Computer and
  Communications Security (CCS)}}}. \bibinfo{publisher}{ACM},
  \bibinfo{pages}{259--274}.
\newblock


\bibitem[\protect\citeauthoryear{Johnson, Pollard, Shen, Li-wei, Feng,
  Ghassemi, Moody, Szolovits, Celi, and Mark}{Johnson et~al\mbox{.}}{2016}]%
        {johnson2016mimic}
\bibfield{author}{\bibinfo{person}{Alistair~EW Johnson}, \bibinfo{person}{Tom~J
  Pollard}, \bibinfo{person}{Lu Shen}, \bibinfo{person}{H~Lehman Li-wei},
  \bibinfo{person}{Mengling Feng}, \bibinfo{person}{Mohammad Ghassemi},
  \bibinfo{person}{Benjamin Moody}, \bibinfo{person}{Peter Szolovits},
  \bibinfo{person}{Leo~Anthony Celi}, {and} \bibinfo{person}{Roger~G Mark}.}
  \bibinfo{year}{2016}\natexlab{}.
\newblock \showarticletitle{MIMIC-III, A Freely Accessible Critical Care
  Database}.
\newblock \bibinfo{journal}{\emph{Scientific Data}}  \bibinfo{volume}{3}
  (\bibinfo{year}{2016}), \bibinfo{pages}{160035}.
\newblock


\bibitem[\protect\citeauthoryear{Jordon, Yoon, and van~der Schaar}{Jordon
  et~al\mbox{.}}{2019}]%
        {JYS19}
\bibfield{author}{\bibinfo{person}{James Jordon}, \bibinfo{person}{Jinsung
  Yoon}, {and} \bibinfo{person}{Mihaela van~der Schaar}.}
  \bibinfo{year}{2019}\natexlab{}.
\newblock \showarticletitle{{PATE-GAN: Generating Synthetic Data with
  Differential Privacy Guarantees}}.
\newblock \bibinfo{journal}{\emph{{OpenReview}}} (\bibinfo{year}{2019}).
\newblock


\bibitem[\protect\citeauthoryear{Karras, Aila, Laine, and Lehtinen}{Karras
  et~al\mbox{.}}{2018}]%
        {KALL18}
\bibfield{author}{\bibinfo{person}{Tero Karras}, \bibinfo{person}{Timo Aila},
  \bibinfo{person}{Samuli Laine}, {and} \bibinfo{person}{Jaakko Lehtinen}.}
  \bibinfo{year}{2018}\natexlab{}.
\newblock \showarticletitle{{Progressive Growing of GANs for Improved Quality,
  Stability, and Variation}}. In \bibinfo{booktitle}{\emph{{International
  Conference on Learning Representations (ICLR)}}}.
\newblock


\bibitem[\protect\citeauthoryear{Karras, Laine, and Aila}{Karras
  et~al\mbox{.}}{2019}]%
        {KLA19}
\bibfield{author}{\bibinfo{person}{Tero Karras}, \bibinfo{person}{Samuli
  Laine}, {and} \bibinfo{person}{Timo Aila}.} \bibinfo{year}{2019}\natexlab{}.
\newblock \showarticletitle{{A Style-Based Generator Architecture for
  Generative Adversarial Networks}}. In \bibinfo{booktitle}{\emph{{IEEE
  Conference on Computer Vision and Pattern Recognition (CVPR)}}}.
  \bibinfo{publisher}{IEEE}, \bibinfo{pages}{4401--4410}.
\newblock


\bibitem[\protect\citeauthoryear{Karras, Laine, Aittala, Hellsten, Lehtinen,
  and Aila}{Karras et~al\mbox{.}}{2020}]%
        {KLAHLA20}
\bibfield{author}{\bibinfo{person}{Tero Karras}, \bibinfo{person}{Samuli
  Laine}, \bibinfo{person}{Miika Aittala}, \bibinfo{person}{Janne Hellsten},
  \bibinfo{person}{Jaakko Lehtinen}, {and} \bibinfo{person}{Timo Aila}.}
  \bibinfo{year}{2020}\natexlab{}.
\newblock \showarticletitle{{Analyzing and Improving the Image Quality of
  StyleGAN}}. In \bibinfo{booktitle}{\emph{{IEEE Conference on Computer Vision
  and Pattern Recognition (CVPR)}}}. \bibinfo{publisher}{IEEE},
  \bibinfo{pages}{8107--8116}.
\newblock


\bibitem[\protect\citeauthoryear{Kingma and Ba}{Kingma and Ba}{2015}]%
        {KB15}
\bibfield{author}{\bibinfo{person}{Diederik~P. Kingma} {and}
  \bibinfo{person}{Jimmy Ba}.} \bibinfo{year}{2015}\natexlab{}.
\newblock \showarticletitle{{Adam: A Method for Stochastic Optimization}}. In
  \bibinfo{booktitle}{\emph{{International Conference on Learning
  Representations (ICLR)}}}.
\newblock


\bibitem[\protect\citeauthoryear{Kingma and Dhariwal}{Kingma and
  Dhariwal}{2018}]%
        {KD18}
\bibfield{author}{\bibinfo{person}{Diederik~P. Kingma} {and}
  \bibinfo{person}{Prafulla Dhariwal}.} \bibinfo{year}{2018}\natexlab{}.
\newblock \showarticletitle{{Glow: Generative Flow with Invertible 1x1
  Convolutions}}. In \bibinfo{booktitle}{\emph{{Annual Conference on Neural
  Information Processing Systems (NeurIPS)}}}. \bibinfo{publisher}{NeurIPS},
  \bibinfo{pages}{10236--10245}.
\newblock


\bibitem[\protect\citeauthoryear{Kingma and Welling}{Kingma and
  Welling}{2014}]%
        {KW14}
\bibfield{author}{\bibinfo{person}{Diederik~P. Kingma} {and}
  \bibinfo{person}{Max Welling}.} \bibinfo{year}{2014}\natexlab{}.
\newblock \showarticletitle{{Auto-Encoding Variational Bayes}}. In
  \bibinfo{booktitle}{\emph{{International Conference on Learning
  Representations (ICLR)}}}.
\newblock


\bibitem[\protect\citeauthoryear{Krizhevsky, Sutskever, and Hinton}{Krizhevsky
  et~al\mbox{.}}{2012}]%
        {KSH12}
\bibfield{author}{\bibinfo{person}{Alex Krizhevsky}, \bibinfo{person}{Ilya
  Sutskever}, {and} \bibinfo{person}{Geoffrey~E. Hinton}.}
  \bibinfo{year}{2012}\natexlab{}.
\newblock \showarticletitle{{ImageNet Classification with Deep Convolutional
  Neural Networks}}. In \bibinfo{booktitle}{\emph{{Annual Conference on Neural
  Information Processing Systems (NIPS)}}}. \bibinfo{publisher}{NIPS},
  \bibinfo{pages}{1106--1114}.
\newblock


\bibitem[\protect\citeauthoryear{Larsen, S{\o}nderby, Larochelle, and
  Winther}{Larsen et~al\mbox{.}}{2016}]%
        {LSLW16}
\bibfield{author}{\bibinfo{person}{Anders Boesen~Lindbo Larsen},
  \bibinfo{person}{S{\o}ren~Kaae S{\o}nderby}, \bibinfo{person}{Hugo
  Larochelle}, {and} \bibinfo{person}{Ole Winther}.}
  \bibinfo{year}{2016}\natexlab{}.
\newblock \showarticletitle{{Autoencoding beyond Pixels Using a Learned
  Similarity Metric}}. In \bibinfo{booktitle}{\emph{{International Conference
  on Machine Learning (ICML)}}}. \bibinfo{publisher}{JMLR},
  \bibinfo{pages}{1558--1566}.
\newblock


\bibitem[\protect\citeauthoryear{Li and Wand}{Li and Wand}{2016}]%
        {LW16}
\bibfield{author}{\bibinfo{person}{Chuan Li} {and} \bibinfo{person}{Michael
  Wand}.} \bibinfo{year}{2016}\natexlab{}.
\newblock \showarticletitle{{Precomputed Real-Time Texture Synthesis with
  Markovian Generative Adversarial Networks}}. In
  \bibinfo{booktitle}{\emph{{European Conference on Computer Vision (ECCV)}}}.
  \bibinfo{publisher}{Springer}, \bibinfo{pages}{702--716}.
\newblock


\bibitem[\protect\citeauthoryear{Li and Zhang}{Li and Zhang}{2020}]%
        {LZ20}
\bibfield{author}{\bibinfo{person}{Zheng Li} {and} \bibinfo{person}{Yang
  Zhang}.} \bibinfo{year}{2020}\natexlab{}.
\newblock \showarticletitle{{Label-Leaks: Membership Inference Attack with
  Label}}.
\newblock \bibinfo{journal}{\emph{{CoRR abs/2007.15528}}}
  (\bibinfo{year}{2020}).
\newblock


\bibitem[\protect\citeauthoryear{Liu and Nocedal}{Liu and Nocedal}{1989}]%
        {liu1989limited}
\bibfield{author}{\bibinfo{person}{Dong~C Liu} {and} \bibinfo{person}{Jorge
  Nocedal}.} \bibinfo{year}{1989}\natexlab{}.
\newblock \showarticletitle{On the Limited Memory BFGS Method for Large Scale
  Optimization}.
\newblock \bibinfo{journal}{\emph{Mathematical Programming}}
  \bibinfo{volume}{45}, \bibinfo{number}{1-3} (\bibinfo{year}{1989}),
  \bibinfo{pages}{503--528}.
\newblock


\bibitem[\protect\citeauthoryear{Liu, Luo, Wang, and Tang}{Liu
  et~al\mbox{.}}{2015}]%
        {LLWT15}
\bibfield{author}{\bibinfo{person}{Ziwei Liu}, \bibinfo{person}{Ping Luo},
  \bibinfo{person}{Xiaogang Wang}, {and} \bibinfo{person}{Xiaoou Tang}.}
  \bibinfo{year}{2015}\natexlab{}.
\newblock \showarticletitle{{Deep Learning Face Attributes in the Wild}}. In
  \bibinfo{booktitle}{\emph{{IEEE International Conference on Computer Vision
  (ICCV)}}}. \bibinfo{publisher}{IEEE}, \bibinfo{pages}{3730--3738}.
\newblock


\bibitem[\protect\citeauthoryear{Long, Bindschaedler, Wang, Bu, Wang, Tang,
  Gunter, and Chen}{Long et~al\mbox{.}}{2018}]%
        {LBWBWTGC18}
\bibfield{author}{\bibinfo{person}{Yunhui Long}, \bibinfo{person}{Vincent
  Bindschaedler}, \bibinfo{person}{Lei Wang}, \bibinfo{person}{Diyue Bu},
  \bibinfo{person}{Xiaofeng Wang}, \bibinfo{person}{Haixu Tang},
  \bibinfo{person}{Carl~A. Gunter}, {and} \bibinfo{person}{Kai Chen}.}
  \bibinfo{year}{2018}\natexlab{}.
\newblock \showarticletitle{{Understanding Membership Inferences on
  Well-Generalized Learning Models}}.
\newblock \bibinfo{journal}{\emph{{CoRR abs/1802.04889}}}
  (\bibinfo{year}{2018}).
\newblock


\bibitem[\protect\citeauthoryear{Mehri, Kumar, Gulrajani, Kumar, Jain, Sotelo,
  Courville, and Bengio}{Mehri et~al\mbox{.}}{2017}]%
        {MKGKJSCB17}
\bibfield{author}{\bibinfo{person}{Soroush Mehri}, \bibinfo{person}{Kundan
  Kumar}, \bibinfo{person}{Ishaan Gulrajani}, \bibinfo{person}{Rithesh Kumar},
  \bibinfo{person}{Shubham Jain}, \bibinfo{person}{Jose Sotelo},
  \bibinfo{person}{Aaron~C. Courville}, {and} \bibinfo{person}{Yoshua Bengio}.}
  \bibinfo{year}{2017}\natexlab{}.
\newblock \showarticletitle{{SampleRNN: An Unconditional End-to-End Neural
  Audio Generation Model}}. In \bibinfo{booktitle}{\emph{{International
  Conference on Learning Representations (ICLR)}}}.
\newblock


\bibitem[\protect\citeauthoryear{Melis, Song, Cristofaro, and Shmatikov}{Melis
  et~al\mbox{.}}{2019}]%
        {MSCS19}
\bibfield{author}{\bibinfo{person}{Luca Melis}, \bibinfo{person}{Congzheng
  Song}, \bibinfo{person}{Emiliano~De Cristofaro}, {and}
  \bibinfo{person}{Vitaly Shmatikov}.} \bibinfo{year}{2019}\natexlab{}.
\newblock \showarticletitle{{Exploiting Unintended Feature Leakage in
  Collaborative Learning}}. In \bibinfo{booktitle}{\emph{{IEEE Symposium on
  Security and Privacy (S\&P)}}}. \bibinfo{publisher}{IEEE},
  \bibinfo{pages}{497--512}.
\newblock


\bibitem[\protect\citeauthoryear{Nasr, Shokri, and Houmansadr}{Nasr
  et~al\mbox{.}}{2019}]%
        {NSH19}
\bibfield{author}{\bibinfo{person}{Milad Nasr}, \bibinfo{person}{Reza Shokri},
  {and} \bibinfo{person}{Amir Houmansadr}.} \bibinfo{year}{2019}\natexlab{}.
\newblock \showarticletitle{{Comprehensive Privacy Analysis of Deep Learning:
  Passive and Active White-box Inference Attacks against Centralized and
  Federated Learning}}. In \bibinfo{booktitle}{\emph{{IEEE Symposium on
  Security and Privacy (S\&P)}}}. \bibinfo{publisher}{IEEE},
  \bibinfo{pages}{1021--1035}.
\newblock


\bibitem[\protect\citeauthoryear{Pathak, Kr{\"a}henb{\"u}hl, Donahue, Darrell,
  and Efros}{Pathak et~al\mbox{.}}{2016}]%
        {PKDDE16}
\bibfield{author}{\bibinfo{person}{Deepak Pathak}, \bibinfo{person}{Philipp
  Kr{\"a}henb{\"u}hl}, \bibinfo{person}{Jeff Donahue}, \bibinfo{person}{Trevor
  Darrell}, {and} \bibinfo{person}{Alexei~A. Efros}.}
  \bibinfo{year}{2016}\natexlab{}.
\newblock \showarticletitle{{Context Encoders: Feature Learning by
  Inpainting}}. In \bibinfo{booktitle}{\emph{{IEEE Conference on Computer
  Vision and Pattern Recognition (CVPR)}}}. \bibinfo{publisher}{IEEE},
  \bibinfo{pages}{2536--2544}.
\newblock


\bibitem[\protect\citeauthoryear{Powell}{Powell}{1964}]%
        {powell1964efficient}
\bibfield{author}{\bibinfo{person}{Michael~JD Powell}.}
  \bibinfo{year}{1964}\natexlab{}.
\newblock \showarticletitle{An Efficient Method for Finding the Minimum of a
  Function of Several Variables without Calculating Derivatives}.
\newblock \bibinfo{journal}{\emph{Comput. J.}} \bibinfo{volume}{7},
  \bibinfo{number}{2} (\bibinfo{year}{1964}), \bibinfo{pages}{155--162}.
\newblock


\bibitem[\protect\citeauthoryear{Radford, Metz, and Chintala}{Radford
  et~al\mbox{.}}{2015}]%
        {RMC15}
\bibfield{author}{\bibinfo{person}{Alec Radford}, \bibinfo{person}{Luke Metz},
  {and} \bibinfo{person}{Soumith Chintala}.} \bibinfo{year}{2015}\natexlab{}.
\newblock \showarticletitle{{Unsupervised Representation Learning with Deep
  Convolutional Generative Adversarial Networks}}.
\newblock \bibinfo{journal}{\emph{{CoRR abs/1511.06434}}}
  (\bibinfo{year}{2015}).
\newblock


\bibitem[\protect\citeauthoryear{Rezende, Mohamed, and Wierstra}{Rezende
  et~al\mbox{.}}{2014}]%
        {RMW14}
\bibfield{author}{\bibinfo{person}{Danilo~Jimenez Rezende},
  \bibinfo{person}{Shakir Mohamed}, {and} \bibinfo{person}{Daan Wierstra}.}
  \bibinfo{year}{2014}\natexlab{}.
\newblock \showarticletitle{{Stochastic Backpropagation and Approximate
  Inference in Deep Generative Models}}.
\newblock \bibinfo{journal}{\emph{{CoRR abs/1401.4082}}}
  (\bibinfo{year}{2014}).
\newblock


\bibitem[\protect\citeauthoryear{Russakovsky, Deng, Su, Krause, Satheesh, Ma,
  Huang, Karpathy, Khosla, Bernstein, Berg, and Fei-Fei}{Russakovsky
  et~al\mbox{.}}{2015}]%
        {RDSKSMHKKBBF15}
\bibfield{author}{\bibinfo{person}{Olga Russakovsky}, \bibinfo{person}{Jia
  Deng}, \bibinfo{person}{Hao Su}, \bibinfo{person}{Jonathan Krause},
  \bibinfo{person}{Sanjeev Satheesh}, \bibinfo{person}{Sean Ma},
  \bibinfo{person}{Zhiheng Huang}, \bibinfo{person}{Andrej Karpathy},
  \bibinfo{person}{Aditya Khosla}, \bibinfo{person}{Michael Bernstein},
  \bibinfo{person}{Alexander~C. Berg}, {and} \bibinfo{person}{Li Fei-Fei}.}
  \bibinfo{year}{2015}\natexlab{}.
\newblock \showarticletitle{{ImageNet Large Scale Visual Recognition
  Challenge}}.
\newblock \bibinfo{journal}{\emph{{CoRR abs/1409.0575}}}
  (\bibinfo{year}{2015}).
\newblock


\bibitem[\protect\citeauthoryear{Sablayrolles, Douze, Schmid, Ollivier, and
  J{\'e}gou}{Sablayrolles et~al\mbox{.}}{2019}]%
        {SDSOJ19}
\bibfield{author}{\bibinfo{person}{Alexandre Sablayrolles},
  \bibinfo{person}{Matthijs Douze}, \bibinfo{person}{Cordelia Schmid},
  \bibinfo{person}{Yann Ollivier}, {and} \bibinfo{person}{Herv{\'e}
  J{\'e}gou}.} \bibinfo{year}{2019}\natexlab{}.
\newblock \showarticletitle{{White-box vs Black-box: Bayes Optimal Strategies
  for Membership Inference}}. In \bibinfo{booktitle}{\emph{{International
  Conference on Machine Learning (ICML)}}}. \bibinfo{publisher}{JMLR},
  \bibinfo{pages}{5558--5567}.
\newblock


\bibitem[\protect\citeauthoryear{Salem, Zhang, Humbert, Berrang, Fritz, and
  Backes}{Salem et~al\mbox{.}}{2019}]%
        {SZHBFB19}
\bibfield{author}{\bibinfo{person}{Ahmed Salem}, \bibinfo{person}{Yang Zhang},
  \bibinfo{person}{Mathias Humbert}, \bibinfo{person}{Pascal Berrang},
  \bibinfo{person}{Mario Fritz}, {and} \bibinfo{person}{Michael Backes}.}
  \bibinfo{year}{2019}\natexlab{}.
\newblock \showarticletitle{{ML-Leaks: Model and Data Independent Membership
  Inference Attacks and Defenses on Machine Learning Models}}. In
  \bibinfo{booktitle}{\emph{{Network and Distributed System Security Symposium
  (NDSS)}}}. \bibinfo{publisher}{Internet Society}.
\newblock


\bibitem[\protect\citeauthoryear{Salimans, Goodfellow, Zaremba, Cheung,
  Radford, and Chen}{Salimans et~al\mbox{.}}{2016}]%
        {SGZCRC16}
\bibfield{author}{\bibinfo{person}{Tim Salimans}, \bibinfo{person}{Ian~J.
  Goodfellow}, \bibinfo{person}{Wojciech Zaremba}, \bibinfo{person}{Vicki
  Cheung}, \bibinfo{person}{Alec Radford}, {and} \bibinfo{person}{Xi Chen}.}
  \bibinfo{year}{2016}\natexlab{}.
\newblock \showarticletitle{{Improved Techniques for Training GANs}}. In
  \bibinfo{booktitle}{\emph{{Annual Conference on Neural Information Processing
  Systems (NIPS)}}}. \bibinfo{publisher}{NIPS}, \bibinfo{pages}{2226--2234}.
\newblock


\bibitem[\protect\citeauthoryear{Shokri, Stronati, Song, and Shmatikov}{Shokri
  et~al\mbox{.}}{2017}]%
        {SSSS17}
\bibfield{author}{\bibinfo{person}{Reza Shokri}, \bibinfo{person}{Marco
  Stronati}, \bibinfo{person}{Congzheng Song}, {and} \bibinfo{person}{Vitaly
  Shmatikov}.} \bibinfo{year}{2017}\natexlab{}.
\newblock \showarticletitle{{Membership Inference Attacks Against Machine
  Learning Models}}. In \bibinfo{booktitle}{\emph{{IEEE Symposium on Security
  and Privacy (S\&P)}}}. \bibinfo{publisher}{IEEE}, \bibinfo{pages}{3--18}.
\newblock


\bibitem[\protect\citeauthoryear{Simonyan and Zisserman}{Simonyan and
  Zisserman}{2015}]%
        {SZ15}
\bibfield{author}{\bibinfo{person}{Karen Simonyan} {and}
  \bibinfo{person}{Andrew Zisserman}.} \bibinfo{year}{2015}\natexlab{}.
\newblock \showarticletitle{{Very Deep Convolutional Networks for Large-Scale
  Image Recognition}}. In \bibinfo{booktitle}{\emph{{International Conference
  on Learning Representations (ICLR)}}}.
\newblock


\bibitem[\protect\citeauthoryear{Szegedy, Liu, Jia, Sermanet, Reed, Anguelov,
  Erhan, Vanhoucke, and Rabinovich}{Szegedy et~al\mbox{.}}{2015}]%
        {SLJSRAEVR15}
\bibfield{author}{\bibinfo{person}{Christian Szegedy}, \bibinfo{person}{Wei
  Liu}, \bibinfo{person}{Yangqing Jia}, \bibinfo{person}{Pierre Sermanet},
  \bibinfo{person}{Scott~E. Reed}, \bibinfo{person}{Dragomir Anguelov},
  \bibinfo{person}{Dumitru Erhan}, \bibinfo{person}{Vincent Vanhoucke}, {and}
  \bibinfo{person}{Andrew Rabinovich}.} \bibinfo{year}{2015}\natexlab{}.
\newblock \showarticletitle{{Going Deeper with Convolutions}}. In
  \bibinfo{booktitle}{\emph{{IEEE Conference on Computer Vision and Pattern
  Recognition (CVPR)}}}. \bibinfo{publisher}{IEEE}, \bibinfo{pages}{1--9}.
\newblock


\bibitem[\protect\citeauthoryear{Tieleman and Hinton}{Tieleman and
  Hinton}{2012}]%
        {tieleman2012lecture}
\bibfield{author}{\bibinfo{person}{Tijmen Tieleman} {and}
  \bibinfo{person}{Geoffrey Hinton}.} \bibinfo{year}{2012}\natexlab{}.
\newblock \showarticletitle{Lecture 6.5-rmsprop: Divide the Gradient by a
  Running average of its Recent Magnitude}.
\newblock \bibinfo{journal}{\emph{COURSERA: Neural Networks for Machine
  Learning}} \bibinfo{volume}{4}, \bibinfo{number}{2} (\bibinfo{year}{2012}),
  \bibinfo{pages}{26--31}.
\newblock


\bibitem[\protect\citeauthoryear{van~den Oord, Dieleman, Zen, Simonyan,
  Vinyals, Graves, Kalchbrenner, Senior, and Kavukcuoglu}{van~den Oord
  et~al\mbox{.}}{2016}]%
        {ODZSVGKSK16}
\bibfield{author}{\bibinfo{person}{Aaron van~den Oord}, \bibinfo{person}{Sander
  Dieleman}, \bibinfo{person}{Heiga Zen}, \bibinfo{person}{Karen Simonyan},
  \bibinfo{person}{Oriol Vinyals}, \bibinfo{person}{Alex Graves},
  \bibinfo{person}{Nal Kalchbrenner}, \bibinfo{person}{Andrew Senior}, {and}
  \bibinfo{person}{Koray Kavukcuoglu}.} \bibinfo{year}{2016}\natexlab{}.
\newblock \showarticletitle{{WaveNet: A Generative Model for Raw Audio}}.
\newblock \bibinfo{journal}{\emph{{CoRR abs/1609.03499}}}
  (\bibinfo{year}{2016}).
\newblock


\bibitem[\protect\citeauthoryear{Vinyals, Toshev, Bengio, and Erhan}{Vinyals
  et~al\mbox{.}}{2015}]%
        {VTBE15}
\bibfield{author}{\bibinfo{person}{Oriol Vinyals}, \bibinfo{person}{Alexander
  Toshev}, \bibinfo{person}{Samy Bengio}, {and} \bibinfo{person}{Dumitru
  Erhan}.} \bibinfo{year}{2015}\natexlab{}.
\newblock \showarticletitle{{Show and Tell: A Neural Image Caption Generator}}.
  In \bibinfo{booktitle}{\emph{{IEEE Conference on Computer Vision and Pattern
  Recognition (CVPR)}}}. \bibinfo{publisher}{IEEE},
  \bibinfo{pages}{3156--3164}.
\newblock


\bibitem[\protect\citeauthoryear{Xie, Lin, Wang, Wang, and Zhou}{Xie
  et~al\mbox{.}}{2018}]%
        {XLWWZ18}
\bibfield{author}{\bibinfo{person}{Liyang Xie}, \bibinfo{person}{Kaixiang Lin},
  \bibinfo{person}{Shu Wang}, \bibinfo{person}{Fei Wang}, {and}
  \bibinfo{person}{Jiayu Zhou}.} \bibinfo{year}{2018}\natexlab{}.
\newblock \showarticletitle{{Differentially Private Generative Adversarial
  Network}}.
\newblock \bibinfo{journal}{\emph{{CoRR abs/1802.06739}}}
  (\bibinfo{year}{2018}).
\newblock


\bibitem[\protect\citeauthoryear{Yan, Yang, Sohn, and Lee}{Yan
  et~al\mbox{.}}{2016}]%
        {YYSL16}
\bibfield{author}{\bibinfo{person}{Xinchen Yan}, \bibinfo{person}{Jimei Yang},
  \bibinfo{person}{Kihyuk Sohn}, {and} \bibinfo{person}{Honglak Lee}.}
  \bibinfo{year}{2016}\natexlab{}.
\newblock \showarticletitle{{Attribute2Image: Conditional Image Generation from
  Visual Attributes}}. In \bibinfo{booktitle}{\emph{{European Conference on
  Computer Vision (ECCV)}}}. \bibinfo{publisher}{Springer},
  \bibinfo{pages}{776--791}.
\newblock


\bibitem[\protect\citeauthoryear{Yeom, Giacomelli, Fredrikson, and Jha}{Yeom
  et~al\mbox{.}}{2018}]%
        {YGFJ18}
\bibfield{author}{\bibinfo{person}{Samuel Yeom}, \bibinfo{person}{Irene
  Giacomelli}, \bibinfo{person}{Matt Fredrikson}, {and} \bibinfo{person}{Somesh
  Jha}.} \bibinfo{year}{2018}\natexlab{}.
\newblock \showarticletitle{{Privacy Risk in Machine Learning: Analyzing the
  Connection to Overfitting}}. In \bibinfo{booktitle}{\emph{{IEEE Computer
  Security Foundations Symposium (CSF)}}}. \bibinfo{publisher}{IEEE},
  \bibinfo{pages}{268--282}.
\newblock


\bibitem[\protect\citeauthoryear{Yi, Walia, and Babyn}{Yi
  et~al\mbox{.}}{2019}]%
        {yi2019generative}
\bibfield{author}{\bibinfo{person}{Xin Yi}, \bibinfo{person}{Ekta Walia}, {and}
  \bibinfo{person}{Paul Babyn}.} \bibinfo{year}{2019}\natexlab{}.
\newblock \showarticletitle{Generative Adversarial Network in Medical Imaging:
  A Review}.
\newblock \bibinfo{journal}{\emph{Medical Image Analysis}}
  (\bibinfo{year}{2019}), \bibinfo{pages}{101552}.
\newblock


\bibitem[\protect\citeauthoryear{Yu, Barnes, Shechtman, Amirghodsi, and
  Luk{\'{a}}c}{Yu et~al\mbox{.}}{2019}]%
        {YBSAL19}
\bibfield{author}{\bibinfo{person}{Ning Yu}, \bibinfo{person}{Connelly Barnes},
  \bibinfo{person}{Eli Shechtman}, \bibinfo{person}{Sohrab Amirghodsi}, {and}
  \bibinfo{person}{Michal Luk{\'{a}}c}.} \bibinfo{year}{2019}\natexlab{}.
\newblock \showarticletitle{{Texture Mixer: A Network for Controllable
  Synthesis and Interpolation of Texture}}. In \bibinfo{booktitle}{\emph{{IEEE
  Conference on Computer Vision and Pattern Recognition (CVPR)}}}.
  \bibinfo{publisher}{IEEE}, \bibinfo{pages}{12164--12173}.
\newblock


\bibitem[\protect\citeauthoryear{Yu, Li, Zhou, Malik, Davis, and Fritz}{Yu
  et~al\mbox{.}}{2020}]%
        {YLZMDF20}
\bibfield{author}{\bibinfo{person}{Ning Yu}, \bibinfo{person}{Ke Li},
  \bibinfo{person}{Peng Zhou}, \bibinfo{person}{Jitendra Malik},
  \bibinfo{person}{Larry Davis}, {and} \bibinfo{person}{Mario Fritz}.}
  \bibinfo{year}{2020}\natexlab{}.
\newblock \showarticletitle{{Inclusive GAN: Improving Data and Minority
  Coverage in Generative Models}}. In \bibinfo{booktitle}{\emph{{European
  Conference on Computer Vision (ECCV)}}}. \bibinfo{publisher}{Springer}.
\newblock


\bibitem[\protect\citeauthoryear{Zhang, Isola, Efros, Shechtman, and
  Wang}{Zhang et~al\mbox{.}}{2018a}]%
        {ZIESW18}
\bibfield{author}{\bibinfo{person}{Richard Zhang}, \bibinfo{person}{Phillip
  Isola}, \bibinfo{person}{Alexei~A. Efros}, \bibinfo{person}{Eli Shechtman},
  {and} \bibinfo{person}{Oliver Wang}.} \bibinfo{year}{2018}\natexlab{a}.
\newblock \showarticletitle{{The Unreasonable Effectiveness of Deep Features as
  a Perceptual Metric}}. In \bibinfo{booktitle}{\emph{{IEEE Conference on
  Computer Vision and Pattern Recognition (CVPR)}}}. \bibinfo{publisher}{IEEE},
  \bibinfo{pages}{586--595}.
\newblock


\bibitem[\protect\citeauthoryear{Zhang, Ji, and Wang}{Zhang
  et~al\mbox{.}}{2018b}]%
        {ZJW18}
\bibfield{author}{\bibinfo{person}{Xinyang Zhang}, \bibinfo{person}{Shouling
  Ji}, {and} \bibinfo{person}{Ting Wang}.} \bibinfo{year}{2018}\natexlab{b}.
\newblock \showarticletitle{{Differentially Private Releasing via Deep
  Generative Model (Technical Report)}}.
\newblock \bibinfo{journal}{\emph{{CoRR abs/1801.01594}}}
  (\bibinfo{year}{2018}).
\newblock


\end{thebibliography}

\appendix

\section{Proof}

\begin{customthm}{5.1}
Given the victim model with parameter $\theta_v$, a query dataset $S$, the membership probability of a query sample $x_i$ is well approximated by the sigmoid of minus calibrated reconstruction error. 
\begin{equation}
P(m_i = 1|\theta_v, x_i, S) \approx \sigma ( -L_{\text{cal}} (x_i,\mathcal{R}(x_i|\mathcal{G}_v))  
\end{equation}
And the optimal attack is equivalent to 
\begin{align}
\mathcal{A}(x_i,\mathcal{M}(\theta_v)) = \mathbbm{1}[L_{\text{cal}}(x_i,\mathcal{R}(x_i|\mathcal{G}_v)) < \epsilon ]
\end{align}
i.e., the attacker checks whether the calibrated reconstruction error of the query sample $x_i$ is smaller than a threshold $\epsilon$.

\begin{proof}

By applying the Bayes rule and the property of sigmoid function $\sigma$, the membership probability can be rewritten as follows~\cite{SDSOJ19}:
\begin{align}
\label{eq:sigmoid}
P(m_i=1|\theta_v, x_i, S)&= \sigma \left(\log \left(\frac{P(\theta_v|m_i=1,x_i,S_{-i})P(m_i=1)}{P(\theta_v|m_i=0,x_i,S_{-i})P(m_i=0)}\right)\right)
\end{align}
where $S_{-i}=S\backslash (x_i,m_i)$, i.e., the whole query set except the query sample $x_i$.

Assuming independence of samples in $\mathcal{S}$ while applying Bayes rule and Product rule, we obtain the following posterior approximation

\begin{align}
P(\theta_v|S) &\propto \prod_{\{j|m_j=1\}} P(x_j|\theta_v)P(\theta_v) \label{eq:posterior_1}\\
& \propto \exp(-\sum_j m_j \cdot l(x_j,
\theta_v))  \label{eq:posterior_2}
\end{align}
with $l(x_j,\theta_v)=L(x_j,\mathcal{R}(x|\mathcal{G}_v))$for brevity.
The \autoref{eq:posterior_1} means that the probability of a certain model parameter is determined by its i.i.d. training set samples. Subsequently, by assuming a uniform prior of the model parameter over the whole parameter space %
and plug in the results from \autoref{eq:kernel density 2} we obtain \autoref{eq:posterior_2}.

By normalizing the posterior in~\autoref{eq:posterior_2}, we obtain
\begin{align}
P(\theta_v|m_i=1,x_i,S_{-i})&= \frac{\exp(-\sum_j m_j \cdot l(x_j,\theta_v))}{\int_{\theta'} \exp(-\sum_j m_j\cdot l(x_j,\theta')) d_{\theta'}} \\
P(\theta_v|m_i=0,x_i,S_{-i})&= \frac{\exp(-\sum_{j\neq i} m_j \cdot l(x_j,\theta_v))}{\int_{\theta'} \exp(-\sum_{j\neq i} m_j\cdot l(x_j,\theta')) d_{\theta'}}
\end{align}
with the following ratio:
\begin{align}
\frac{P(\theta_v|m_i=1,x_i,S)}{P(\theta_v|m_i=0,x_i,S)} &= \frac{\exp(-l(x_i,\theta_v))}{\int_{\theta'}\exp(-l(x_i,\theta'))P(\theta'|S_{-i}) d_{\theta'}}
\end{align}
where
$$P(\theta|S_{-i})=\frac{\exp(-\sum_{j\neq i} m_j \cdot l(x_j,\theta))}{\int_{\theta'} \exp(-\sum_{j\neq i} m_j\cdot l(x_j,\theta')) d_{\theta'}} $$
Putting things together, we have 
\begin{align}
P(m_i=1|\theta_v, x_i,S) =\sigma [& \log(\frac{P(m_i=1)}{P(m_i=0)})-l(x_i,\theta_v) \nonumber\\
&-\log\left(\int_{\theta'}\exp(-l(x_i,\theta'))P(\theta'|S_{-i}) d_{\theta'}\right) ]
\end{align}
The first term is equivalent to the log ratio of the prior probability, i.e., the fraction of training data in the query set. In most of our experiments, we use a balanced split which makes this term vanish. Thus, only the second and last term will affect the attacker prediction.
Next, we investigate the last term. By applying Jensen's inequality, we can bound the last term from above.
\begin{align}
-\log\left(\int_{\theta'}\exp(-l(x_j,\theta'))P(\theta'|S_{-i}) d_{\theta'} \right) &=  -\log \mathbb{E}_{\theta'} \exp(-l(x_i,\theta')) \nonumber \\
& \leq - \mathbb{E}_{\theta'} \log \exp (-l(x_i,\theta')) \nonumber\\
& =  \mathbb{E}_{\theta'} l(x_i,\theta')
\end{align}
Additionally, we can obtain the lower bound by taking the optimimum over the full parameter space, i.e.

\begin{align}
-\log\left(\int_{\theta'}\exp(-l(x_j,\theta'))P(\theta'|S_{-i}) d_{\theta'} \right) &\geq  -\log \max_{\theta'} \exp(-l(x_i,\theta')) \nonumber \\
& = \min_{\theta'} l(x_i,\theta') 
\end{align}

Under the assumption of a highly peaked posterior, e.g. uni-modal Gaussian~\cite{SDSOJ19}, we can well approximate this quantity by using one sample, i.e. using one reference model that is not trained on the query sample. Formally,
\begin{align}
P(m_i=1|\theta_v, x_i,S) &\approx \sigma \left[ -l(x_i,\theta_v) + l(x_i,\theta_r)\right] \nonumber\\
& = \sigma \left[ -L(x,\mathcal{R}(x|\mathcal{G}_v))+L(x,\mathcal{R}(x|\mathcal{G}_r)\right] \nonumber\\
&=\sigma \left[ -L_{\text{cal}}(x,\mathcal{R}(x|\mathcal{G}_v)\right]
\end{align}
where the dependence on $S,\theta_v$ is absorbed in the calibrated distance $L_{\text{cal}}(x,\mathcal{R}(x|\mathcal{G}_v))$.

Hence, the optimal attacker classifies $x_i$ as in the training set if the membership probability is sufficiently large, i.e., $L_{\text{cal}}(x,\mathcal{R}(x|\mathcal{G}_v))$ is sufficiently small (than a threshold), following from the non-decreasing property of $\sigma$. 
\end{proof}
\end{customthm}

\section{Experiment Setup}
\subsection{Hyper-parameter Setting}

We fix $k$ to be 20k for evaluating the full black-box attacks.
We set $\lambda_1 = 1.0$, $\lambda_2 = 0.2$, $\lambda_3 = 0.001$ for our partial black-box and white-box attack on CelebA, and set $\lambda_1 = 1.0$, $\lambda_2 = 0.0$, $\lambda_3 = 0.0$ for the other cases. The maximum number of iterations for optimization are set to be 1000 for our white-box attack and 10 for our partial black-box attack.

\subsection{Model Architectures}
We use the official implementations of the victim GAN models.\protect\footnote{\url{https://github.com/tkarras/progressive_growing_of_gans},\\
\url{https://github.com/igul222/improved_wgan_training},\\
\url{https://github.com/carpedm20/DCGAN-tensorflow},\\
\url{https://github.com/mp2893/medgan},\\
\url{https://drive.google.com/drive/folders/10RCFaA8kOgkRHXIJpXIWAC-uUyLiEhlY}
} We re-implement WGANGP model with a fully-connected structure for non-image datasets. The network architecture is summarized in \autoref{table:architecture}. The depth of both the generator and discriminator is set to 5. The dimension of the hidden layer is fix to be 512 . We use ReLU as the activation function for the generator and Leaky ReLU with $\alpha=0.2$ for the discriminator, except for the output layer where either the sigmoid or identity function is used. 

\begin{table}[!h]
\centering
 \begin{adjustbox}{max width=\columnwidth}
\begin{tabular}{l|l|l}
\toprule
Generator& Generator  & Discriminator \\
 (MIMIC-\uppercase\expandafter{\romannumeral3}) & (Instagram) & (MIMIC-\uppercase\expandafter{\romannumeral3} and Instagram)\\
\midrule
FC (512) & FC (512) & FC (512) \\
ReLU & ReLU & LeakyReLU (0.2) \\
FC (512) & FC (512) & FC (512) \\
ReLU & ReLU & LeakyReLU (0.2) \\
FC (512) & FC (512) & FC (512) \\
ReLU & ReLU & LeakyReLU (0.2) \\
FC (512) & FC (512) & FC (512) \\
ReLU & ReLU & LeakyReLU (0.2) \\
FC ($\text{dim}(x)$) & FC ($\text{dim}(x)$) & FC (1) \\
Sigmoid & Identity & Identity \\
\bottomrule
\end{tabular}
\end{adjustbox}
\caption{Network architecutre of WGANGP on MIMIC-\uppercase\expandafter{\romannumeral3} and Instagram. }
\label{table:architecture}
\end{table}

\subsection{Implementation of Baseline Attacks}

We provide more details of implementing baseline attacks that are discussed in \autoref{sec:comparisons}. 

\subsubsection{LOGAN} For CelebA, we employ DCGAN as the attack model, which is the same as in the original paper~\cite{HMDC19}. For  MIMIC-\uppercase\expandafter{\romannumeral3} and Instagram, we use WGANGP as the attack model. 

\subsubsection{MC} For implementing MC in the full black-box setting on CelebA, we apply the same process of their best attack on the RGB image dataset: First, we employ principal component analysis (PCA) on a data subset disjoint from the query data. Then, we keep the first 120 PCA components as suggested in the original paper~\cite{HHB19} and apply dimensionality reduction on the generated and query data. Finally, we calculate the Euclidean distance of the projected data and use the median heuristic to choose the threshold for MC attack. 

\section{Additional Results}
\subsection{Sanity-check in the White-box Setting}
\label{sec:wb sanity check}

\subsubsection{Analysis on optimization initialization} Due to the non-convexity of our optimization problem, the choice of initialization is of great importance. We explore three different initialization heuristics in our experiments, including mean ($z_0= \mu$), random ($z_0\sim \mathcal{N}(\mu,\Sigma)$), and nearest neighbour ($z_0 = \text{argmin}_{z\in \{z_i\}_{i=1}^k} \Vert \mathcal{G}_v(z) - x\Vert_2^2 $). We find that the mean and nearest neighbor initializations perform well in practice, and are in general better than random initialization in terms of the successful reconstruction rate (reconstruction error smaller than 0.01). Therefore, we apply the mean and nearest neighbor initialization in parallel, and choose the one with smaller reconstruction error for the attack.

\subsubsection{Analysis on Optimization Method} 
\begin{figure}[!h]
\centering
\includegraphics[width=0.9\columnwidth]{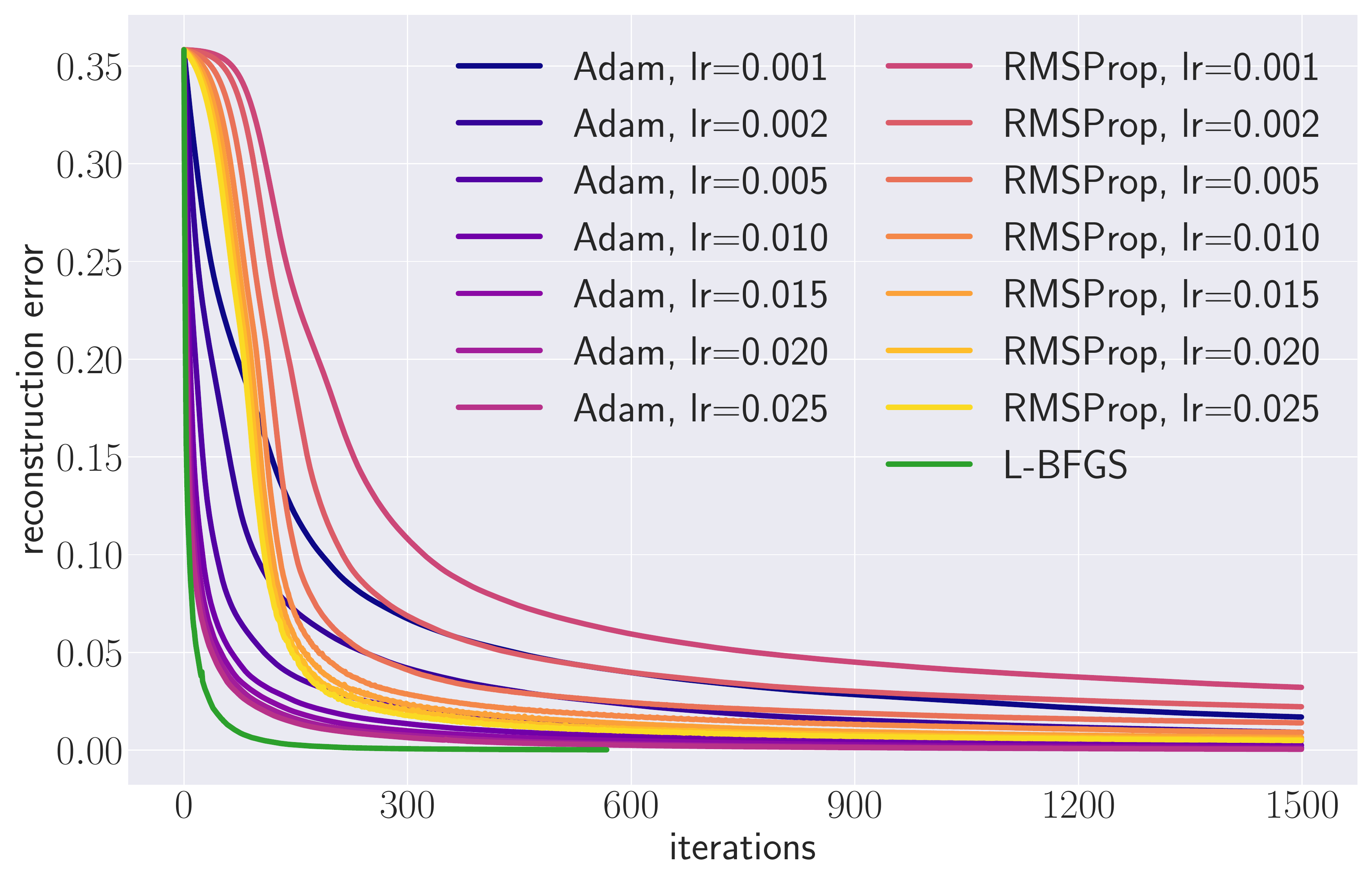}
\caption{Convergence rate of various optimizers (Adam, RMSProp, L-BFGS) with different learning rates. Mean initialization ($z_0 = \mu$) is applied in this analysis study.}
\label{fig:optimizers}
\end{figure}

We explore three optimizers with a range of hyper-parameter search: Adam~\cite{KB15}, RMSProp~\cite{tieleman2012lecture}, and L-BFGS~\cite{liu1989limited} for reconstructing generated samples of PGGAN on CelebA. \autoref{fig:optimizers} shows that L-BFGS achieves superior convergence rate with no additional hyper-parameter. Therefore, we select L-BFGS as our default optimizer in the white-box setting. 

\begin{figure*}[!t]
\centering
\subfigure[]{
\label{fig:reconstruc_l2}
\includegraphics[width=0.6\columnwidth]{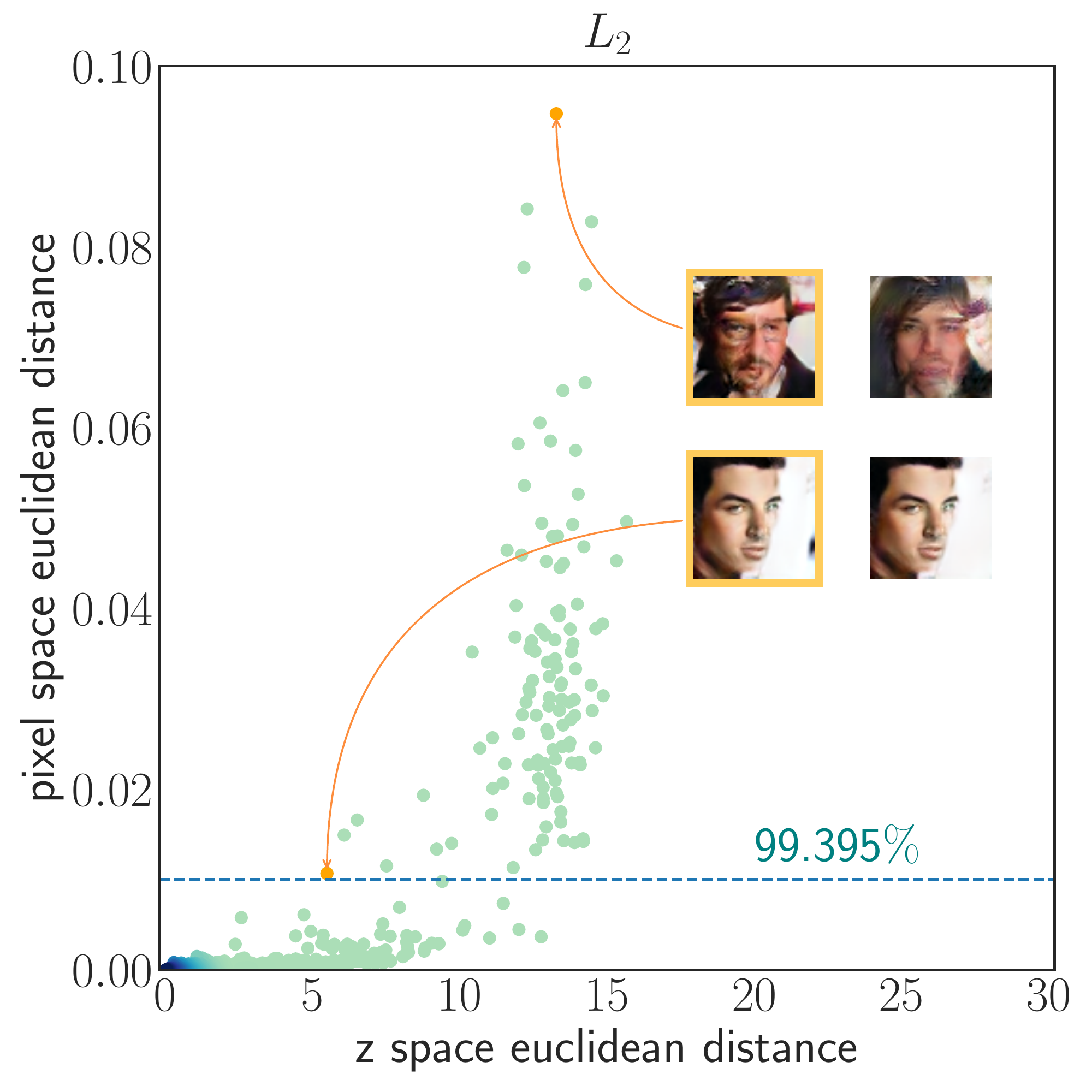}
}
\subfigure[]{
\label{fig:reconstruc_l2lpips}
\includegraphics[width=0.6\columnwidth]{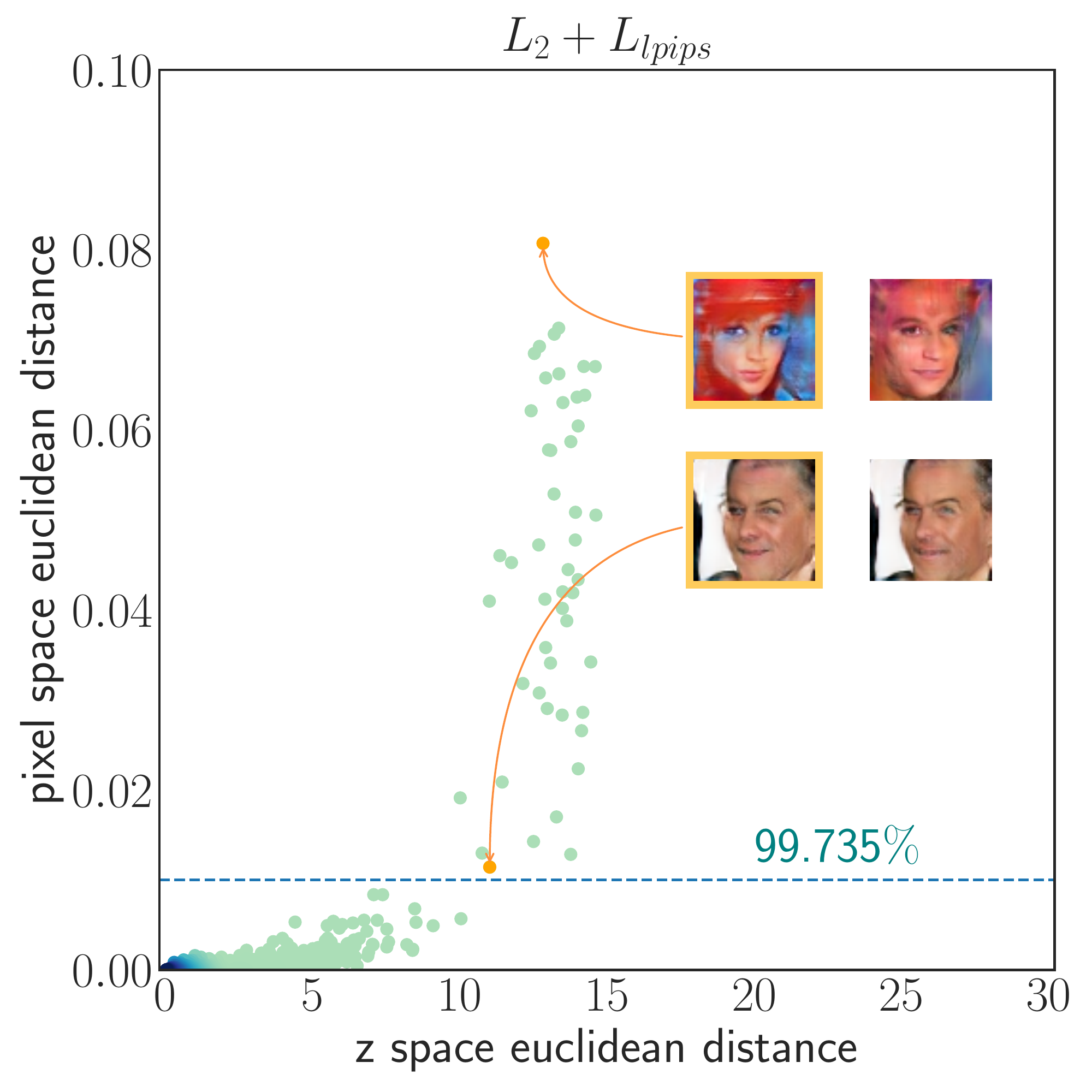}
}
\subfigure[]{
\label{fig:reconstruc_l2lpipsnorm}
\includegraphics[width=0.6\columnwidth]{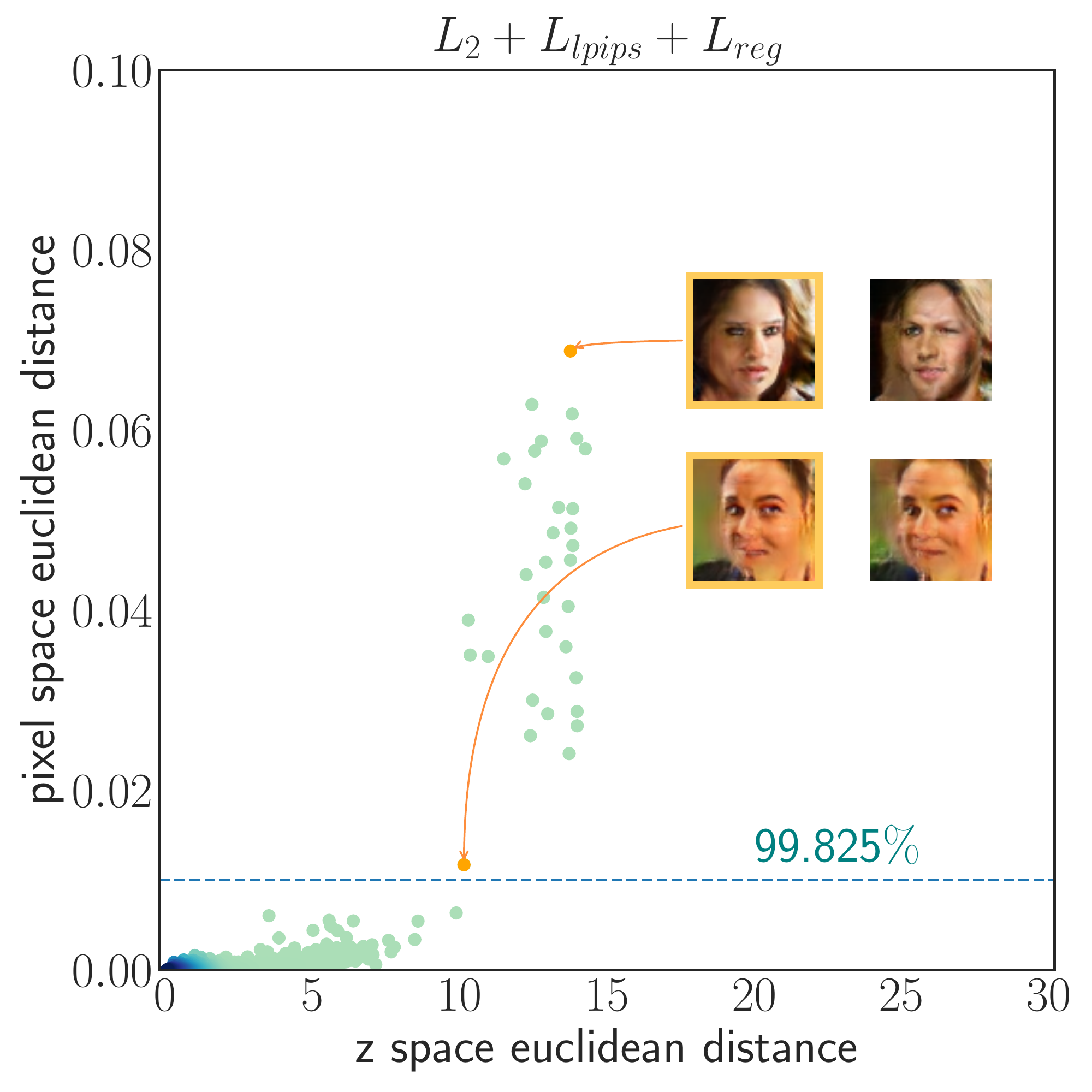}
}
\caption{Reconstruction error plots of PGGAN-generated samples on CelebA. The x-axis represents the Euclidean distance between a reconstructed latent code to its ground truth value. The y-axis represents the $L_2$ residual in the image domain. The images in orange frame are generated samples. Their reconstructed copies are shown on their right. Samples below the dashed line have reconstruction residuals smaller than 0.01, where no visual difference can be observed. Therefore, the reconstruction is in general better if there is a higher portion of sample points below the dashed line (a higher successful reconstruction rate). (a) Reconstruction results when disabling $L_{\text{lpips}}$ and $L_{\text{reg}}$ ($\lambda_1=1.0$, $\lambda_2=0$, $\lambda_3=0$). (b) Reconstruction results when disabling $L_{\text{reg}}$ ($\lambda_1=1.0$, $\lambda_2=0.2$, $\lambda_2=0$). (c) Reconstruction results when enabling all the  $L_2, L_{\text{lpips}}$ and $L_{\text{reg}}$ terms ($\lambda_1=1.0$, $\lambda_1=0.2$, $\lambda_2=0.001$). We find that using all the terms most benefits the reconstruction.}
\label{fig:gen_reconstruction} 
\end{figure*}

\subsubsection{Analysis on Distance Metric Design for Optimization} We show the effectiveness of our objective design (\autoref{eq:distance_metric}). Although optimizing only for element-wise difference term $L_2$ yields reasonably good reconstruction in most cases, we observe undesired blur in reconstruction for CelebA images. Incorporating deep image feature term $L_{\text{lpips}}$ and regularization term $L_{\text{reg}}$ benefits the successful reconstruction rate. See \autoref{fig:gen_reconstruction} for a demonstration.

\begin{table}[!h]
\centering
\caption {Successful reconstruction rate for generated samples from different GANs.}
\label{table:generated_successrate}
\begin{tabular}{lcccc}
\toprule
&  DCGAN & PGGAN & WGANGP & VAEGAN \\  
\midrule
Success rate (\%) & 99.89 & 99.83 &  99.55  & 99.25  
\\
\bottomrule
\end{tabular}
\end{table}

\subsubsection{Sanity Check on Distance Metric Design for Optimization} In addition, we check if the non-convexity of our objective function affects the feasibility of attack against different victim GANs. We apply optimization to reconstruct generated samples. Ideally, the reconstruction should have no error because the query samples are directly generated by the model, i.e., their preimages exist. We set a threshold of $0.01$ to the reconstruction error for counting successful reconstruction rate, and evaluate the success rate for four GAN models trained on CelebA. \autoref{table:generated_successrate} shows that we obtained more than 99\% success rate for all the GANs, which verifies the feasibility of our optimization-based attack. 

\subsubsection{Analysis on Distance Metric Design for Classification} 
\begin{figure}[!h]
\centering
\includegraphics[width=0.7\columnwidth]{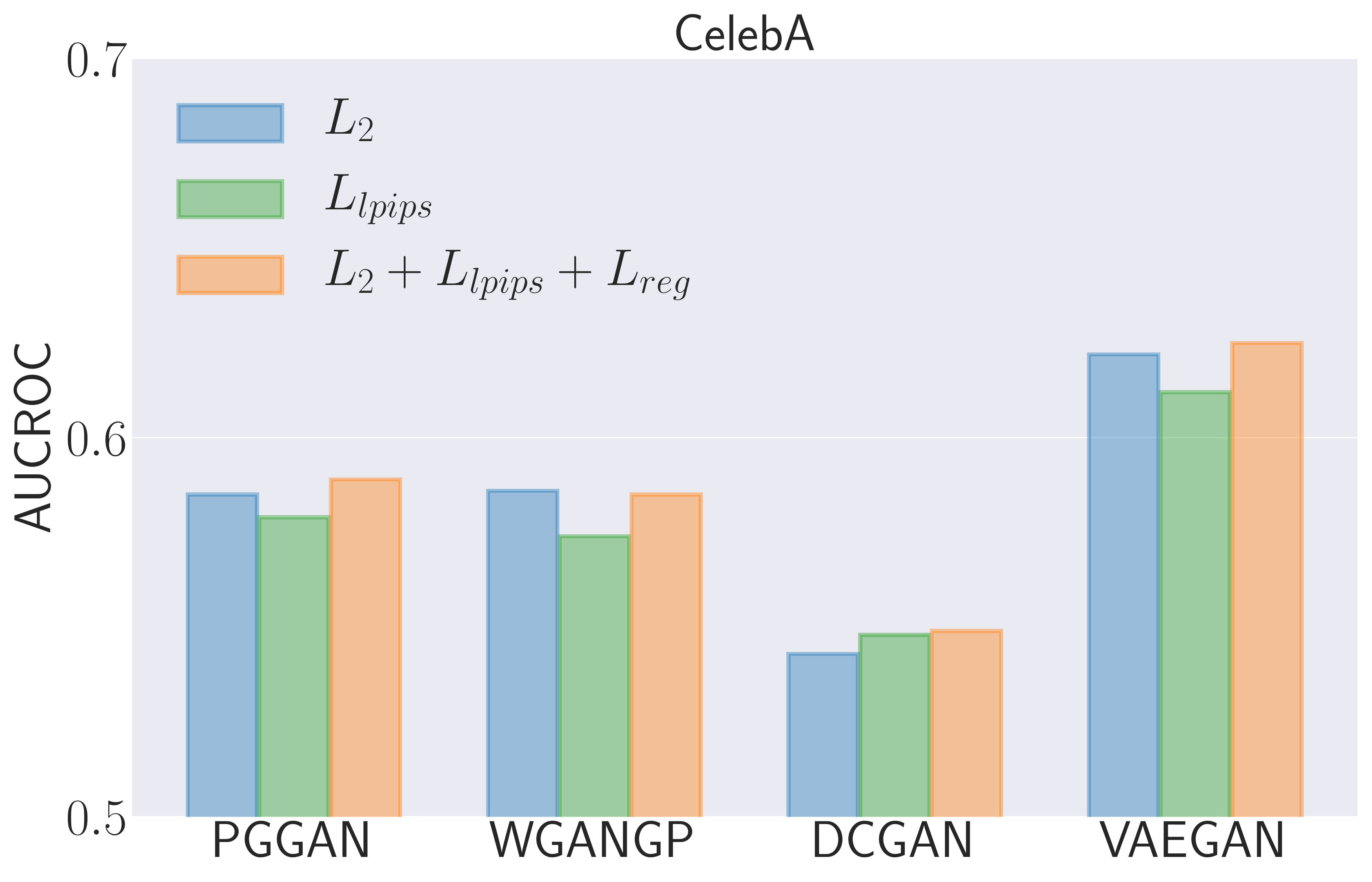}
\caption{White-box attack performance against GANs on CelebA, w.r.t. distance metric design for classification.}
\label{fig:celeba_wb_distance}
\end{figure}

We propose to enable/disable $\lambda_1$, $\lambda_2$, or $\lambda_3$ in \autoref{eq:distance_metric} to investigate the contribution of each term towards classification thresholding (membership inference) on CelebA. In detail, we consider using (1) the element-wise difference term $L_2$ only, (2) the deep image feature term $L_{\text{lpips}}$ only, and (3) all the three terms together to evaluate attack performance. \autoref{fig:celeba_wb_distance} shows the AUCROC of attack against each various GANs. We find that our complete distance metric design achieves general superiority to single terms. Therefore, we use the complete distance metric for classification thresholding.

\subsection{Additional Quantitative Results}

 \subsubsection{Evaluation on Full Black-box Attack}

\myparatight{Attack Performance w.r.t. Training Set Size} \autoref{table:bb_size} corresponds to \autoref{fig:bb_size_1}, \autoref{fig:bb_size_2}, and \autoref{fig:bb_size_3} in the main paper. 

\begin{table}[!h]
\captionsetup{width=\columnwidth}
\centering
\subtable[CelebA]{
\begin{adjustbox}{max width=\columnwidth}
\begin{tabular}{lcccccccc}
\toprule
& 64 & 128 & 256 & 512 & 1024 & 2048 & 4096 & 20k  \\  
\midrule
PGGAN & 1.00 & 1.00 & 1.00 & 0.99 & 0.95 & 0.79 & 0.58 & 0.51 \\
WGANGP & 1.00 & 1.00 & 1.00 & 0.97 & 0.89 & 0.72 & 0.62 & 0.51\\
\bottomrule
\end{tabular}
\end{adjustbox}}
\subtable[MIMIC-\uppercase\expandafter{\romannumeral3}]{
\begin{adjustbox}{max width=\columnwidth}
\begin{tabular}{lcccccccc}
\toprule
& 64 & 128 & 256 & 512 & 1024 & 2048 & 4096 & 8192 \\  
\midrule
WGANGP & 0.98 & 0.97 & 0.93 & 0.87 & 0.81 & 0.68 & 0.54 & 0.52 \\
MEDGAN & 0.78 & 0.65 & 0.57 & 0.54 & 0.52 & 0.52 & 0.51 & 0.51 \\
\bottomrule
\end{tabular}
\end{adjustbox}}
\subtable[Instagram]{
\begin{adjustbox}{max width=\columnwidth}
\begin{tabular}{lccccccccc}
\toprule
& 64 & 128 & 256 & 512 & 1024 & 2048 & 4096 & 8192 & 10k\\  
\midrule
WGANGP & 1.00 & 1.00 & 0.97 & 0.90 & 0.72 & 0.54 & 0.50 & 0.50 & 0.50 \\
\bottomrule
\end{tabular}
\end{adjustbox}
}
\caption{Full black-box attack performance w.r.t. training set size.} 
\label{table:bb_size}
\end{table}

\myparatight{Attack Performance w.r.t. Training Set Selection} \autoref{table:celeba_iden_rand} corresponds to  \autoref{fig:celeba_iden_rand} in the main paper. 

\subsubsection{Evaluation on Partial Black-box Attack} 

\myparatight{Attack Performance w.r.t. Training Set Selection} \autoref{table:celeba_iden_rand} corresponds to  \autoref{fig:celeba_iden_rand} in the main paper. 

\begin{table}[!h]
\captionsetup{width=\columnwidth}
\centering
\subtable[Full black-box]{
\begin{adjustbox}{max width=\columnwidth}
\begin{tabular}{lcccc}
\toprule
&  PGGAN & WGANGP & DCGAN & VAEGAN \\  
\midrule
random & 0.51 & 0.51 & 0.51 & 0.50 \\
identity & 0.53 & 0.53 & 0.51 & 0.51\\
\bottomrule
\end{tabular}
\end{adjustbox}
\label{table:celeba_iden_rand_bb}
}

\subtable[Partial black-box]{
\begin{adjustbox}{max width=\columnwidth}
\begin{tabular}{lcccc}
\toprule
&  PGGAN & WGANGP & DCGAN & VAEGAN \\  
\midrule
random & 0.55 &  0.53 & 0.51 &  0.55 \\
identity & 0.57 & 0.60 & 0.55 & 0.58 \\
 \bottomrule
\end{tabular}
\end{adjustbox}
\label{table:celeba_iden_rand_pbb}
}
    
\subtable[White-box]{
\begin{adjustbox}{max width=\columnwidth}
\begin{tabular}{lcccc}
\toprule
&  PGGAN & WGANGP & DCGAN & VAEGAN \\  
\midrule
random & 0.54 &  0.53 & 0.52 &  0.61 \\
identity & 0.59 & 0.59 & 0.55 & 0.63 \\
\bottomrule
\end{tabular}
\end{adjustbox}
\label{table:celeba_iden_rand_wb}
}
\caption {Attack performance on the random v.s. identity-based GAN training set selection. We only focus on CelebA across attack settings.}
\label{table:celeba_iden_rand}
\end{table}

\subsubsection{Evaluation on White-box Attack}

\myparatight{Ablation on Distance Metric Design for Classification} \autoref{table:celeba_wb_distance} corresponds to \autoref{fig:celeba_wb_distance}. 

\begin{table}[!h]
\captionsetup{width=\columnwidth}
\centering
\begin{adjustbox}{max width=\columnwidth}
\begin{tabular}{lcccc}
\toprule
&  DCGAN & PGGAN & WGANGP & VAEGAN \\  
\midrule
$L_2$ & 0.54 & 0.59 & 0.59 & 0.62\\
$L_{\text{lpips}}$  & 0.55 & 0.58 & 0.57 & 0.61 \\
$L_2 + L_{\text{lpips}} + L_{\text{reg}} $ & 0.55  & 0.59 & 0.59 & 0.63
\\
\bottomrule
\end{tabular}
\end{adjustbox}
\caption {White-box  attack  performance  against  various GANs on CelebA, w.r.t.  distance metric design for classification.}
\label{table:celeba_wb_distance}
\end{table}

\begin{table}[!h]
\captionsetup{width=\columnwidth}
\subtable[CelebA]{
\begin{adjustbox}{max width=\columnwidth}
\begin{tabular}{lcccccccc}
\toprule
& 64 & 128 & 256 & 512 & 1024 & 2048 & 4096 & 20k  \\  
\midrule
PGGAN & 1.00 & 1.00 &  1.00 & 0.99 & 0.95 & 0.83 & 0.62 & 0.55\\
WGANGP & 1.00 & 1.00 &  0.99 & 0.97 & 0.89 & 0.78 & 0.69 & 0.53\\
\bottomrule
\end{tabular}
\end{adjustbox}
}

\subtable[MIMIC-\uppercase\expandafter{\romannumeral3}]{
\begin{adjustbox}{max width=\columnwidth}
\begin{tabular}{lcccccccc}
\toprule
& 64 & 128 & 256 & 512 & 1024 & 2048 & 4096 & 8192 \\  
\midrule
WGANGP & 0.98 & 0.96 & 0.92 & 0.87 & 0.82 & 0.80 & 0.67 & 0.54\\
MEDGAN & 0.99 & 0.88 & 0.77 & 0.72 & 0.61 & 0.59 & 0.55 & 0.52\\
\bottomrule
\end{tabular}
\end{adjustbox}
}

\subtable[Instagram]{
\begin{adjustbox}{max width=\columnwidth}
\begin{tabular}{lccccccccc}
\toprule
& 64 & 128 & 256 & 512 & 1024 & 2048 & 4096 & 8192 & 10k\\  
\midrule
WGANGP & 1.00 & 1.00 & 0.97 & 0.90 & 0.72 & 0.55 & 0.50 & 0.50 & 0.49\\ 
\bottomrule
\end{tabular}
\end{adjustbox}
}
\caption{White-box attack performance w.r.t. training set size.} 
\label{table:wb_size}
\end{table}

\myparatight{Attack Performance w.r.t. Training Set Size} \autoref{table:wb_size} corresponds to \autoref{fig:wb_size_1}, \autoref{fig:wb_size_2}, and \autoref{fig:wb_size_3}.

\begin{table}[!h]
\centering
\subtable[Full black-box]{
\begin{adjustbox}{max width=\columnwidth}
\begin{tabular}{lcccc}
\toprule
&  PGGAN & WGANGP & DCGAN & VAEGAN \\  
\midrule
before calibration & 0.53 & 0.53 & 0.51 & 0.51 \\
after calibration & 0.54 & 0.54 & 0.52 & 0.51 \\
\bottomrule
\end{tabular}
\end{adjustbox}
}

\subtable[Partial black-box]{
\begin{adjustbox}{max width=\columnwidth}
\begin{tabular}{lcccc}
\toprule
&  PGGAN & WGANGP & DCGAN & VAEGAN \\  
\midrule
before calibration & 0.57 & 0.60 & 0.55 & 0.58 \\
after calibration & 0.58 & 0.63 & 0.56 & 0.59 \\
\bottomrule
\end{tabular}
\end{adjustbox}
}

\subtable[White-box]{
\begin{adjustbox}{max width=\columnwidth}
\begin{tabular}{lcccc}
\toprule
&  PGGAN & WGANGP & DCGAN & VAEGAN \\  
\midrule
before calibration & 0.59 & 0.59 & 0.55 & 0.63 \\
after calibration & 0.68 & 0.64 & 0.55 & 0.76 \\
\bottomrule
\end{tabular}
\end{adjustbox}
}
\caption {Attack performance before and after calibration on CelebA.} 
\label{table:celeba_calibration}
\end{table}

\begin{table}[!h]
\centering
\subtable[MIMIC-\uppercase\expandafter{\romannumeral3} (WGANGP)]{
\begin{adjustbox}{max width=\columnwidth}
\begin{tabular}{lcccccccc}
\toprule
& 64 & 128 & 256 & 512 & 1024 & 2048 & 4096 & 8192  \\
\midrule
full bb  & 0.98 & 0.97 & 0.93 & 0.87 & 0.81 & 0.68 & 0.54 & 0.52\\
full bb (calibrated) & 1.00 & 0.99 & 0.97 & 0.94 & 0.89 & 0.84 & 0.67 & 0.56\\
wb  & 0.98 & 0.96 & 0.92 & 0.87 & 0.82 & 0.80 & 0.67 & 0.54\\
wb (calibrated) & 0.98 & 0.97 & 0.93 & 0.90 & 0.87 & 0.85 & 0.75 & 0.59 \\
\bottomrule
\end{tabular}
\end{adjustbox}
}

\subtable[MIMIC-\uppercase\expandafter{\romannumeral3} (MEDGAN)]{
\begin{adjustbox}{max width=\columnwidth}
\begin{tabular}{lcccccccc}
\toprule
& 64 & 128 & 256 & 512 & 1024 & 2048 & 4096 & 8192 \\  
\midrule
full bb &  0.78 & 0.65 & 0.57 & 0.54 & 0.52 & 0.52 & 0.51 & 0.51 \\
full bb (calibrated) & 0.91 & 0.71 & 0.63 & 0.58 & 0.55 & 0.53 & 0.52 & 0.51 \\
wb & 0.99 & 0.88 & 0.77 & 0.72 & 0.61 & 0.59 & 0.55 & 0.52\\
wb (calibrated) & 0.96 & 0.87 & 0.81 & 0.75 & 0.65 & 0.62 & 0.57 & 0.55\\
\bottomrule
\end{tabular}
\end{adjustbox}
}

\subtable[Instagram (WGANGP)]{
\begin{adjustbox}{max width=\columnwidth}
\begin{tabular}{lccccccccc}
\toprule
& 64 & 128 & 256 & 512 & 1024 & 2048 & 4096 & 8192 & 10k\\  
\midrule
full bb & 1.00 & 1.00 & 0.97 & 0.90 & 0.72 & 0.54 & 0.50 & 0.50 & 0.49\\
full bb (calibrated) & 1.00 & 1.00 & 0.98 & 0.91 & 0.80 & 0.72 & 0.65 & 0.57 & 0.56\\
wb & 1.00 & 1.00 & 0.97 & 0.90 & 0.72 & 0.55 & 0.50 & 0.50 & 0.49\\ 
wb (calibrated) & 1.00 & 1.00 & 0.98 & 0.92 & 0.79 & 0.73 & 0.67 & 0.58 & 0.57\\
\bottomrule
\end{tabular}
\end{adjustbox}
}
\caption{Attack performance before and after calibration for non-image datasets w.r.t. GAN training set sizes. \textbf{bb}: black-box; \textbf{wb}: white-box.} 
\label{table:wbbb_calibrate}
\end{table}

\myparatight{Attack Performance w.r.t. Training Set Selection}
\autoref{table:celeba_iden_rand} corresponds to  \autoref{fig:celeba_iden_rand} in the main paper. 

\subsubsection{Attack Calibration}
\autoref{table:celeba_calibration} corresponds to  \autoref{fig:celeba_calibrate_iden} in the main paper. \autoref{table:wbbb_calibrate} corresponds to \autoref{fig:wbbb_calibrate} in the main paper.

\subsubsection{Comparison to Baseline Attacks}
\autoref{table:celeba_allattack} corresponds to \autoref{fig:celeba_iden_allattack} in the main paper. \autoref{table:mimic_instagram_all_attack} corresponds to \autoref{fig:mimic_instagram_allattack} in the main paper. \autoref{table:fbb_numqueries} corresponds to \autoref{fig:fbb_numqueries} in the main paper.

\begin{table}[H]
\centering
\begin{adjustbox}{max width=\columnwidth}
\begin{tabular}{lccccc}
\toprule
&  PGGAN & WGANGP & DCGAN & VAEGAN & VAE \\  
\midrule
full bb (LOGAN) & 0.56 & 0.57 & 0.52 & 0.50 & 0.52\\
full bb (MC) & 0.52 & 0.52 &  0.51 & 0.50 & 0.51\\
full bb (ours calibrated)& 0.54 & 0.54 & 0.52 & 0.51 & 0.54\\
partial bb (ours calibrated)& 0.58 & 0.63 & 0.56 & 0.59& 0.73\\
wb (ours calibrated)& 0.68 & 0.66 & 0.55 & 0.76 & 0.94\\
full (LOGAN/MC) & 0.91 &  0.83 & 0.83 & 0.61 & 0.90\\
\bottomrule
\end{tabular}
\end{adjustbox}
\caption{Comparison of different attacks on CelebA. \textbf{bb}: black-box; \textbf{wb}: white-box; \textbf{full}: accessible discriminator (full model).}
\label{table:celeba_allattack}
\end{table}

 \begin{table}[H]
\captionsetup{width=\columnwidth}
\centering
\subtable[MIMIC-\uppercase\expandafter{\romannumeral3} (WGANGP)]{
\begin{adjustbox}{max width=\columnwidth}
\begin{tabular}{lcccccccc}
\toprule
& 64 & 128 & 256 & 512 & 1024 & 2048 & 4096 & 8192  \\
\midrule
full bb (LOGAN) & 0.98 & 0.97 & 0.96 & 0.94 & 0.92 & 0.83 & 0.65 & 0.54 \\
full bb (ours calibrated) & 1.00 & 0.99 & 0.97 & 0.94 & 0.89 & 0.84 & 0.67 & 0.56\\
wb (ours calibrated) & 0.98 & 0.97 & 0.93 & 0.90 & 0.87 & 0.85 & 0.75 & 0.59 \\
dis (LOGAN) & 1.00 &  1.00 & 1.00 & 1.00 & 1.00 & 1.00 & 0.99 & 0.98 \\
\bottomrule
\end{tabular}
\end{adjustbox}
}
\subtable[MIMIC-\uppercase\expandafter{\romannumeral3} (MEDGAN)]{
\begin{adjustbox}{max width=\columnwidth}
\begin{tabular}{lcccccccc}
\toprule
& 64 & 128 & 256 & 512 & 1024 & 2048 & 4096 & 8192 \\  
\midrule
full bb (LOGAN) & 0.45 & 0.57 & 0.53 & 0.52 & 0.51 & 0.52 & 0.50 & 0.51 \\
full bb (ours calibrated) & 0.91 & 0.71 & 0.63 & 0.58 & 0.55 & 0.53 & 0.52 & 0.51 \\
wb (calibrated) & 0.96 & 0.87 & 0.81 & 0.75 & 0.65 & 0.62 & 0.57 & 0.55\\
dis (LOGAN) & 1.00 & 0.92 & 0.96 & 0.90 & 0.85 & 0.90 & 0.80 & 0.73 \\
\bottomrule
\end{tabular}
\end{adjustbox}
}

\subtable[Instagram (WGANGP)]{
\begin{adjustbox}{max width=\columnwidth}
\begin{tabular}{lccccccccc}
\toprule
& 64 & 128 & 256 & 512 & 1024 & 2048 & 4096 & 8192 & 10k\\  
\midrule
full bb (LOGAN)& 1.00 & 0.99 & 0.96 & 0.91 & 0.68 & 0.55 & 0.58 & 0.55 & 0.55\\
full bb (calibrated) & 1.00 & 1.00 & 0.98 & 0.91 & 0.80 & 0.72 & 0.65 & 0.57 & 0.56\\
wb (calibrated) & 1.00 & 1.00 & 0.98 & 0.92 & 0.79 & 0.73 & 0.67 & 0.58 & 0.57\\
dis (LOGAN) & 1.00 & 1.00 & 1.00 & 1.00 & 1.00 & 1.00 & 0.98 & 0.96 & 0.93\\
\bottomrule
\end{tabular}
\end{adjustbox}
}
\caption{Comparison of different attacks on the other two non-image datasets w.r.t. GAN training set size. \textbf{bb}: black-box; \textbf{wb}: white-box; \textbf{dis}: accessible discriminator.} 
\label{table:mimic_instagram_all_attack}
\end{table}

\begin{table}[H]
\centering
\begin{adjustbox}{max width=\columnwidth}
\begin{tabular}{lcccccccccccccc}
\toprule
$k$ &64 &128 &256 &512 &1024 &2048 &4096 &8192 &15k &20k &40k &60k &80k &100k \\  
\midrule
LOGAN & 0.51 & 0.51 & 0.51 & 0.52 & 0.53 & 0.53 & 0.53 & 0.54  & 0.55 & 0.56 & 0.57 & 0.58 & 0.57 & 0.57\\
MC & 0.50 & 0.50 & 0.51 & 0.51 & 0.51 & 0.51 & 0.52 & 0.52  & 0.52 & 0.52 & 0.52 & 0.53 & 0.53 & 0.53 \\
ours calibrated &  0.51 & 0.51 & 0.51 & 0.52 & 0.52 & 0.52 & 0.53 & 0.53 & 0.54 & 0.54 & 0.54 & 0.55 & 0.55 & 0.55 \\
\bottomrule
\end{tabular}
\end{adjustbox}
\caption{Full black-box attack performance against PGGAN on CelebA w.r.t. $k$ in {\upshape \autoref{eq:fbb_rec}}, the number of generated samples.}
\label{table:fbb_numqueries}
\end{table}

 \subsubsection{Defense}
\autoref{table:defense} corresponds to \autoref{fig:celeba_defense_allattack} in the main paper. \autoref{table:defense_size} corresponds to \autoref{fig:defense_size} in the main paper.

\begin{table}[H]
\centering
\begin{adjustbox}{max width=\columnwidth}
\begin{tabular}{lccc}
\toprule
& full black-box & partial black-box & white-box \\
\midrule
w/o DP & 0.54 & 0.58 & 0.68 \\
w/ DP & 0.53 & 0.56  & 0.59 \\
\bottomrule
\end{tabular}
\end{adjustbox}
\caption {Attack performance against PGGAN on CelebA with or without DP defense.}
\label{table:defense}
\end{table}
 
\begin{table}[H]
\centering
\begin{adjustbox}{max width=\columnwidth}
\begin{tabular}{lccccccc}
\toprule
& 64 & 128 & 256 & 512 & 1024 & 2048 & 4096 \\
\midrule
white-box w/o DP & 1.00 & 1.00 & 1.00 & 0.99 & 0.95 & 0.83 & 0.62 \\
white-box w/ DP & 1.00 & 1.00 & 0.99 & 0.98  & 0.90 & 0.70 & 0.56 \\
full black-box w/o DP & 1.00 & 1.00 & 1.00 & 0.99 & 0.95 & 0.79 & 0.57 \\
full black-box w/ DP & 1.00 & 1.00 & 0.99 & 0.98 & 0.89 & 0.68 & 0.53 \\
\bottomrule
\end{tabular}
\end{adjustbox}
\caption {Attack performance against PGGAN on CelebA with or without DP defense, w.r.t. GAN training set size.}
\label{table:defense_size}
\end{table}

\subsection{Additional Qualitative Results}

Given query samples $x$, we show their reconstruction copies $R(x\vert\mathcal{G}_v)$ and $R(x\vert\mathcal{G}_r)$ obtained in our white-box attack.  

\newpage
\begin{figure*}[!h]
\subfigure[Query (real) images]{
\includegraphics[width=\textwidth]{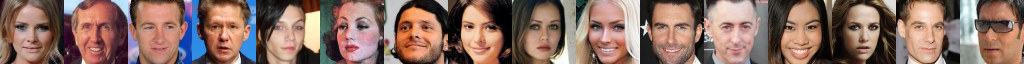}
}

\subfigure[PGGAN victim model reconstruction]{
\includegraphics[width=\textwidth]{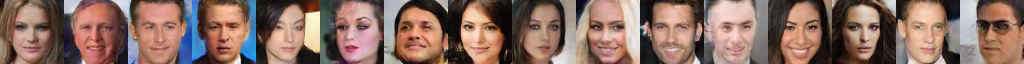}
}

\subfigure[PGGAN (w/ DP) victim model reconstruction]{
\includegraphics[width=\textwidth]{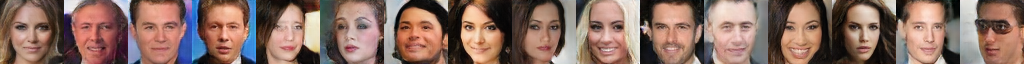}
}

\subfigure[PGGAN reference model reconstruction]{
\includegraphics[width=\textwidth]{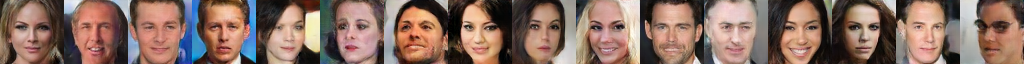}
}

\subfigure[WGANGP victim model reconstruction]{
\includegraphics[width=\textwidth]{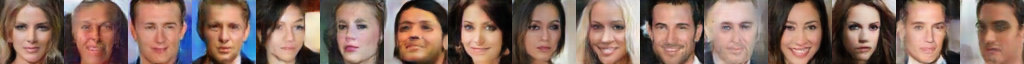}
}

\subfigure[WGANGP reference model reconstruction]{
\includegraphics[width=\textwidth]{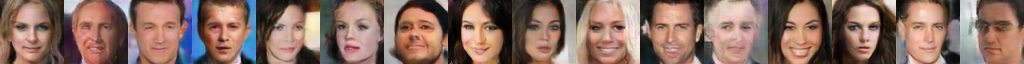}
}

\subfigure[DCGAN victim model reconstruction]{
\includegraphics[width=\textwidth]{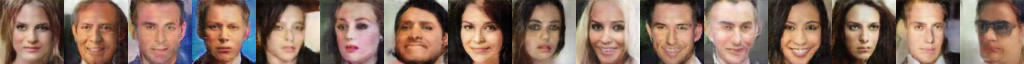}
}

\subfigure[DCGAN reference model reconstruction]{
\includegraphics[width=\textwidth]{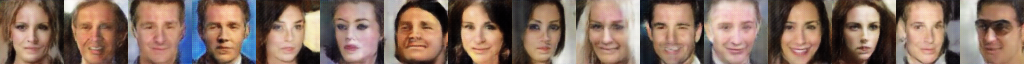}
}

\subfigure[VAEGAN victim model reconstruction]{
\includegraphics[width=\textwidth]{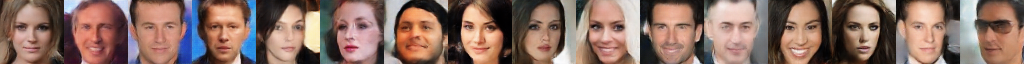}
}

\subfigure[VAEGAN reference model reconstruction]{
\includegraphics[width=\textwidth]{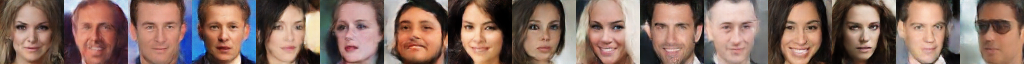}
}
\caption{Reconstruction of query samples $x$ that are in the training set, i.e., $x\in D_{\text{train}}$.}
\label{fig:pos_reconstruction}
\end{figure*}

\begin{figure*}[!h]
\subfigure[Query (real) images]{
\includegraphics[width=\textwidth]{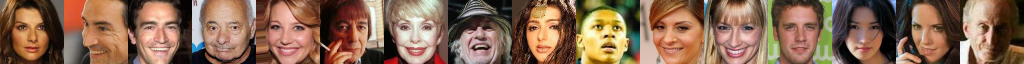}
}
\subfigure[PGGAN victim model reconstruction]{
\includegraphics[width=\textwidth]{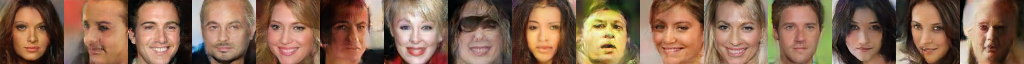}
}
\subfigure[PGGAN (w/ DP) victim model reconstruction]{
\includegraphics[width=\textwidth]{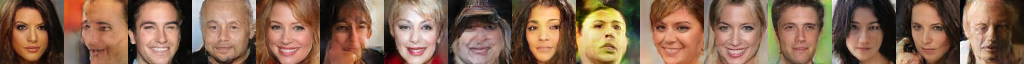}
}
\subfigure[PGGAN reference model reconstruction]{
\includegraphics[width=\textwidth]{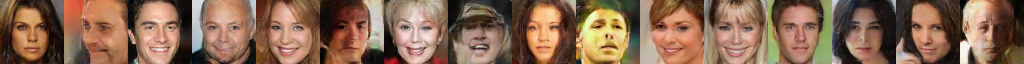}
}
\subfigure[WGANGP victim model reconstruction]{
\includegraphics[width=\textwidth]{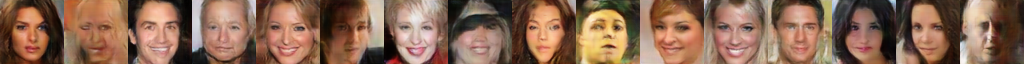}
}
\subfigure[WGANGP reference model reconstruction]{
\includegraphics[width=\textwidth]{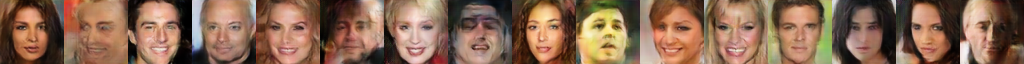}
}
\subfigure[DCGAN victim model reconstruction]{
\includegraphics[width=\textwidth]{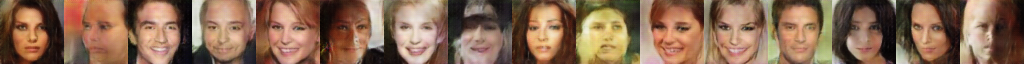}
}
\subfigure[DCGAN reference model reconstruction]{
\includegraphics[width=\textwidth]{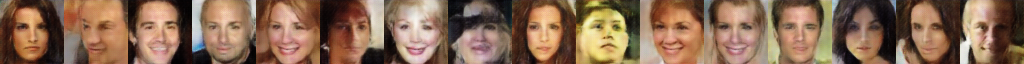}
}
\subfigure[VAEGAN victim model reconstruction]{
\includegraphics[width=\textwidth]{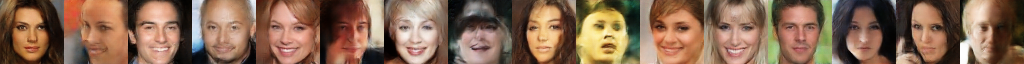}
}
\subfigure[VAEGAN reference model reconstruction]{
\includegraphics[width=\textwidth]{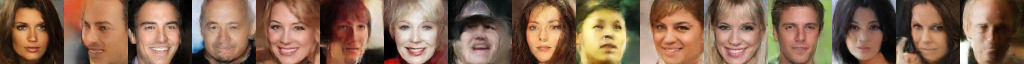}
}
\caption{Reconstruction of query samples $x$ that are \textbf{not} in the training set, i.e., $x\notin D_{\text{train}}$.}
\label{fig:neg_reconstruction}
\end{figure*}

\end{document}